\newif\ifsimplesvp
\simplesvpfalse	

\newif\ifdraft
\draftfalse	

\documentclass[11pt]{article}
\usepackage[verbose, tmargin= .8in, lmargin= .7in, rmargin= .7in, bmargin=.8in]{geometry}

\usepackage{graphicx,amsmath,comment,framed,amssymb,algorithm,algorithmic,subcaption,url,xspace,bbm,amsthm,cite,array}
\usepackage{etoolbox,appendix,color}
\usepackage[pdfusetitle,
bookmarks= true,bookmarksnumbered= false,bookmarksopen= false,
breaklinks= false,pdfborder= {0 0 1}, backref= false, colorlinks= false]
{hyperref}

\newtheorem{thm}{Theorem}
\newtheorem{prop}{Proposition}
\newtheorem{lem}{Lemma}

\newtheorem{dfn}{Definition}

\newtheorem{mdl}{Model}

\newtheorem{conjecture}{Conjecture}

\newcommand{\Prob}{\ensuremath{\mathbb{P}}}
\newcommand{\Exs}{\ensuremath{\mathbb{E}}}
\newcommand{\defn}{\ensuremath{: \, =  }}

\newcommand{\real}{\ensuremath{\mathbb{R}}}
\newcommand{\SymMat}[1]{\ensuremath{\mathcal{S}^{#1 \times #1}}}
\newcommand{\Id}{\ensuremath{\mathcal{I}}} 
\newcommand{\onevec}{\ensuremath{\mathbf{1}}} 

\newcommand{\inprod}[2]{\ensuremath{\langle #1 , \, #2 \rangle}}

\DeclareMathOperator{\rank}{rank}

\DeclareMathOperator{\diag}{diag}

\newcommand{\tru}{M^*}
\newcommand{\truf}{F^*}

\newcommand{\MC}{\textup{MC}}
\newcommand{\SMC}{\textup{SMC}}

\newcommand{\inco}{\mu}
\newcommand{\ob}{\Omega}
\newcommand{\rk}{r}
\newcommand{\dm}{d}
\newcommand{\proj}{\Pi_{\ob}}
\newcommand{\projk}[1]{\Pi_{\ob_{#1}}}
\newcommand{\rproj}{\mathcal{R}_\ob}
\newcommand{\rprojk}[1]{\mathcal{R}_{\ob_{#1}}}
\newcommand{\hp}{\mathcal{H}_\ob}
\newcommand{\hpk}[1]{\mathcal{H}_{\ob_{#1}}}
\newcommand{\hpp}[1]{\mathcal{H}_\ob^{(#1) }}
\newcommand{\iter}[1]{F^{#1}}
\newcommand{\iiter}[2]{F^{#1, #2}}

\newcommand{\er}[1]{E^{#1}}
\newcommand{\err}[2]{E^{#1,#2}}
\newcommand{\ddf}[1]{\Delta^{#1}}

\newcommand{\inc}{\sqrt{ \frac{\inco \rk}{\dm} }}
\newcommand{\add}{\frac{1}{64^2 \inco \rk^2}}
\newcommand{\lic}[1]{\left(\frac{1}{2}\right)^{#1}}

\newcommand{\incc}{\frac{\inco \rk}{\dm} }
\newcommand{\td}[1]{\tilde{D}^{#1}}
\newcommand{\rhoo}{\frac{1}{64^2}}

\newcommand{\pj}{\mathcal{P}_{\mathcal{T}}}
\newcommand{\pjp}{\mathcal{P}_{\mathcal{T}^\perp}}
\newcommand{\To}{t_0}
\newcommand{\hpw}{\mathcal{H}_{\ob}^{(-w)}}
\newcommand{\hpww}{\mathcal{H}_{\ob}^{(w)}}
\newcommand{\whp}{\textit{w.h.p.}\xspace}
\newcommand{\W}{W}
\newcommand{\Z}{Z}
\newcommand{\svp}{\mathcal{P}_r}
\newcommand{\iterl}[1]{\Lambda^{#1}}
\newcommand{\iterm}[1]{M^{#1}}

\newcommand{\sss}{\sigma_{\rk}}
\newcommand{\ls}{\sigma_{1}}
\newcommand{\cond}{\kappa }
\newcommand{\fulladd}{\frac{1}{2^{16} \cond\inco \rk^2}}
\newcommand{\Hp}[1]{H^{#1}}
\newcommand{\G}[1]{G^{#1}}
\newcommand{\true}{\Lambda^*}
\newcommand{\tot}{5\log _2\dm +12}
\newcommand{\fullindb}[1]{\fulladd\lic{#1}\inc}
\newcommand{\pw}{\mathcal{P}_w}
\newcommand{\g}{\mathcal{F}}
\newcommand{\iq}[1]{\theta^{#1}}
\newcommand{\qd}{\delta}
\newcommand{\qdt}[1]{\delta_{#1}}
\newcommand{\loo}{leave-one-out\xspace}
\newcommand{\Loo}{Leave-One-Out\xspace}
\newcommand{\NNM}{NNM\xspace}
\newcommand{\PGD}{PGD\xspace}

\newcommand{\matsnorm}[2]{\| #1\|_{#2}}
\newcommand{\vecnorm}[2]{\| #1\|_{#2}}
\newcommand{\abs}[1]{| #1 |}
\newcommand{\twonorm}[1]{\vecnorm{#1}{2}}

\newcommand{\opnorm}[1]{\ensuremath{\matsnorm{#1}{\mbox{\tiny{op}}}}}
\newcommand{\nucnorm}[1]{\ensuremath{\matsnorm{#1}{\mbox{\tiny{nuc}}}}}
\newcommand{\fronorm}[1]{\ensuremath{\matsnorm{#1}{\mbox{\tiny{F}}}}}
\newcommand{\infnorm}[1]{\ensuremath{\vecnorm{#1}{\infty}}}
\newcommand{\twoinfnorm}[1]{\ensuremath{\matsnorm{#1}{2,\infty}}}
\newcommand{\twotwoinfnorm}[1]{\ensuremath{\matsnorm{#1}{(2,\infty)^2}}}

\newcommand{\bigo}{\mathcal{O}}
\newcommand{\indic}{\mathbbm{1}}

\newcommand{\niea}{\overset{(a)}{\leq}}
\newcommand{\nieb}{\overset{(b)}{\leq}}
\newcommand{\niec}{\overset{(c)}{\leq}}

\newcommand{\nia}{(a)}
\newcommand{\nib}{(b)}
\newcommand{\nic}{(c)}

\newcommand{\citep}[2][]{%
	\ifstrempty{#1}{%
		\hspace{1sp}\cite{#2}%
	}{%
		\hspace{1sp}\cite[#1]{#2}%
	}%
}
\newcommand{\citet}[2][]{%
	\ifstrempty{#1}{%
		\hspace{1sp}\cite{#2}%
	}{%
		\hspace{1sp}\cite[#1]{#2}%
	}%
}
\newcommand{\citealt}[2][]{%
	\ifstrempty{#1}{%
		\hspace{1sp}\cite{#2}%
	}{%
		\hspace{1sp}\cite[#1]{#2}%
	}%
}
\newcommand{\citealp}[2][]{%
	\ifstrempty{#1}{%
		\hspace{1sp}\cite{#2}%
	}{%
		\hspace{1sp}\cite[#1]{#2}%
	}%
}

\newcommand{\blue}[1]{\textcolor{blue}{#1}}

\newcommand{\yccomment}[1]{\ifdraft{\bf{{\blue{{Yudong--- #1}}}}}\else\fi}

\begin{document}
\title{Leave-one-out Approach for Matrix Completion:\\Primal and Dual Analysis}

\author{Lijun Ding
        and Yudong Chen
\thanks{L. Ding and Y. Chen are  with the School of Operations Research and Information 
Engineering, Cornell University, Ithaca, NY, 14850 USA. E-mail: \{ld446, yudong.chen\}@cornell.edu
}}

\date{}


%
%

\maketitle

\begin{abstract}
In this paper, we introduce a powerful  technique based on \Loo analysis to the study of low-rank matrix completion problems. Using this technique, we  develop a general approach for obtaining fine-grained, \emph{entrywise} bounds for iterative stochastic procedures in the presence of probabilistic dependency. We demonstrate the power of this approach in analyzing two of the most  important algorithms for matrix completion: (i) the non-convex approach based on Projected Gradient Descent (\PGD) for a rank-constrained formulation, also known as the Singular Value Projection algorithm, and (ii) the convex relaxation approach  based on nuclear norm minimization (\NNM). 

Using this approach, we establish the first convergence guarantee for the original form of \PGD \emph{without regularization or sample splitting}, and in particular shows that it converges linearly in the \emph{infinity norm}. For \NNM, we use this approach to study a fictitious iterative procedure that arises in the \emph{dual analysis}. Our results show that \NNM recovers an $ \dm $-by-$ \dm $ rank-$ \rk $  matrix with  $\mathcal{O}(\inco \rk \log(\inco \rk)\dm \log\dm )$ observed entries. This bound has optimal dependence on the matrix dimension and is independent of the condition number. To the best of our knowledge, none of previous sample complexity results for tractable matrix completion algorithms satisfies these two properties simultaneously.
\end{abstract}


\section{Introduction}

The matrix completion problem  concerns recovering a low-rank matrix given a (typically random) subset of its entries. To study the sample complexity and algorithmic behaviors of this problem, one often needs to analyze an iterative procedure in the presence of dependency across the iterations and the entries of the iterates. Such dependency creates significant difficulties in both the design and analysis of algorithms, often leading to sub-optimal bounds as well as complicated and unrealistic algorithms that are not used in practice. 

To overcome these challenges, in this paper we introduce a powerful technique, based on the \Loo argument, to the study of matrix completion problems. \Loo, as an \emph{analytical} technique, allows one to isolate the effect of  dependency on individual entries, and establish \emph{entrywise} bounds for the iterative procedures.  We use this technique to obtain new theoretical guarantees for two archetypal algorithms for matrix completion: (i) the non-convex approach based on applying Projected Gradient Descent (\PGD) to a rank-constrained formulation, also known as the Singular Value Projection (SVP) algorithm~\citealt{jain2010guaranteed}; (ii) the convex relaxation method based on Nuclear Norm Minimization (\NNM)~\citealt{candes2009exact}.  We use \loo in two distinct ways. For \PGD, we employ this technique to study the \emph{primal} solution path of the algorithm.  For \NNM, we analyze an iterative procedure that arises in the \emph{analysis}, particularly for constructing a \emph{dual} solution that certifies the optimality of the desired primal solution. \\

Concretely, consider the problem of recovering a rank-$ \rk $ matrix $\tru\in \real^{\dm_1 \times \dm_2}$  given a subset of its entries,  $\{\tru_{ij}: (i,j)\in \ob \}$. As this problem is ill-posed for general $ \ob $, it is standard to assume that $ \ob $ is generated according to some probabilistic model. 
To recover $ \tru $, a natural idea is to seek a low-rank matrix that is consistent with the observations in $ \ob $. 
Based on this idea, two most representative algorithms for matrix completion are the following:

\paragraph{Projected gradient descent for rank-constrained formulation} 

This approach is based on solving a natural rank-constrained, least-squares formulation for matrix completion:
\begin{align}\label{minisvp} 
\mbox{minimize}_{X \in\real^{\dm_1\times \dm_2}} \; \frac{1}{2}\fronorm{\proj(X)-\proj(\tru)}^2 \quad \mbox{subject to}\; \rank{(X)}\leq \rk,
\end{align}
where $ \proj(\cdot): \real^{\dm_1 \times \dm_2} \to \real^{\dm_1 \times \dm_2}$ is the linear operator that zeros out the entries outside $ \ob $. \PGD applied to the above optimization problem takes the form 
\begin{equation}\label{realsvp}
\begin{aligned} 
\iterm{0} = 0; \qquad  
\iterm{t+1}= \svp \Big(\iterm{t} -\eta_t \big( \proj(\iterm{t})-\proj(\tru) \big) \Big), \;\; t = 0,1,2,\ldots
\end{aligned} 
\end{equation}
where $ \eta_t $ is the step size and $\svp$ is the projection operator onto the set of rank-$ \rk $ matrices.  Note that while this set is non-convex, the projection $ \svp $ can be efficiently computed by the rank-$ \rk $ singular value decomposition (SVD). This approach is also known as Singular Value Projection (SVP) or iterative hard thresholding~\cite{jain2010guaranteed}.

\paragraph{Nuclear norm minimization} 

Since the rank function is non-convex, another popular approach for matrix completion is based on replacing the rank with a convex surrogate, namely the nuclear norm. This relaxation leads to the following convex nuclear norm minimization (\NNM) problem~\cite{candes2009exact}: 
\begin{align}\label{mmininuc} 
\mbox{minimize}_{X \in \real^{\dm_1 \times \dm_2}}\; \nucnorm{X} \quad \mbox{subject to }\; \proj(X) = \proj(\tru),
\end{align}
where $ \nucnorm{X} $ denotes the nuclear norm of $ X $, defined as the sum of its singular values. 
\bigskip

Both \PGD and \NNM can be efficiently computed/solved. The key statistical question here is when these two approaches recover the true low-rank matrix $ \tru $ under natural probabilistic models for the observed data~$ \ob $. Perhaps surprisingly, while matrix completion has been  extensively studied, a complete answer to the above question remains elusive. As we elaborate below, existing techniques are fundamentally insufficient in this regard, either relying on assumptions that are difficult (sometimes impossible) to verify, or resulting in performance bounds that are inherently sub-optimal.

\subsection{Our contributions}

Our key insight to the answer of the above question, is as follows. While the \PGD  and \NNM approaches appear completely different, their analysis can both be reduced to studying a \emph{stochastic iterative procedure} of  the form
\begin{equation}\label{p1e1} 
\begin{aligned} 
\iq{t+1} =  \g (\iq{t};\qd), \qquad \text{for } t= 0,1,\dots
\end{aligned}
\end{equation} 
Here $\g(\,\cdot\,; \qd)$ is a possibly nonlinear and implicit mapping with a fixed point $ \iq{*} $, and $\qd $ represents a random data vector. Note that the same $ \qd $ is used in all iterations. Therefore, a major challenge here is that the iterates $ \{\iq{t}\} $ are dependent through the common data~$ \qd $. The \Loo analysis allows us to isolate such dependency and provide fine-grained, entrywise convergence guarantees, namely, bounds on $ \infnorm{\iq{t} - \iq{*}} \defn \max_{i} |\iq{t}_i - \iq{*}_i |. $ As discussed below, these bounds are crucial to the analysis of \PGD and \NNM. We now elaborate.


\subsubsection{Projected gradient descent} 

The \PGD algorithm~\eqref{realsvp} can be recognized as a special case of the iteration~\eqref{p1e1}, where the nonlinear map $ \g $ is given implicitly by the projection~$ \svp $ (i.e., an SVD),  and the random data $ \qd $ corresponds to the observed  indices $ \ob $. A major roadblock in analyzing this iterative procedure involves showing that for all $ t $, the \emph{differences} of iterates $ \iterm{t+1} -\iterm{t}$ remain incoherent, which roughly means that they are entrywise well-bounded. Such bounds are challenging to obtain due to the probabilistic dependency across the iterations.
The seminal work~\cite{jain2010guaranteed} only provides partial results, \emph{imposing as an assumption} that $ \iterm{t+1} -\iterm{t}$ is incoherent. Another set of work~\cite{jain2013low,hardt2014fast,jain2015fast} resorts to a \emph{sample splitting} trick, that is, assuming that a fresh set of \emph{independent} observations $ \ob $ is used in each iteration. As we comment on in greater details in Section~\ref{sec:svp}, this trick is artificial, difficult to implement, and unnecessary in practice; moreover it leads to sample complexity bounds that are either non-rigorous or inherently suboptimal.

Using \loo, we are able to study the \emph{original} form~\eqref{p1e1} of \PGD---without sample splitting---and rigorously prove that the iterates indeed remain entrywise small. In fact, a stronger conclusion is established: the iterates {converge geometrically} in entrywise norm to $ \tru $; see Theorem~\ref{fullmt0} for details.


\subsubsection{Nuclear norm minimization} 

To prove that the convex \NNM program~\eqref{mmininuc}  recovers the underlying matrix $ \tru $ as the optimal solution, it suffices to show that $ \tru $ satisfies the first-order optimality condition, which stipulates the existence of a corresponding \emph{dual optimal solution} (often called a ``dual certificate''). Such a dual certificate can be constructed using an iterative procedure, akin to dual ascent, in the form of the iteration~\eqref{p1e1}. The celebrated Golfing Scheme argument~\citep{gross2011recovering} implements such a procedure, but it crucially relies on the sample splitting trick to circumvent the dependency across iterations. While this argument has proved to be fruitful and led to the best sampling complexity bounds to date~\citep{recht2011simpler,candes2011robust,chen2015incoherence}, it is well-recognized that sample splitting is a workaround and results in fundamentally sub-optimal bounds. 

Using \loo, we are able to analyze the dual ascent procedure with correlated iterations and establish \emph{entrywise} bounds, which ensure dual feasibility. Our results imply that \NNM recovers a $ \dm $-by-$ \dm $ rank-$ \rk $ matrix $ \tru $ given a number of $C \inco \rk \log(\inco\rk) \dm \log \dm $ observed entries, where $ \inco $ is the incoherence parameter of $ \tru $ and $C$ is a universal constant; see Theorem~\ref{mt1} for details. To the best of our knowledge, this is the first sample complexity result, for a tractable algorithm, that has the optimal scaling with the dimension~$\dm$ while at the same time carries absolutely no dependence on the condition number of $ \tru $. We believe that our result paves the way to finally matching the information-theoretic lower bound $C \inco \rk \dm \log \dm $~\citep{candes2010power}. \\

We emphasize that in both settings above, the mapping $ \g(\iq{}; \qd) $ \emph{cannot} satisfy a contraction property over all $ \iq{} $'s, even when restricted to those that are low-rank and incoherent; see Sections~\ref{s2}  and~\ref{lootemplate} for detailed discussion. Therefore, standard techniques from stochastic approximation~\cite{kushner2003SA} are insufficient for our problem. The key step in our analysis is to show that with high probability, a type of contraction is satisfied \emph{by the sequence of iterates $ \{ \iq{t} \}$ generated by the procedure~\eqref{p1e1}}; that is, the iterative procedure avoids the bad regions of $ \theta $ in which contraction fails to hold. The \loo argument plays a key role in establishing this {probabilistic}, {sequence-specific} convergence result.

\subsection{Discussion}

In this paper we focus on the \PGD and \NNM approaches. While algorithms for matrix completion abound, \PGD and \NNM are of fundamental importance.
In particular, \NNM, and more broadly convex relaxation methods, remains one of the most versatile, robust and statistically efficient approaches to high-dimensional statistical problems. Similarly, \PGD plays a unique role in a growing line of  work on non-convex methods. It is recognized as a particularly natural and simple algorithm, does not require a two-step procedure of ``initialization + local refinement", and in fact is often used as an initialization procedure for other algorithms \citep{jain2013low,chen2015fast,sun2016guaranteed,zheng2016convergence}. Moreover, \PGD involves one of the most important numerical procedure: computing the best low-rank approximation using SVD.  Many other algorithms can be either viewed as approximate or perturbed versions of \PGD/SVD, or as computationally efficient procedures for solving \NNM.

In this sense, while we apply \Loo to  \PGD  and \NNM specifically, we believe that this technique is useful more broadly in studying other iterative procedures for statistical problems with complex probabilistic structures. 
\yccomment{say something about your work on regular SDP, but only if you think it fits well here.} 
Indeed, when preparing an early version of this manuscript~\cite{ding2018loo_arxiv}, we became aware of the independent work in~\cite{ma2017implicit,chen2018gradient}, which uses a related technique to analyze other iterative methods for non-convex formulations of matrix completion and phase retrieval problems. We discuss this work in more details in Section~\ref{sec:related}.

Our results can also be viewed as establishing a form of \emph{implicit regularization}. In particular, note that neither the rank-constrained formulation~\eqref{minisvp}, nor the iterative procedures we consider for \PGD and \NNM, has an explicit mechanism for regularizing their solutions to have small entrywise ($ \ell_\infty $) norms. The goal of the \loo analysis is precisely to show that this property is satisfied automatically, with high probability, by the solution sequence generated by the iterative procedures. From this perspective, our results are complementary to a very recent line of work on $ \ell_1 $/$ \ell_2 $  implicit regularization of (stochastic) gradient descent methods~\cite{gunasekar2017implicit,li2018algorithmic}.


\paragraph*{Paper Organization} 

In Section~\ref{sec:related}, we review and compare with existing work in the literature. In Section~\ref{s2}, we present our main results for \PGD and \NNM. In Section \ref{lootemplate},  we outline the main ingredients of our \Loo based technique. Using this technique, we prove our results for \PGD for \NNM in Sections~\ref{fullsvp} and~\ref{sec: prfofthmmt1full}, respectively. 
The proofs of some technical lemmas are deferred to the appendix.
 
\section{Related work and comparison} 
\label{sec:related}

\Loo has many incarnations, and is often used as an \emph{algorithmic} technique, e.g., for cross-validation. As an \emph{analytical} technique, \loo has been employed to study robust M-estimation \citep{el2013robust}, de-biased Lasso estimators \citep{javanmard2015biasing}, and spectral and MLE methods for ranking problems~\cite{chen2017spectral}.

Most related to our work are several recent papers that use \loo in problems involving low-rank matrix estimation. The work in \cite{zhong2017near} studies the generalized power method for phase synchronization problems. The work in \cite{abbe2017entrywise} derives general entrywise perturbation bounds for spectral decomposition. The contemporary work in~\cite{ma2017implicit,chen2018gradient} also studies nonconvex formulations of matrix completion, but focuses on a different algorithm, namely, gradient descent applied to an unconstrained and factorized objective function.

Besides the differences in problem settings, our use of \loo differs from the above work in the following three main aspects: 
\begin{enumerate}
	\item The work in~\cite{abbe2017entrywise,chen2017spectral} consider ``one-shot'' spectral methods, which only involve a single SVD operation. In contrast, the \PGD method we study is an iterative procedure with multiple sequential SVD operations. As will become clear in our proofs, even studying the \emph{second} iteration of \PGD involves a very different analysis than that for one-shot algorithms. In particular, we need to track the propagation of errors and dependency through many (potentially infinite) iterations, which requires careful induction and probabilistic arguments. 
	\item The work in~\cite{ma2017implicit,zhong2017near,chen2018gradient} study gradient descent and power methods, which are iterative procedures in the form of~\eqref{p1e1}. Both methods correspond to an explicit and relatively simple mapping $ \g $. The \PGD algorithm is much more complicated, as it involves computing the SVD, a highly nonlinear operation that is defined implicitly and variationally. The analysis of \PGD is hence significantly harder, requiring quite delicate use of matrix perturbation and concentration bounds.
	\item In our analysis of  \NNM,  \loo is used in a  different context: instead of studying an actual algorithmic procedure, we  use  \loo to study an (unimplementable) iterative procedure that arises in the \emph{dual analysis} of the convex program.
\end{enumerate} 
The exact low-rank matrix completion problem is studied in the seminar work~\cite{candes2009exact}, which initialized the use of the \NNM approach. Follow-up work on \NNM includes~\cite{candes2010power,gross2011recovering,recht2011simpler}, with the best existing sample complexity result given in~\cite{chen2015incoherence}.
The \PGD algorithm is proposed in~\cite{jain2010guaranteed} under the name SVP, although no rigorous guarantees are provided for matrix completion. 
Other iterative algorithms based on non-convex formulations of matrix completion have been proposed; a partial list of work in this line includes~\cite{keshavan2010matrix,hardt2014fast,sun2016guaranteed,chen2015fast,zheng2016convergence}. These algorithms are different from \PGD: they are typically based on a factorized formulation (rather than the rank-constrained formulation~\eqref{minisvp}), and often require explicit $\ell_\infty$ projection/regularization and a careful initialization procedure via SVD. With the $ \ell_\infty $ regularization, a remarkable recent result shows that this factorized formulation in fact has no spurious local minima~\cite{ge2016matrix,ge2017no}, though the resulting iteration complexity bounds therein are quite pessimistic. 
After presenting our main theorem, we provide a more quantitative comparison with the above results.

\PGD and \NNM have been well studied for the related problem of \emph{matrix sensing}~\cite{jain2010guaranteed,recht2010guaranteed}, whose standard formulation can be viewed as a simpler version of matrix completion with the projection $ \proj $ replaced by a linear operator $ \mathcal{A} $ that satisfies certain restricted isometry property (RIP) over \emph{all} low-rank matrices. The lack of such a global RIP/contraction property in matrix completion makes it a more challenging problem and is precisely the reason why \loo is needed.

\section{Problem setup and main theorems}\label{s2} 

In this section, we describe the formal setup of the matrix completion problem and present our main results, namely convergence and sample complexity guarantees for \PGD and \NNM.

\paragraph*{Notations}

For each integer $d>0$, define the set $[d] \defn \{1,2,\dots,d\}$. We write  $f = \bigo (g)$ or $f \lesssim g$ if $ f \le C\cdot g $ for a universal numerical constant $ C $.
Denote by $e_i$  the $ i $-th standard basis vector, $\onevec$ the all-one vector, and $I$ the identity matrix, in appropriate dimensions. 
The set of symmetric matrices in $\real^{\dm\times \dm}$ is denoted by $\SymMat{\dm}$.  For a matrix $Z$, let $Z_{i\cdot}$ be its  $i$-th row and $Z_{\cdot j}$ its $j$-th column. The Frobenius norm, operator norm (maximum singular value) and nuclear norm (sum of singular values) of a matrix are denoted by $\fronorm{\cdot},\opnorm{\cdot}$ and $\nucnorm{\cdot}$, respectively. 
Two other matrix norms are used:  $\infnorm{Z} \defn \max_{i,j}|Z_{ij}|$ for the entrywise $ \ell_\infty $  norm, and   $\twoinfnorm{Z} \defn \max_{i}\twonorm{Z_{i\cdot}}$ for the maximum row $ \ell_2 $ norm.  
The $i$-th largest singular value of a matrix $ Z $ is $\sigma_i(Z)$, and the  best rank-$k$ approximation of $Z$ in Frobenius norm is $\svp(Z)$. For two matrices $A$ and $B$, we write $A\otimes B \defn  AB^T$ for their outer product.  We denote by $\Id$ the identity operator on matrices.  

\subsection{Matrix completion setup}\label{sec:setup}

In matrix completion, the goal is to recover an unknown rank-$ \rk $ matrix $ \tru \in \real^{\dm_1\times \dm_2} $ given partial observations of its entries indexed by the set $ \ob \subset [\dm_1] \times [\dm_2]  $. Define the observation indicator $ \delta_{ij} \defn \indic\{(i,j)  \in \ob \}$, as well as the sampling operator $\proj:\real^{\dm\times \dm} \rightarrow \real^{\dm\times \dm}$ via
\begin{align*}
\big( \proj(Z) \big)_{ij} = Z_{ij} \delta_{ij} = 
\begin{cases}
Z_{ij}, & (i,j) \in \ob, \\
0, & (i,j) \not\in \ob. 
\end{cases}
\end{align*}
It is well known that when most of the entries of $\tru$ equal zero. it is impossible to recover $\tru$ unless all of its entries are observed~\citep{candes2009exact}. To avoid such pathological situations, we impose the standard assumption that $ \tru $ is incoherent in the following sense: 
\begin{dfn}[Incoherence] \label{dfn:inco}
	A matrix $M \in \real ^{\dm_1\times \dm_2}$ with rank-$ \rk $ SVD  $M = U\Sigma V^T$ is $\inco$-incoherent if 
	$$ \twoinfnorm{U}\leq \sqrt{\frac{\inco \rk}{\dm_1}} 
	\qquad \text{and} \qquad 
	\twoinfnorm{V}\leq \sqrt{\frac{\inco \rk}{\dm_2}}.$$
\end{dfn} 

In the sequel, we assume that $ \tru $ has rank $ \rk $ and is $ \inco $-incoherent. If another matrix $ Z \in  \real ^{\dm_1\times \dm_2}$ is $ \bigo(\inco)  $-incoherent, we simply say that $ Z $ is incoherent. 
Denote by $ \kappa := \sigma_1(\tru)/\sigma_\rk (\tru) $ the condition number of $ \tru $. Throughout this paper, by \emph{with high probability} (\whp), we mean with probability at least $ 1-c_1(\dm_1+\dm_2)^{-c_2}$ for some universal constants $c_1,c_2>0$.\footnote{In all our proofs, the value of $c_2$ can be made arbitrarily large as long as the constant $C_0 $ in Theorems~\ref{fullmt0} and~\ref{mt1} is sufficiently large. In this case, if each of a polynomial (in $\dm_1$ and $\dm_2$) number of events holds \whp~(with probability $\ge 1-c_1(\dm_1+\dm_2)^{-c_2}$), then by the union bound the interaction of these events also holds \whp~(with probability $ \ge 1-c_1 (\dm_1+\dm_2)^{-c_2'} $ for a constant $c_2' < c_2$).  }


\subsection{Analysis of projected gradient descent}
\label{sec:svp}

To study the \PGD algorithm~\eqref{realsvp}, we consider a standard probabilistic setting where the observation indices~$ \ob $ are randomly generated.  We focus on the following symmetric and positive semidefinite setting.\footnote{We consider this setting for the sake of a streamlined presentation of our main techniques. Our results can be extended to the general asymmetric case either via a direct analysis, or by using an appropriate form of the standard  dilation argument (see, e.g., \cite{zheng2016convergence}), though the proofs will become more tedious. } 
\begin{mdl}\label{dfn:SMC}
	Under the model \SMC$(\tru, p)$, the matrix $\tru \in \SymMat{\dm}$ is symmetric positive semidefinite, and the observation indices $ \ob $ is such that $\Prob \big( (i,j) \in \ob \big) =p $ independently for all $ i\ge j $, where $ p \in (0,1] $, and that $ (j,i) \in \ob $ if and only if $ (i,j) \in \ob $.
\end{mdl} 



The \PGD algorithm~\eqref{realsvp} is first proposed by Jain et al.\ in~\cite{jain2010guaranteed}. They observe empirically that the objective value $\fronorm{\proj(\iterm{t}-\tru)}^2$ of the \PGD iterate decreases quickly to zero; accordingly,  they conjecture that the iterate $ \iterm{t} $ is guaranteed to converge to $ \tru $ \cite[Conjecture 4.3]{jain2010guaranteed}. Below we reproduce their conjecture, which is rephrased under our symmetric setting:

\begin{conjecture} \label{conjecture} 
	For some numbers $ C, C'>0$ depending on $ \rk $ and $ \inco $, the following holds under the model \SMC$(\tru, p)$. If $p \ge C \frac{\log \dm }{\dm}$, then with high probability, the \PGD algorithm~\eqref{realsvp} with a fixed step size $\eta_t \equiv \eta = \mathcal{O}(\frac{1}{p})$  outputs a matrix $M^t$ of rank at most $\rk$ such that $\fronorm{\proj(M^t)-\proj(\tru)}^2 \leq \epsilon$ after $C'\log \frac{1}{\epsilon}$ iterations; moreover, $ \iterm{t} $ converges to $ \tru $.  
\end{conjecture} 

This conjecture remains open since the proposal of \PGD. 
The original \PGD paper~\cite{jain2010guaranteed} argues that \PGD would converge if the operator $ \proj $ is \emph{assumed} to satisfy a form of Restricted Isometry Property (RIP), i.e., $\proj(\iterm{t} - \tru)  $ preserves the Frobenius norm of the error matrix $\iterm{t} - \tru $. However, RIP \emph{cannot} hold uniformly for all low-rank matrices---just consider matrices with only one non-zero entry. Even when one restricts to \emph{incoherent} iterates $ \iterm{t}$, the error matrix $ \iterm{t} - \tru  $,  being the difference of two incoherent matrices, need not be incoherent itself. 

Instead of relying on RIP and the Frobenius (Euclidean) norm geometry, we directly control the entries of the error matrix using \Loo. In particular, we show that \emph{every entry} of the eigenvectors of $\iterm{t}$ converges to that of $\tru$ simultaneously and geometrically, hence $\iterm{t}$ converges entrywise to $ \tru  $ as well. This result formally proves Conjecture~\ref{conjecture}.

\begin{thm}\label{fullmt0} 
	Under the model \SMC$(\tru,p)$, if $p \geq C_0\frac{\cond^6\inco^4 \rk^6\log \dm}{\dm}$ for some universal constant $C_0$, then with high probability the \PGD algorithm~\eqref{realsvp} with fixed step size $ \eta_t \equiv \frac{1}{p} $ satisfies the bound
	\begin{align}
	\infnorm{M^t-\tru} \leq \lic{t}\ls (\tru), \quad \text{for all }t=1,2,3,\ldots
	\end{align} 
Moreover, the first few iterations satisfy the tighter bound 	$ \infnorm{M^t-\tru} \leq \lic{t} \frac{\ls (\tru)}{\dm}$  for all $t\in[\log\dm]$.
 \yccomment{Added the last sentence.}
\end{thm}
We prove Theorem~\ref{fullmt0} in Section~\ref{fullsvp}, by casting \PGD into a stochastic iterative procedure in the form of~\eqref{p1e1}, namely $ \iq{t+1} =  \g (\iq{t};\qd).$ In particular,  the random data $ \qd $ consists of the observation indicators $ \delta_{ij} := \indic\{(i,j) \in \Omega \} $ generated according to Model~\ref{dfn:SMC}, and the \PGD iteration~\eqref{realsvp} acts on the primal variable $ \iq{t} \equiv \iterm{t} $  with the map $ \g$ given by 
	\begin{align}\label{eq:svd_f_map}
	\g(\cdot \, ; \delta)  = \svp \Bigl( \big(\Id - p^{-1} \proj \big) (\cdot ) + p^{-1} \proj(\tru) \Bigr).
	\end{align}
We establish entrywise geometric convergence of this procedure using the \loo technique; the main ideas of the analysis is outlined in Section~\ref{lootemplate}.

\subsubsection{Discussion and comparison}

Theorem~\ref{fullmt0} establishes, for the first time, the convergence of the \emph{original form} of \PGD. Our result holds when the same set of observed entries $ \Omega $ is used in all iterations, without any sampling splitting. 

In comparison, existing work in~\cite{jain2013low,hardt2014fast,jain2015fast} has considered certain \emph{modified} versions of \PGD. These algorithms are significantly more complicated than the original \PGD: they typically proceed in a stagewise fashion and rely on sampling splitting, i.e., using an independent set of observations $ \ob $  in each iteration. 
It is well recognized that sample splitting has several major drawbacks~\cite{hardt2014fast,sun2016guaranteed}. Firstly, it is a wasteful way of using the data, and leads to sample complexity bounds that grow (unnecessarily) with the number of  iterations. Secondly, the use of sampling splitting is artificial, resulting in algorithmic complications that are not needed in practice. Finally, as observed in \cite{hardt2014fast,sun2016guaranteed},  naive sample splitting (i.e., partitioning $ \ob $ into disjoint subsets) in fact does \emph{not} ensure the required independence; rigorously addressing this technical subtlety (as done in \cite{hardt2014fast})  leads to even more complicated algorithms that are sensitive to the generative model of $\ob$ and hence hardly practical.  

A consequence of Theorem~\ref{fullmt0} is that the \PGD iterates~$\iterm{t} $ remains incoherent throughout the iterations (though incoherence and RIP are no longer needed explicitly in our convergence proof). Note that \PGD, and the nonconvex program~\eqref{minisvp} it aims to solve, have no explicit regularization mechanism to ensure incoherence. Therefore, while natural and simple, the \PGD algorithm is effective for quite delicate probabilistic reasons, which are tied to the specific algorithmic procedure and cannot be simply explained by the geometry of the optimization problem~\eqref{minisvp}. 



\subsection{Analysis of nuclear norm minimization}
\label{sec:nnm}

To present our results on \NNM, we consider the following setting in which the ground-truth $ \tru $ is allowed to be a general rectangular and asymmetric matrix.

\begin{mdl}\label{dfn:MC}
	Under the model \MC$(\tru, p)$, $ \tru $ is a $ \dm_1  $-by-$\dm_2  $ matrix, and the observation indices $ \ob $ is such that  $\Prob \big( (i,j) \in \ob \big) =p $ independently for all $ i, j $, where $ p \in (0,1] $
\end{mdl} 

Starting with the seminar papers~\cite{candes2009exact,keshavan2010matrix}, a long line of work has been devoted to proving sample complexity results for the above model, that is, sufficient conditions for recovering $ \tru $ using \NNM and other algorithms. We summarize the state-of-the-art in Table~\ref{table1},  omitting other existing results that are strictly dominated by those in the table.

Using \Loo, we are able to improve upon this long line of work and establish the following new sample complexity result for \NNM.

\begin{thm}\label{mt1}
	Under the model \MC($\tru,p)$, if $p \geq C_0\inco \rk \log(\inco \rk) \frac{\log (\max\{d_1,d_2\})}{\min\{d_1,d_2\}}$ for some universal constant $C_0$, then with high probability $\tru$ is the unique minimizer of the \NNM program~\eqref{mmininuc}. 
\end{thm} 

We prove this theorem in Section~\ref{sec: prfofthmmt1full}, by connecting \NNM to the stochastic iterative procedure~\eqref{p1e1} in the form $ \iq{t+1} =  \g (\iq{t};\qd).$ 
In particular, we consider an iterative procedure acting on the dual variable $ \iq{t} $ of \NNM, with the map $ \g $ given by 
\begin{align}\label{eq:nnm_f_map}
\g(\cdot \, ; \delta) =  \big(\pj - p^{-1}\pj\proj \big) (\cdot) ;
\end{align}
here $ \pj $ is the projection onto the tangent space at $ \tru $ with respect to the set of low-rank matrices (the explicit expression of $ \pj $ is given in Section~\ref{sec: prfofthmmt1full}), and the data $ \qd $ consists of the observation indicators $ \delta_{ij} := \indic\{(i,j) \in \Omega \} $ under Model~\ref{dfn:MC}. We show that with high probability, the above procedure converges to an optimal dual solution that certifies the primal optimality of $ \tru $ to the \NNM program~\eqref{mmininuc}. A key step is the proof is to show the iterates are dual feasible, which in turn requires bounding their $ \ell_{\infty} $ norm. 
We do so using the \loo technique, with the main ideas of the analysis outlined in Section~\ref{lootemplate}.

\subsubsection{Discussion and comparison}

In the setting with $ \dm_1 = \dm_2 =\dm $, Theorem~\ref{mt1} shows that \NNM recovers $ \tru $ \whp provided that the expected number of observed entries satisfies $ p \dm^2 \gtrsim \inco \rk \log (\inco \rk) \dm \log \dm $. Note that this bound is independent of the condition number $\cond$ of $ \tru $. An information-theoretic lower bound on the sample complexity is established in~\cite{candes2010power}, which shows that $ p \dm \gtrsim \inco \rk \dm \log\dm $ is necessary for \emph{any} algorithm. Our bound hence has the optimal dependence on $ \dm $ and $ \cond $, and is sub-optimal by a logarithmic term of the incoherence parameter $ \inco $ and rank~$ \rk $.

\begin{table}
	\begin{center}
		\renewcommand{\arraystretch}{1.1}
		\begin{tabular}{ r | l }
			\hline
			Work & Sample Complexity $p\dm^2$ \\  
			\hline\hline
			\cite{chen2015incoherence} & $\mathcal{O}(\inco \rk \dm \log^2\dm)$ \\ 
			\cite{keshavan2010matrix} & $ \mathcal{O} \big( \cond^2  \rk \dm \max \bigr\{ \inco \log\dm, \inco^2 \rk \cond^4 \bigr\} \big)$\\
			\cite{sun2016guaranteed} & $\mathcal{O} \big(\cond^2 \rk \dm \max\bigr  \{\inco\log \dm, \inco^2 \rk^6\cond^4\bigr\} \big)$\\
			\cite{zheng2016convergence}& $\mathcal{O}(\cond^2 \inco \rk ^2\dm \max\{\inco, \log \dm\})$\\
			\cite{balcan2018matrix} & $\mathcal{O}(\cond^2 \inco \rk \dm \log \dm \log_{2\cond}\dm)$\\
			\textbf{This Paper} &  $ \boldsymbol{ \mathcal{O}(\inco \rk\log(\inco \rk) \dm \log\dm ) }$ \\
			\hline
			Lower Bound \cite{candes2010power} & $\Omega(\inco \rk \dm \log\dm )$ \\
			\hline
		\end{tabular}
	\end{center}
	\caption{Comparison of sample complexity results for tractable algorithms for exact matrix completion under the setting with $ \dm_1 = \dm_2 =\dm $.} \label{table1}
\end{table}

Let us compare Theorem~\ref{mt1} with the sample complexity results in Table~\ref{table1}. The best existing result for \NNM appears in~\cite{chen2015incoherence}, which establishes a bound that scales sub-optimally with $ \log^2 \dm $. This gap is a fundamental consequence of their proof techniques, as they rely on the Golfing Scheme~\cite{gross2011recovering} that splits $ \Omega $ into $ \log\dm $ disjoint subsets to ensure independence.
All other previous results in the table have non-trivial dependence on the condition number $ \cond $. While it is common to see dependency on $ \cond $ in the \emph{time complexity}, the appearance of $ \cond $ in the \emph{sample complexity} is unnecessary. To the best of our knowledge, our result is the only one that achieves optimal dependence on both the condition number and the dimension for tractable algorithms; in particular, our result is not dominated by any existing results. 

We note that the very recent work in~\cite{balcan2018matrix} obtains a sample complexity result that matches the lower bound; their bound, however, is achieved by an algorithm with running time exponential in~$ \dm $.

\section{\Loo analysis of stochastic iterative procedures}\label{lootemplate} 

As mentioned, we prove our main results for \PGD and \NNM by using \Loo to analyze certain stochastic iterative procedures and obtain entrywise bounds. In this section, we present the main ingredients of this approach. 
We first describe the general ideas of \Loo in the context of the abstract stochastic iteration~\eqref{p1e1}, and then discuss the additional steps needed for the concrete settings of \PGD and \NNM. The complete proofs are given in Sections~\ref{fullsvp} and~\ref{sec: prfofthmmt1full} to follow.


\subsection{Stochastic iterative procedures}

Consider the stochastic iterative procedure in \eqref{p1e1}, namely, $ \iq{t+1} =  \g (\iq{t};\qd).$
Here $\qd = (\qdt{1},\dots ,\qdt{N})\in \real^{N}$ is a random data vector with independent coordinates, and $\g(\cdot \, ; \qd) : \real^N \to \real^N$ is a nonlinear map with a fixed point $\iq{*}$. For simplicity, we assume that $ \iq{*} =0 $.  Our goal is to study the convergence behavior of the iterates $\iq{t}$ to the fixed point $\iq{*}$. 

If $ \g $ is a contraction in $ \ell_2 $ norm in the sense that with high probability,
\begin{align*}
\twonorm{\g (\iq{};\qd) - \g(\iq{\prime};\qd)} &\leq \alpha \twonorm{{\iq{}} -\iq{\prime} }, \qquad \text{uniformly for all }\iq{}, \iq{\prime}
\end{align*}
for some $ \alpha <1 $, then it is straightforward to show that the $ \ell_2 $ distance to the fixed point, $ \twonorm{\iq{t}} $, decreases geometrically to zero. This contraction argument is classical, but often insufficient. 
\begin{itemize}
\item In some cases, one is interested in controlling the \emph{entrywise} behaviors of the iterates, i.e., bounding its $ \ell_\infty $ norm $ \infnorm{\iq{t}} $. Using the worst-case inequality $ \infnorm{\iq{t}} \le \twonorm{\iq{t}} $, together with the above $ \ell_2 $ distance bound, is often far too loose. This is the situation we will encounter in the analysis of \NNM.

\item Worse yet, there are settings where the $ \ell_2 $ contraction does not hold uniformly for all $ \iq{} $ and $ \iq{\prime} $; instead, only a restricted version holds:
\begin{align} \label{restrict_l2}
\twonorm{\g (\iq{};\qd) - \g(\iq{\prime};\qd)} &\leq \alpha \twonorm{\iq{} -\iq{\prime} }, 
\qquad \forall \iq{}, \iq{\prime}: \infnorm{\iq{}}, \infnorm{\iq{\prime}} \le b
\end{align}
for some small number $ b $. In this case, establishing convergence of $ \iq{t} $ requires one to first control the $ \ell_\infty $ norm of $ \iq{t} $. This is the situation we will encounter in  the analysis of \PGD. \medskip
\end{itemize}

In both situations, one needs to control the individual coordinates of the iterates. The \Loo argument allows us to do so by exploiting the independence of the coordinates of the data vector $ \qd $, and by exploiting the fine-grained structures of the map $ \g $. 

For illustration, we assume that iteration \eqref{p1e1} is \emph{separable} w.r.t.\ the data vector $\qd$, in the sense that 
\begin{equation}
\label{separabilitiy}
\iq{t+1}_i = \g _i(\iq{t};\qdt{i}).
\end{equation}
That is, the $i$-th coordinate of the iterate has explicit dependence only on the $i$-th coordinate of the data~$\qd$. Note that $\iq{t+1}_i$ also depends on all coordinates of $\iq{t}$, which in turn depends on the entire vector $\qd$. Consequently, all coordinates of $\iq{t+1}$, for all iterations $t$, are still correlated with each other.

Our crucial observation is that the map $\g (\iq{}; \qd)$ is often not too sensitive to individual coordinates of~$\iq{}$.
In this case, we expect that the randomness of $\qd$ propagates slowly across the coordinates, so the correlation between $\iq{t+1}_i$ and $\{\delta_j,j\not=i\}$ is relatively weak even though they are not independent.
To formalize this insensitivity property, we assume that $ \g $ satisfies, in addition to the restricted $ \ell_2 $-contraction bound~\eqref{restrict_l2}, the following $ \ell_\infty /  \ell_2 $ Lipschitz condition
\begin{align}
\infnorm{\g (\iq{};\qd) - \g(\iq{\prime};\qd)} &\leq \beta \twonorm{\iq{} -\iq{\prime}},
\qquad \forall \iq{}, \iq{\prime}.
\label{infLipschitz}
\end{align}
The value of $ \beta $ is often small, since we are comparing $\ell_\infty$ norm with $\ell_2$ norm.  However, one should expect that $\beta > \frac{1}{\sqrt{N}}$, as otherwise we would have  $\ell_\infty$ contraction/non-expansion, which is what we try to prove in the first place. 

\subsection{\Loo analysis}

We are now ready to describe the \loo argument, which allows us to exploit the properties~\eqref{restrict_l2}--\eqref{infLipschitz} and isolate the behavior of individual coordinates. For each $i\in[N]$, let $\qd^{(-i)} \defn (\qdt{1},\dots, \qdt{i-1},0,\qdt{i+1},\dots, \qdt{N})$ be the vector obtained from the original data vector $\qd$ by zeroing out its $i$-th coordinate. Consider the fictitious iteration (used only in the analysis)
\begin{equation}
\begin{aligned}\label{pe1} 
\iq{0,i} =\iq{0}; \qquad  \iq{t+1,i} = \g (\iq{t,i};\qd^{(-i)}), \quad t=0,1,\ldots .
\end{aligned} 
\end{equation}
Crucially, $\iq{t+1,i}$ is independent of $\qdt{i}$ by construction. Our strategy is to show that the \loo iterates $ \iq{t,i} $ closely approximate the original iterates $ \iq{t} $, thereby leveraging the independence in $ \iq{t,i} $ to bound the coordinates of $ \iq{t} $. 

To this end, we use induction on $ t $, with the hypothesis that $ \twonorm{ \iq {t}-\iq {t,i}} $ (\emph{proximity}) and $ \infnorm{\iq{t}} $ (\emph{$ \ell_\infty $ bound}) are small in an appropriate sense. For the next iteration $ t+1$, it is intuitive that $\twonorm{ \iq {t+1}-\iq {t+1,i}}$ should remain small,  since $\iq{t+1,i}$ and $\iq{t+1}$ are computed using two data vectors different at only one coordinate. More quantitatively,  we can establish the proximity property by
\begin{equation}
\begin{aligned} \label{pkey2} 
\twonorm{\iq{t+1} -\iq{t+1,i}} &= \twonorm{ \g(\iq{t};\qd) -\g (\iq{t,i};\qd^{(-i)})}\\
& \leq \twonorm{\g (\iq{t};\qd)-\g(\iq{t,i};\qd)} + \twonorm{\g(\iq{t,i};\qd)-\g(\iq{t,i};\qd^{(-i)})}\\
& = \underbrace{ \twonorm{ \g(\iq {t};\qd)- \g(\iq {t,i};\qd )} }_{\text{$\ell_2$-Lipschitz term}} + \underbrace{ \big| \g_i (\iq{t,i};\qd_i)- \g_i(\iq{t,i};0) \big| }_{\text{discrepancy term}},
\end{aligned}
\end{equation}
where the last step follows from the separability assumption~\eqref{separabilitiy}.  The discrepancy term above involves two independent quantities $ \iq{t} $ and $ \qd_i $, and can typically be controlled by standard concentration arguments.  To bound the $ \ell_2 $-Lipschitz term above, we invoke the restricted $ \ell_2 $-contraction condition~\eqref{restrict_l2} under the $ \ell_\infty $ induction hypothesis, thus obtaining
$
\twonorm{ \g(\iq {t};\qd)- \g(\iq {t,i};\qd ) }  \le \alpha \twonorm{\iq{t}-\iq{t,i}} .
$
The above bounds combined with the proximity hypothesis on $\twonorm{ \iq {t}-\iq {t,i}}$, yield an (often contracting) upper bound on $\twonorm{\iq{t+1}-\iq{t+1,i}}$, so the proximity bound holds for the next iteration.

Turning to the coordinates of the original iterate $ \iq{t+1} $, we use separability~\eqref{separabilitiy} to compute
\begin{equation} \label{pkey1}
\begin{aligned}
| \iq{t+1}_i |  = | \g_i (\iq{t};\qdt{i}) |
& = | \g_i (\iq {t};\qdt{i})- \g_i (\iq {t,i};\qdt{i}) + \g_i (\iq {t,i};\qdt{i}) |\\
& \leq  \infnorm{ \g(\iq {t};\qd)- \g(\iq {t,i};\qd )} +  | \g_i(\iq {t,i};\qdt{i}) | \\
& \leq \beta \twonorm{ \iq {t}-\iq {t,i}} + | \g_i(\iq {t,i};\qdt{i}) |,
\end{aligned} 
\end{equation}
where the last step follows from the Lipschitz conditions~\eqref{infLipschitz}.  
The first term $ \twonorm{ \iq {t}-\iq {t,i}}  $ above is bounded under the proximity hypothesis; the second term $\g(\iq {t,i};\qdt{i})$ again involves two independent quantities and can be handled as before. Putting together, we have established an upper bound on each coordinate $ | \iq{t+1}_i | $ of the original iteration~\eqref{p1e1}, as desired.\\

To sum up, by using the above arguments, we reduce the challenging problem of controlling the individual coordinates of $\iq{t}$ to two easier tasks:
\begin{enumerate}
	\item Control the quantity $\g_i (\iq{}; \qdt{i})$ when $\iq{}$ and~$\qdt{i}$ are independent. This quantity measures the sensitivity of $ \g $ under an independent random perturbation to one coordinate of the data vector $ \qd $.
	\item Control the quantity $\g(\iq{}; \qd) -\g(\iq{\prime};\qd)$ in various norms when $ \iq{},\iq{\prime} $ are small entrywise. This quantity measures the sensitively of $ \g $ with respect to the iterate $ \iq{} $. This task can be accomplished using the (restricted) Lipschitz properties~\eqref{restrict_l2} and~\eqref{infLipschitz} of $\g$, which can often be established even when $ \ell_2 $- or $ \ell_\infty $-contraction fails to hold uniformly.
\end{enumerate}

\subsection{Analysis of \PGD and \NNM using \Loo} 

To study \PGD and \NNM,  we instantiate the abstract procedure~\eqref{p1e1} as in equations~\eqref{eq:svd_f_map} and~\eqref{eq:nnm_f_map}, respectively. 
The analysis of these two procedures follows the general strategy outlined above, though the proof involves several technical complications:
\begin{itemize}
	\item In the matrix completion setting, the iterates $ \iq{t} $ and the random data $ \qd{} $ are both matrices, so the  separability property~\eqref{separabilitiy}, and accordingly the \loo sequences $ \{\iq{t,i}\} $, take a more complicated form involving the rows and columns of these matrices. 
	\item Consequently, in addition to bounding the entrywise norm of the iterates, we often need to bound their row-wise norm as well. In the case of \PGD, we in fact do so for the eigenvectors of the iterates.
	\item We need to establish the Lipschitz properties~\eqref{restrict_l2} and~\eqref{infLipschitz} with small enough $ \alpha $ and $\beta$, which requires the use of appropriate concentration and matrix perturbation bounds. 
\end{itemize}

In addition, \PGD involves an unbounded number of iterations, yet the high probability bounds obtained by \loo are only valid for $ \text{poly}(\dm) $ iterations, due to the use of union bounds. Fortunately, after this many iterations, \PGD already enters a small neighborhood of the fixed point $\iq{*}$, within which one can establish certain uniform concentration bounds that are valid for an arbitrary number of iterations.

For \NNM, a direct application of \loo as in the last subsection would establish a sample complexity result of the form $p \gtrsim \text{poly} (\inco \rk) \frac{\log \dm}{\dm}$.\footnote{This is done in an earlier version of this paper~\cite{ding2018loo_arxiv}.} 
This bound has the right dependence on~$ \dm $, but it is vastly sub-optimal in terms $ \inco $ and $ \rk $. To remove these superfluous $ \inco \rk $ factors and establish the tighter bound $p \gtrsim  \inco \rk \log(\inco \rk) \frac{\log \dm}{\dm}$ in Theorem~\ref{mt1}, we take a hybrid approach that ``warm-starts'' the iterative procedure by running a small number (in particular, $ \bigo(\log\inco \rk) $) of iterations with sample splitting. As mentioned in Section~\ref{sec:nnm}, this approach gives the best sample complexity upper bound to date, but we have not been able to remove the extra $ \log\inco \rk$ factor that is absent in the information-theoretic lower bound (cf.\ Table~\ref{table1}).

\renewcommand{\hpp}[1]{\mathcal{H}_\ob^{(-#1) }}

\section{Proof of Theorem~\ref{fullmt0}} \label{fullsvp} 

In this section, we prove our convergence guarantee for \PGD in Theorem~\ref{fullmt0}. The proof makes use of the auxiliary lemmas given in Appendix~\ref{auxillary lemmas}.
Let $\ls$ denote the largest eigenvalue of $\tru$ and $\sss$ the smallest nonzero eigenvalue. Recall that $\kappa := \frac{\ls}{\sss}$ is the condition number of $\tru$.

\paragraph*{Proof outline} After recording some preliminary steps in~Section~\ref{sec:svp_proof_prelim},  we present the main steps of the proof in two parts, following the strategy given in Section~\ref{lootemplate}.  Let $ t_0 \defn \tot $.
\begin{itemize}
	\item Part 1: We prove that \whp there holds the infinity norm bound
	\begin{align} \label{fullind1} 
	\infnorm{ \iterm{t} - \tru} \leq \frac{ 1}{d}\lic{t} \ls \qquad \text{for all $t=1,2, \ldots t_0 .$}
	\end{align}
	\yccomment{Changed $ \frac{\rk}{\dm} $ in the above bound to $ \frac{1}{\dm} $, in view of~\eqref{eqn:iterm-truminfnorm}. This tighter bound seems necessary for the final arguments in Section~\ref{fullsubsection5}. Please double check.}
	 This bound is proved using an induction argument and the \loo technique. The proof proceeds in two sub-steps.
	 \begin{itemize}
	 	\item Part 1(a): We first establish the base case, that is, the bound \eqref{fullind1} holds for $ t=1 $. This step is itself non-trivial, and is presented in Section~\ref{fullsubsection1}.
	 	\item Part 1(b): We next perform the induction step, in which we assume that  the induction hypothesis holds for $ t $ and show that it is also valid for $ t+1 $.  This step is presented in Section~\ref{fullsubsection2}.
	 \end{itemize} 
	\item  Part 2: We show in Section \ref{fullsubsection5} that \whp  there holds the Frobenius norm bound
	\begin{align}\label{fullind2} 
	\fronorm{\iterm{t} -\tru} \leq \lic{t-t_0} \fronorm{\iterm{t_0} - \tru} \qquad \text{	for all $t \geq t_0,$}
	\end{align}  
	thereby controlling the error of \PGD for an infinite number of iterations.
\end{itemize}
Combining the above two bounds \eqref{fullind1} and \eqref{fullind2}, we conclude that  \whp
$
\infnorm{M^t -\tru} \leq \lic{t}\ls \text{ for all $t\geq 1,$}
$ 
which establishes the first part of Theorem~\ref{fullmt0}. The second part of the theorem is exactly the bound~\eqref{fullind1}.

\subsection{Preliminaries} \label{sec:svp_proof_prelim}

Throughout the proof, we use $c$ and $C$ to denote sufficiently large universal constants that may differ from line to line. Recall the assumption $p \ge C \frac{\cond^6\inco^4 \rk^6\log \dm}{\dm}$.


Recall that a constant step size  $\eta_t \equiv \frac{1}{p}$ is used in the \PGD iteration~\eqref{realsvp}; accordingly, we define the operator $\hp \defn \Id-\frac{1}{p} \proj$. With this notation, the \PGD iteration~\eqref{realsvp} can be written compactly as
\begin{equation}
\begin{aligned} \label{eq: originalsvp}
\iterm{0} = 0;  \quad \iterm{t+1}= \svp(\tru +\hp(\iterm{t}-\tru))  \qquad t=0,1,\ldots 
\end{aligned} 
\end{equation} 
We write the rank-$ \rk $ eigenvalue decompositions of $\iterm{t}$ and $M^*$ as 
$\iterm{t} = (\iter{t}\iterl{t}) \otimes \iter{t} $ 
and $M^* = (F^*\Lambda^*) \otimes F^*$ respectively. Both $\iter{t}$ and $F^*$ are in $\real^{\dm \times \rk}$
and have orthonormal columns. The matrices $\iterl{t}$ and $\Lambda^*$ are in $\real^{r\times r}$ and are diagonal 
matrices. Note that $\Lambda^* = \diag(\sigma_1,\dots,\sigma_r)$.

For the purpose of analysis, we consider a  \loo version of \PGD. Let $\hpp{m}$ be the operator derived from $\hp$  with the $m$-th row and column observed; that is, 
$$ \biggr (\hpp{m} Z \biggr )_{ij} = 
\begin{cases}
 (1- \frac{1}{p}\delta_{ij} )Z_{ij},  & i\not =m, j\not =m, \\ 
 0, &  i = m \;\text {or}\; j = m  .
 \end{cases}$$ 
For each $m\in [d]$, define the following \loo sequence:
\begin{equation}
\begin{aligned} \label{eq: loosequence}
\iterm{0,m} = 0;  \quad \iterm{t+1,m}  = \svp\bigr [ \tru + \hpp{m} \bigr ( \iterm{t,m}  - \tru \bigr ) \bigr], \qquad t=0,1,\ldots 
\end{aligned} 
\end{equation}
We write the rank-$ \rk $ eigenvalue decomposition of $ \iterm{t,m} $ as $\iterm{t,m} = (\iter{t,m}\iterl{t,m}) \otimes \iter{t,m}$. Here $\iter{t,m}\in \real^{\dm \times \rk}$ has orthonormal columns and $\iterl{t,m}\in \real^{\rk \times \rk}$ is diagonal.  By construction, the sequence $(\iiter{t}{m}, \iterl{t,m})_{t=0,1,\ldots }$ is independent of $\delta_{m \cdot}$ and $\delta_{\cdot m}$, i.e., the $m$-th row and column of $\ob$. It is convenient to let $ m=0 $ correspond the original \PGD iteration, e.g.,  $\iter{t,0} \equiv \iter{t}$ and $ \hpp{0} \equiv \hp$. 
\yccomment{Throughout the section, I changed $ \mathcal{H}_\Omega^{(i)} $ to $ \mathcal{H}_\Omega^{(-i)} $ so as to be consistent with the NNM proof. This is done by redefining a macro. Please double check for consistency.}

A few notations are needed for measuring the distance between the column spaces of two matrices $ \iter{} ,  {\iter{}}' \in \real^{\dm \times \rk} $. For each $m\in \{0\}\cup [\dm]$, set $\Hp{t,m} \defn  (\truf)^T  \iter{t,m}$ and its rank-$ \rk $ SVD be $\Hp{t,m} = \bar{U}\bar{\Sigma}\bar{ V}^T$. It is known that the orthogonal matrix $\G{t,m} \defn \bar{U}\bar{V}^T$  is the minimizer of the problem $\min_{O\in \real^{\rk \times \rk}: OO^T = I}\fronorm{\iter{t,m}-\truf O}$ \citep[Lemma 6]{ge2017no}. Similarly, for each pair of \loo iterates $\iter{t,i}$ and $\iter{t,m}$, we set $\Hp{t,i,m} \defn ( \iter{t,m} )^T\iter{t,i}$ and define the orthogonal matrix $\G{t,i,m}$ accordingly. We again use the convention that  $\Hp{t} \equiv \Hp{t,0}, \G{t} \equiv  \G{t,0}$.

The following notations are defined for each step $ t =0,1,\ldots $ Denote the residual matrix  of the original \PGD \eqref{eq: originalsvp}
by $\er{t} \equiv \err{t}{0} \defn \hp ( \iterm{t} - \tru ) $, 
and the residual matrix of the $m$-th \loo sequence~\eqref{eq: loosequence} by 
$\er{t,m} \defn \hpp{m} ( \iterm{t,m} - \tru ) $. 
For each $i,m \in\{0\}\cup [\dm]$, denote the difference of the iterates from the true $\truf$ by $\ddf{t,m} := \iter{t,m}-\truf\G{t,m}$, the distance between a pair of iterates by $D^{t,i,m} \defn \iter {t,i} - \iter{t,m}\G{t,i,m}$, and the non-commutativity measure matrix by $S^{t,i,m}:= \iterl{t,m}\G{t,i,m}- \G{t,i,m}\iterl{t,i}$. Finally, define the shorthands $\ddf{t,\infty} := \arg\max _{\ddf{t,m}:0\leq m\leq \dm} \twoinfnorm{\ddf{t,m}}$, $D^{t,\infty} :=\arg\max_{D^{t,i,m}:0\leq i,m\leq \dm }\fronorm{D^{t.i,m}}$, $E^{t,\infty} :=\arg\max_{E^{t,i}:0\leq i\leq m}\opnorm{E^{t,i}}$ and $S^{t,0,\infty} := \arg \max_{S^{t,0,m}:0\leq m\leq \dm}\fronorm{ S^{t,0,m}}$.\\

\paragraph*{Unequal eigenvalues and non-commutativity}

Since the eigenvalues of $ \iterm{t,m}$ are unequal in general, the proof is complicated by the fact the the diagonal eigenvalue matrix $ \iterl{t,m} $ does not commute with the matrices $\Hp{t,i,m}$ and $G^{t,i,m}$. We record two technical lemmas for handling this issue. 
The first lemma is proved in Section~\ref{sec:proof_clemma1} using techniques from~\cite{abbe2017entrywise,fan2017distributed}.

\begin{lem}\label{clemma1}
	Suppose that $W \defn \tru + E$, where $E \in \SymMat{\dm}$. Let $F\in \real^{\dm \times \rk}$ be the matrix whose columns are the top-$\rk$ eigenvectors of $W$. Let the SVD of the matrix $H \defn (\truf)^T F$  be $H = U\Sigma V^T$. Let $G \defn UV^T$. If  $\opnorm{E}<\frac{1}{2}\sss$, then we have the bounds
	\begin{align*}
	 \opnorm{\true G- G\true} &\leq \left( 2 + 2\ls \frac{1}{\sss-\opnorm{E}} \right)\opnorm{E},\\
	\opnorm{\true H- G\true} &\leq \left( 2+ \ls\frac{1}{\sss-\opnorm{E}} \right)\opnorm{E},\\
	\opnorm{\true G- H\true} &\leq \left( 2+ \ls \frac{1}{\sss-\opnorm{E}} \right)\opnorm{E}.
	\end{align*} 
\end{lem}

The next lemma, proved in Section~\ref{sec:proof_clemma2}, is useful for controlling $\fronorm{ S^{t,i,m}}$. Recall that $\lambda_i(A)$ denotes the $i$-th largest eigenvalue of a symmetric matrix $A$.
\begin{lem}\label{clemma2} 
		Suppose that  $A \in \SymMat{\dm}$ has eigen decomposition $A = F_1 \Lambda _1 F_1^T + F_2 \Lambda_2 F_2^T$, where $\Lambda_1$ is the diagonal matrix consisting of the top-$\rk$ eigenvalues of $A$. Suppose that  $\tilde{A} = A+E$ with $ E\in \SymMat{\dm}$, and similarly let $ \tilde{\Lambda} $ and $ \tilde{F} $ be the matrices of the top-$\rk$ eigenvalues and eigenvectors of $ \tilde{A} $, respectively.  Suppose that the top-$r$ eigenvalues of $A$ are positive, and the smallest positive eigenvalue is larger in absolute value than the negative ones.  Let $ H \defn F_1^T \tilde{F} $ has SVD $ U\Sigma V^T$, and $G \defn UV^T$.  If
		$\opnorm{E}<\frac{1}{2}( \lambda_{r}(A) -|\lambda_{r+1}(A)|)$, then 
		 \begin{align*}
		 \matsnorm{ \Lambda_1 G - G\tilde{\Lambda} }{} 
		 & \leq \Bigr(\frac{ 2\lambda_{1}(A)+\opnorm{E}}{\lambda_{r}(A) -\opnorm{E}}+1\Bigr) \matsnorm{E\tilde{F}}{} ,
		 \end{align*}
		 where the norm $\matsnorm{\cdot}{}$ can be either $\fronorm{\cdot}$ or $\opnorm{\cdot}$,
\end{lem}

\subsection{Part 1(a): Induction hypothesis and base case $t=1$} \label{fullsubsection1} 

Following our proof outline, we first establish the $ \ell_\infty $ bound~\eqref{fullind1} for $1\le t\le t_0 \defn \tot$ by induction on $t$.  Our induction hypothesis is that \whp,
 \begin{subequations} 
 \label{eq: inductionsvphyothesis}
 \begin{align}
  (\text{operator norm bound}) & \qquad \opnorm{E^{{t-1},\infty }} \leq \fulladd \lic{t} \sss, 
  \label{eq:hypothesis_opnorm}\\
 (l_{2,\infty} \text{ norm bound}) & \qquad \twoinfnorm{\ddf{t,\infty} } \leq \fulladd \lic{t} \inc,
 \label{eq:hypothesis_2inf}\\
 (\text{proximity}) &\qquad  \fronorm{D^{t,\infty}} \leq \fulladd \lic{t}\inc, \label{eq:hypothesis_proximity}\\
 (\text{non-commutativity bound}) & \qquad \fronorm{S^{t,0,\infty}} \leq \ls  \fulladd \lic{t}\inc.  \label{eq:hypothesis_commute}
 \end{align}
 \end{subequations}
By applying Lemma \ref{lemma: inftwoinfnorm}, we see that the above bounds in~\eqref{eq: inductionsvphyothesis} immediately imply the desired bound~\eqref{fullind1} on the original iterate $\iterm{t}$. 
 
We first prove the base case $t=1$ of the induction hypothesis \eqref{eq: inductionsvphyothesis}. In the proof we shall show that various inequalities hold \whp for each fixed indices $i\in \{0\}\cup [\dm]$ and $m\in [\dm]$. By the union bound, these inequalities  hold simultaneously for all indices \whp

\subsubsection{Operator norm bound} 
 
We have \whp
\begin{equation}
\begin{aligned}\label{fullopnorm0} 
\opnorm{\er{0,i}}  = \opnorm{ \hpp{i} \bigr (\tru\bigr)}
 \niea 2c \sqrt{ \frac{\dm \log \dm}{p}}\frac{\inco \rk }{\dm}\ls 
 \nieb \frac{1}{C\cond} \fulladd\lic{}\sss,
\end{aligned}
\end{equation} 
where step $\nia$  is due to Lemma \ref{candes0} and $\infnorm{\tru}\leq \incc\ls$, and step $\nib$ is due to  $p \gtrsim \frac{\cond^6\inco^4\rk^6\log d}{d}$. The maximum of the last LHS over $ i  \in \{0\} \cup [\dm]$ is $ \opnorm{E^{{0},\infty }} $. Taking this maximum proves the desired operator norm bound~\eqref{eq:hypothesis_opnorm} in the induction hypothesis for $ t=1 $. 

\subsubsection{$ \ell_{2,\infty} $, proximity and non-commutativity bounds}

We claim that the following three intermediate inequalities hold \whp for all $i,m$:
\begin{subequations} \label{eq:svp_base_intermediate}
\begin{align}
 \twonorm{ \ddf{1,i}_{m\cdot}}
&\leq   \frac{1}{4} \fulladd \lic{}  \inc  + \frac{2}{15} \twoinfnorm{\ddf{1,m}} + \frac{2}{15}  \fronorm{D^{1,i,m}}, 
\label{eq: boundofDeltainitialstep} \\
\fronorm{D^{1,i,m}} 
&\leq  \frac{1}{8} \twoinfnorm{\ddf{1,\infty}} +\frac{1}{8}\fullindb{}, \label{eq: boundofDinitalstep} \\
\fronorm{\big[ (\hpp{0}-\hpp{m})(\tru) \big] \iter{1,m}} 
&\leq  \sss \biggr[\frac{1}{32}\twoinfnorm{\ddf{1,m}} +  \frac{1}{32}\lic{}\fulladd \sqrt{\frac{\inco \rk}{\dm}}\biggr]. 
\label{eq: boundofWinitialstep}
\end{align}
\end{subequations} 
We postpone the proofs of~\eqref{eq: boundofDeltainitialstep}, \eqref{eq: boundofDinitalstep} and~\eqref{eq: boundofWinitialstep} to Sections~\ref{sec: prelimstep0fullsvp}, \ref{sec:proof_boundofDinitialstep} and~\ref{sec:proof_boundofWinitialstep}, respectively. With these three inequalities, the last three bounds in the induction hypothesis follow easily, as we show below.

First, plugging~\eqref{eq: boundofDinitalstep} 
into~\eqref{eq: boundofDeltainitialstep}, we obtain that \whp
\begin{equation}
\begin{aligned} 
\twonorm{\ddf{1,i}_{m\cdot}} 
&\leq  \frac{1}{2}\fullindb{} + \frac{1}{6}\twoinfnorm{\ddf{1,\infty}} . 
\end{aligned}
\end{equation}
The maximum of the last LHS over $i\in \{0\}\cup [\dm]$ and $m\in [\dm]$ is $ \twoinfnorm{ \ddf{1,\infty}} $. Taking this maximum 
and rearranging terms, we obtain the desired $\ell_{2,\infty}$ bound~\eqref{eq:hypothesis_2inf} in the induction hypothesis for $ t=1 $.

Next, plugging the $\ell_{2,\infty}$ bound~\eqref{eq:hypothesis_2inf} we just proved into~\eqref{eq: boundofDinitalstep}, we obtain that \whp
\begin{equation*}
\begin{aligned}
\fronorm{D^{1,i,m}} 
&\leq \frac{1}{8} \twoinfnorm{\ddf{1,\infty}} + \frac{1}{8}\fullindb{}
\leq \frac{1}{4}\fullindb{}.
\end{aligned}
\end{equation*}
The maximum of the last LHS over $i$ and $m$ is $ \fronorm{D^{1,\infty} } $. 
Taking this maximum proves the desired proximity bound~\eqref{eq:hypothesis_proximity} in the induction hypothesis for $ t=1 $.

Finally, we have \whp
\begin{align*}
\fronorm{S^{1,0,m}}  
=\fronorm{S^{1,m,0}} 
&\niea 4\cond \cdot \fronorm{\big[ (\hpp{0}-\hpp{m})(\tru) \big] \iter{1,m}} \\
&\nieb \frac{\ls}{2}\fullindb{},
\end{align*}
where step $ \nia $ follows from Lemma \ref{clemma2} whose premise is satisfied because of the bound~\eqref{fullopnorm0}, 
\yccomment{May want to add more explanation for step $ \nia $ here}
and step $ \nib $ follows from the inequalities~\eqref{eq: boundofWinitialstep} and~\eqref{eq:hypothesis_2inf}.
The maximum of the last LHS over $m$ is $ \fronorm{S^{t,0,\infty}}  $. Taking this maximum proves the desired non-commutativity bound~\eqref{eq:hypothesis_commute} in the induction hypothesis for $ t=1 $.
We have thus completed the proof of the base case of the hypothesis.

\subsubsection{Proof of Intermediate Inequality~\eqref{eq: boundofDeltainitialstep}}
\label{sec: prelimstep0fullsvp}

We focus on the $m$-th row of the difference matrix $\ddf{1,i} \defn \iter{1,i}- \truf \G{1,i}$. Using the the fact $\iter{1,i}\Lambda^{1,i}=(\tru +\er{0,i})\iter{1,i}$,  we have the expression
 \begin{align*}
 \ddf{1,i}_{m\cdot} =\iter{1,i}_{m\cdot}- \truf_{m\cdot }\G{1,i} = e_m^T (\tru +\er{0,i}) \iter{1,i} ( \Lambda ^{1,i}) ^{-1}-e_m^T \truf \G{1,i}.
 \end{align*}  
Rearranging the last RHS yields
\begin{equation*}
\begin{aligned} 
& \ddf{1,i}_{m\cdot} \\
=&  e_m^T \truf \true \bigr [ (\truf)^T\iter{1,i}( \Lambda ^{1,i}) ^{-1}-(\true)^{-1}\G{1,i}\bigr]+e_m^T \er{0,i} \iter{1,i} ( \Lambda ^{1,i}) ^{-1}\\
=&  \underbrace{e_m^T \truf\true\bigr [ (\truf)^T\iter{1,i}(\true)^{-1}-(\true)^{-1}\G{1,i}\bigr]}_{T_1}
+ \underbrace{e_m^T \truf \true (\truf)^T\iter{1,i}\bigr[( \Lambda ^{1,i}) ^{-1}-(\true)^{-1}\bigr]}_{T_2}+\underbrace{e_m^T \er{0,i} \iter{1,i} ( \Lambda ^{1,i}) ^{-1}}_{T_3}.
\end{aligned} 
\end{equation*}
Below we bound each of the three terms $T_1,T_2$ and $T_3$.

 \paragraph{Bounding $T_1$} 
 
 \if
By Davis-Kahan $ \sin\Theta $ Theorem (Lemma~\ref{daviskahan}), we have 
 \begin{align}
 \label{boundofsinthetafirststepfullsvp}
 \opnorm{\sin \Theta }\leq \frac{\opnorm{\er{0,i}}}{1- \opnorm{\er{0,i}}}.
 \end{align}
\yccomment{Why do we need the above discussion and bound?} \else\fi

 Note that $ T_1 =  e_m^T\truf R$, where 
 \[
 R \defn  \true\bigr [ (\truf)^T\iter{1,i}(\true)^{-1}-(\true)^{-1}\G{1,i}\bigr]
 =\bigr [ \true \Hp{1,i}-\G{1,i}\true\bigr](\true)^{-1}.
 \]
 Recall that by definition $\Hp{1,i}\defn (\truf )^T\iter{1,i} $ has SVD
$ \bar{U} \bar{\Sigma} \bar{V}^T$ and $\G{1,i} \defn \bar{U} \bar{V}^T$. Therefore,  Lemma \ref{clemma1} is applicable, which gives that \whp
\[
\opnorm{R}
\leq \left( 2 + 2\ls \frac{1}{\sss-\opnorm{\er{0,i}}} \right)\opnorm{\er{0,i}} \cdot  \frac{1}{\sss} 
\leq \frac{1}{C} \fulladd\lic{},
\] 
where the last step is due to the bound~\eqref{fullopnorm0} on $\opnorm{\er{0,i}}$.
Thus $T_1$ can be bounded \whp as
 \begin{align*}
 \twonorm{T_1} & = \twonorm{e_m^T\truf R}\leq \twonorm{e_m^T\truf}\opnorm{R}\leq  \frac{1}{C} \fulladd\lic{} \inc.
 \end{align*} 
 
\paragraph{Bounding $T_2$} 
\yccomment{Below I changed all $\Lambda^{0,i}$ to $\Lambda^{1,i}$; please double check.}
Using Weyl's inequality $ \opnorm{\Lambda^{1,i}-\true} \leq \opnorm{\er{0,i}} $ and the bound~\eqref{fullopnorm0} on $\opnorm{\er{0,i}}$, we have \whp  
\begin{align}
 \label{eq:Lambdainv}
\opnorm{(\Lambda ^{1,i})^{-1}-(\true)^{-1}}\leq \frac{1}{C\cond\sss} \fulladd\lic{}.
 \end{align}
 It follows that \whp  
 \[
 \twonorm{T_2} 
 \leq  \twonorm{e_m^T \truf } \cdot \ls \cdot \opnorm{(\truf)^T\iter{1,i}} \cdot \opnorm{( \Lambda ^{1,i}) ^{-1}-(\true)^{-1}}
 \leq \frac{1}{C} \fulladd\lic{}\inc .
 \] 
 
\paragraph{Bounding $T_3$} 
Using the above bound~\eqref{eq:Lambdainv}, we have \whp
 \begin{align*}
 \twonorm{e_m^T \er{0,i} \iter{1,i} ( \Lambda ^{1,i})^{-1}}  
 \leq \twonorm{e_m^T \er{0,i} \iter{1,i}} \frac{1}{\sss-\frac{\sss}{C\cond\sss} \fulladd\lic{}}.
 \end{align*} 
  
Combining the above bounds for $T_1, T_2$, and $T_3$, we obtain that \whp
 \begin{equation}
 \begin{aligned}\label{fullkey0}
  \twonorm{ \ddf{1,i}_{m\cdot}}
  \leq   \frac{1}{C} \fulladd \sqrt{\frac{\inco \rk}{\dm}} \lic{} 
  +  \twonorm{e_m^T \er{0,i} \iter{1,i}} \frac{C+1}{C\sss} .
 \end{aligned}
 \end{equation}
To proceed, we further control $\twonorm{e_m^T \er{0,i} \iter{1,i}}$. 
If $i=m$, then $e_m^T\er{0,i}=0$ and we are done. Below we assume $i\not=m$.
Let $Q\in \real^{r\times r}$ be an orthogonal matrix whose value is to be determined later. We write
\begin{equation}\label{fullef0}
 \begin{aligned}
 e_m^T \er{0,i} \iter{1,i}
 &= e_m^T \er{0,i} \iter{1,m}Q + e_m^T\er{0,i}(\iter{1,i}-\iter{1,m}Q)\\ 
 &=\underbrace{  e_m^T\hpp{i}(-\tru)\iter{1,m}Q}_{\tilde{T}_1} + \underbrace{ e_m^T\er{0,i}(\iter{1,i}-\iter{1,m}Q)}_{\tilde{T}_2}.
 \end{aligned}
\end{equation}
We bound $\tilde{T}_1$ and $\tilde{T}_2$ below. 
 
\paragraph{Bounding $\tilde{T}_1$} 
We have the bound
 \begin{equation}
 \begin{aligned}\label{fulls1t100}
 \twonorm{\tilde{T}_1} 
 & = \twonorm{ e_m^T\hpp{i}(-\tru)\iter{1,m}}
 \leq \sqrt{\rk}\max_{j \in [\rk]}  \biggr|\sum_{k=1}^d \Big( 1-\frac{1}{p}\delta_{mk} \Big)(-\tru_{mk})\iter{1,m}_{kj}\biggr|.
 \end{aligned}
 \end{equation} 
Note that $\iter{1,m}$ is independent of $\{\delta_{mk},k\in[\dm]\}$ by construction. Therefore,  Bernstein's inequality (Lemma~\ref{bern})
ensures that for each $j \in [\rk]$, with probability at least $1-\dm^{-12}$, there holds the inequality
 \begin{align}\label{fulls1t100_a}
 &\bigg| \sum_{k=1}^d \Big(1-\frac{1}{p}\delta_{mk}\Big)(-\tru_{mk})\iter{1,m}_{kj} \bigg|  \leq   \underbrace{\sqrt{\frac{C\log d}{p}} \twoinfnorm{\iter{1,m}} \twonorm{\tru_{m\cdot}}}_{\tilde{T}_{1a}}
 + \underbrace{ \frac{C\log\dm}{p} \twoinfnorm{\iter{1,m}}\infnorm{\tru}}_{\tilde{T}_{1b}},
 \end{align}
For the first term $\tilde{T}_{1a}$, we have
\begin{align*}
\tilde{T}_{1a} 
 & \niea \sqrt{\frac{C\log d}{p}} \Big(\twoinfnorm{\ddf{1,m}} + \sqrt{\frac{\inco \rk}{\dm}} \Big) \ls \inc \\
 &\nieb \sss \biggr[ \frac{1}{16\sqrt{\rk}}\twoinfnorm{\ddf{1,m}} + \frac{1}{16\times 2\sqrt{\rk}}\fulladd\inc  \biggr].
\end{align*}
where step $\nia$ is due to the facts that $\twoinfnorm{\iter{1,m}} \le \twoinfnorm{\ddf{1,m}} + \twoinfnorm{\truf} $, that $ \twonorm{\tru_{m\cdot}} \le \twonorm{\truf_{m\cdot }} \cdot \ls \cdot \opnorm{\truf} $ and that $ \twonorm{\truf_{m\cdot}}  \le \twoinfnorm{\truf}\leq \inc$, and step $ \nib $ is due to the assumption $p \gtrsim  \frac{\cond^4\log (d) \inco^4 \rk^6}{\dm}$.
For the second term  $\tilde{T}_{1b}$, we follow a similar argument as above to obtain that \whp
 \begin{align*} 
\tilde{T}_{1b} 
 & \leq \sss  \biggr[ \frac{1}{16\sqrt{\rk}}\twoinfnorm{\ddf{1,m}}+\frac{1}{16\times 2\sqrt{\rk}} \fulladd \inc \biggr].
 \end{align*}
We plug the bounds for $\tilde{T}_{1a}$ and $\tilde{T}_{1b}$ into the inequality~\eqref{fulls1t100_a}, and take a union bound over all $ j\in[\rk] $ and $ m\in[\dm] $. Combining with the inequality~\eqref{fulls1t100}, we obtain that \whp
 \begin{equation}\label{fulls1t10}
 \begin{aligned}
 \twonorm{ \tilde{T}_1 } 
 & \leq \sss \biggr[ \frac{1}{8}\twoinfnorm{\ddf{1,m}}+\frac{1}{8\times 2}\fulladd \inc \biggr].
 \end{aligned}
 \end{equation}

\paragraph{Bounding $\tilde{T}_2$}  
Recalling the definition of $\G{t,i,m}$ in Section~\ref{sec:svp_proof_prelim}, we choose $Q=\G{1,i,m}$ so that $\iter{1,i}-\iter{1,m}Q = D^{1,i,m}$. We thus have
 $ \tilde{T}_2 
 = e^T\hpp{i}(\tru) D^{1,i,m} $
by definition of $ \tilde{T}_2 $.  It follows that \whp
 \begin{equation*}
 \begin{aligned}
 \twonorm{\tilde{T}_2}
 \niea  \opnorm{\hpp{i}(\tru)} \fronorm{D^{1,i,m}}
 \nieb 2c \sqrt{\frac{d\log d}{p}}\infnorm{\tru}\fronorm{D^{1,i,m}},
 \end{aligned}
 \end{equation*}
 where step $\nia$ is due to Cauchy-Schwarz, and step $ \nib $ is due to Lemma \ref{candes0}. Using the assumptions that $\infnorm{\tru}\leq \incc$ as $\tru$ is $\inco$-incoherent and $p\gtrsim\frac{\cond^2\inco^4 \rk^6 \log \dm }{\dm}$, we obtain that \whp
\begin{equation}
\begin{aligned}\label{fulls1t1T20}  
\twonorm{\tilde{T}_2}\leq \frac{\sss}{8} \fronorm{D^{1,i,m}}. 
\end{aligned}
\end{equation} 
Plugging the bounds~\eqref{fulls1t10} and~\eqref{fulls1t1T20} for $ \tilde{T}_1 $ and $ \tilde{T}_2 $ into~\eqref{fullef0}, we get that \whp 
\[
 \twonorm{e_m^T \er{0,i} \iter{1,i}}  \le \sss \biggr[ \frac{1}{8}\twoinfnorm{\ddf{1,m}} +  \frac{1}{8} \fronorm{D^{1,i,m}} +\frac{1}{8\times 2}\fulladd \inc \biggr].
\]
Further plugging this bound into~\eqref{fullkey0}, we obtain the first intermediate inequality~\eqref{eq: boundofDeltainitialstep}.

 
\subsubsection{Proof of Intermediate Inequality~\eqref{eq: boundofDinitalstep}}
\label{sec:proof_boundofDinitialstep}
 
To bound  $D^{1,i,m}$, we begin by recalling that by definition,
$$\iterm{1,j} = (\iter{1,j}\iterl{1,j}) \otimes \iter{1,j} = \svp \bigr [ \tru + \hpp{j} ( - \tru ) \bigr ], \qquad \text{for $ j=i $ and $ m $.} $$
Weyl's inequality ensures that the eigen gap $ \delta $ between the $r$-th and $(r+1)$-th eigenvalues of the matrix $\tru + \hpp{i} ( - \tru )  $ is at least $\delta \ge \sss-2\opnorm{\er{0,i}} \ge \frac{15}{16}\sss$ \whp, where the last inequality follows from the bound~\eqref{fullopnorm0} on $ \opnorm{\er{0,i}} $.
Let $W^{i,m} \defn (\hpp{i}-\hpp{m})(\tru)$, for which we have \whp
$\opnorm{W^{i,m}} \leq \opnorm{\er{0,i}}+\opnorm{\er{0,m}} \leq \frac{1}{16}\sss$ again thanks to~\eqref{fullopnorm0}. 
Recalling the definition of $ D^{1,i,m}  $ and applying the Davis-Kahan Theorem (Lemma \ref{daviskahan}), we obtain that \whp
 \begin{align}\label{fulls2t1D10}
 \fronorm{D^{1,i,m}} \leq \frac{\sqrt{2}\fronorm{ W^{i,m} \iter{1,m}}}{\delta-\opnorm{W^{i,m}}}
 \leq \frac{\sqrt{2}\fronorm{ W^{i,m} \iter{1,m}}}{(7/8)\sss} \leq\frac{2}{\sss} \fronorm{W^{i,m} \iter{1,m}}.
 \end{align}
To proceed, we consider two cases: $i=0$ and $i\not=0$.

\paragraph{The case $i=0$} 
In this case, the quantity $ \fronorm{W^{0,m} \iter{1,m}} $ is bounded in the inequality~\eqref{eq: boundofWinitialstep} to be proved below in Section~\ref{sec:proof_boundofWinitialstep}. Plugging~\eqref{eq: boundofWinitialstep} into the inequality~\eqref{fulls2t1D10},  we have \whp
\begin{align}\label{eq:D10m_bound}
\fronorm{D^{1,0,m}}  \leq \frac{1}{16}\twoinfnorm{\ddf{1,m}} + \frac{1}{16}\fullindb{}.
\end{align}

\paragraph{The case $i\neq 0$} 

For any orthonormal matrices $Q_1, Q_2 \in \real^{\rk \times \rk}$, the optimality of $D^{1,i,m}$ implies that
\begin{equation}
\begin{aligned}\label{eq: boundofDintiialiequal0}
\fronorm{D^{1,i,m}} \leq  \fronorm{ \iter{1,i}-\iter{1,m}Q_1}
\leq \fronorm{ \iter{1,i} -\iter{1,0}Q_2}+\fronorm{(\iter{1,0}Q_2Q_1^{-1}-\iter{1,m})Q_1}.
\end{aligned} 
\end{equation}
Choosing $Q_2$ to minimize $\fronorm{ \iter{1,i} -\iter{1,0} Q_2}$ and then $Q_1$ to minimize $\fronorm{\iter{1,0} Q_2Q_1^{-1}-\iter{1,m}}=\fronorm{\iter{1,0} -\iter{1,m}Q_1Q_2^{-1}}$, we have $\fronorm{ \iter{1,i} -\iter{1,0} Q_2}=\fronorm{D^{1,0,i}}$ and  $\fronorm{(\iter{1,0}Q_2Q_1^{-1}-\iter{1,m})Q_1}=\fronorm{\iter{1,0} -\iter{1,m}Q_1Q_2^{-1}}=  \fronorm{D^{1,0,m}}$. It then follows from~\eqref{eq: boundofDintiialiequal0} that \whp
\begin{equation}
\begin{aligned} \label{eq:D1im_bound}
\fronorm{D^{1,i,m}}\leq  \fronorm{D^{1,0,i}} + \fronorm{D^{1,0,m}}\niea \frac{1}{8} \twoinfnorm{\ddf{1,\infty}} +\frac{1}{8}\fullindb{},
\end{aligned}
\end{equation}
where step $\nia$ is  due to the inequality \eqref{eq:D10m_bound} proved above.

In view of the bounds~\eqref{eq:D10m_bound} and~\eqref{eq:D1im_bound} for two cases, we have established the second intermediate inequality~\eqref{eq: boundofDinitalstep}.

\subsubsection{Proof of Intermediate Inequality~\eqref{eq: boundofWinitialstep}}
\label{sec:proof_boundofWinitialstep}
  
We begin by observing that the operator $\hpp{0}-\hpp{m}$ is supported only on the $m$-th row and column. Decomposing the matrix $ (\hpp{0}-\hpp{m})(\tru) $ into two terms accordingly, we have
 \begin{align*}
 &\fronorm{ \big[ (\hpp{i}-\hpp{m})(\tru) \big] \iter{1,m}} \\
 =& \sqrt{ \sum_{j\leq r}\sum_{k\leq \dm, k\not=m} \Bigr( \big(1-\frac{\delta_{mk}}{p} \big)\tru_{mk}\iter{1,m}_{mj}\Bigr)^2 + \sum_{j\leq r}\Bigr(\sum_{k\leq \dm} \big(1-\frac{\delta_{mk}}{p} \big)\tru_{mk}\iter{1,m}_{kj}\Bigr)^2}\\
 \leq & \underbrace{\sqrt{ \sum_{j\leq r}  \big(\iter{1,m}_{mj} \big)^2\sum_{k\leq \dm, k\not=m} \Big( \big(1-\frac{\delta_{mk}}{p}\big)\tru_{km} \Big)^2}}_{B_1}+ 
 \underbrace{\sqrt{\sum_{j\leq r}\Bigr(\sum_{k\leq \dm} \big(1-\frac{\delta_{mk}}{p}\big)\tru_{mk}\iter{1,m}_{kj}\Bigr)^2}}_{B_2}.
 \end{align*} 
For the first term $B_1$, note that $\sqrt{\sum_{k\leq \dm, k\not=m} ((1-\frac{\delta_{mk}}{p})\tru_{km})^2}=\twonorm{\hp(\tru)e_m}$, whence 
 \[
 B_1 \leq  \twoinfnorm{\iter{1,m}}\twonorm{\hp(\tru)e_m}\leq (\twoinfnorm{\ddf{1,m}} +\twoinfnorm{\truf})\opnorm{\hp(\tru)}.
 \] 
Lemma~\ref{candes0} ensures that $\opnorm{\hp(\tru)}\leq2c\sqrt{\frac{\dm \log \dm}{p}}\frac{\inco \rk\ls}{\dm}$. Moreover, we have $ \twoinfnorm{\truf} \le \sqrt{\frac{\inco \rk}{\dm}}  $ and $p \gtrsim  \frac{\log(d)\inco^4\rk^6\cond^4}{ \dm}$ by assumption. Combining pieces, we obtain that \whp 
 \[
 B_1 \leq  \sss \biggr[\frac{1}{64}\twoinfnorm{\ddf{1,m}} +  \frac{1}{64}\lic{}\fulladd \sqrt{\frac{\inco \rk}{\dm}}\biggr].
 \]
For the second term $B_2$, we have 
 \[
 B_2 \leq \sqrt{r} \max_{j\leq r} \bigg| \sum_{k\leq \dm} \Big(1-\frac{\delta_{mk}}{p}\Big)\tru_{mk}\iter{1,m}_{kj} \bigg|
\niea \frac{\sss}{64}\twoinfnorm{\ddf{1,m}} +  \frac{\sss}{64}\fullindb{}, 
 \]
where in step $ \nia $ we follow the same arguments used in bounding $\tilde{T}_1$ in equation~\eqref{fulls1t100}. 
Combining the above bounds for $B_1$ and $B_2$, we obtain the third intermediate inequality~\eqref{eq: boundofWinitialstep}.

\subsection{Part 1(b): Induction step}
 \label{fullsubsection2} 
 
Suppose that the induction hypothesis~\eqref{eq: inductionsvphyothesis} holds for the $t$-th iteration, where $t\geq 1$. We shall prove that it also holds for the $(t+1)$-th iteration. Again, in the proof we shall show that various inequalities hold \whp for each fixed indices $i\in \{0, 1,\ldots, \dm\}$ and $m\in [\dm]$. By the union bound, these inequalities  hold simultaneously for all indices \whp 

\subsubsection{Operator norm bound~\eqref{eq:hypothesis_opnorm}}  

To bound $\er {t,i} \defn \hpp{i} (\iterm{t,i} -\tru) $, we shall  apply  Lemma \ref{yudong0}, which requires an $\ell_\infty$ norm bound on $\iterm{t,i}-\tru$. To this end, let us record several useful bounds. The $\ell_{2,\infty}$ bound~\eqref{eq:hypothesis_2inf} in the induction hypothesis  implies that $ \twoinfnorm{\ddf{t,i} } \leq \fulladd \lic{t} \inc $ \whp; consequently, $\twoinfnorm{\iter{t,i}} \leq2\inc$ \whp Wely's inequality together with the operator norm bound~\eqref{eq:hypothesis_opnorm} in the induction hypothesis implies  that $\opnorm{\iterl{t,i} - \true}\leq \opnorm{E^{t-1,i}}\leq \fulladd \lic{t} \sss $ \whp; consequently, $ \opnorm{\iterl{t,i}} \le 2 \ls $ \whp With these bounds, we may apply Lemma~\ref{lemma: inftwoinfnorm} to obtain that \whp
\begin{equation}
\begin{aligned}\label{eqn:iterm-truminfnorm}
\infnorm{\iterm {t}-\tru}\leq 20\kappa  \ls \fulladd\lic{t}\incc. 
\end{aligned}
\end{equation}
Using Lemma \ref{yudong0} in the following step $\nia$, we obtain that \whp
 \begin{equation}
 \begin{aligned} \label{fullopnorm} 
 \opnorm{E^{t,i}}  
 &\niea c \sqrt{\frac{\rk \dm \log \dm}{p}} \infnorm{\iterm{t,i}-\tru}\\
 & \leq 20\kappa \ls \fulladd\lic{t}\incc \cdot c \sqrt{\frac{\rk \dm\log \dm }{p}}
 \nieb \frac{\sss}{C\cond}\fulladd \lic{t+1}, 
 \end{aligned}
 \end{equation} 
where step $ \nib $ is due to the assumption $p \gtrsim \cond^6 \frac{\inco ^2 \rk^2\log \dm }{\dm}$. We have proved that the operator norm bound~\eqref{eq:hypothesis_opnorm} holds for the next iteration.

\subsubsection{$ \ell_{2,\infty}$ norm bound~\eqref{eq:hypothesis_2inf}} 

\yccomment{Below I changed all $\Lambda^{t,i}$ to $\Lambda^{t+1,i}$, and $ \G{t,i} $ to $ \G{t+1,i} $; please double check.} 
We focus on the $ m $-th row of the difference matrix $ \ddf{t+1,i} \defn \iter{t+1,i} - \truf \G{t+1,i} $. 
By definition, the matrices $\Lambda^{t+1,i}$ and $\iter{t+1,i}$ correspond to the top $r$ eigenvalues and eigenvectors of $\tru + \er{t,i}$, whence
 $$ (\tru +\er{t,i}) \iter{t+1,i} = \iter{t+1,i}\Lambda^{t+1,i}\implies \iter{t+1,i}= (\tru +\er{t,i}) \iter{t+1,i} ( \Lambda ^{t+1,i}) ^{-1}$$
Recalling $\tru = (\truf \true) \otimes \truf$, we the have  
 \begin{align*}
 \ddf{t+1,i}_{m\cdot} 
 = \iter{t+1,i}_{m\cdot}- \truf_{m\cdot }\G{t+1,i} 
 =&  e_m^T \truf \true (\truf)^T \iter{t+1,i}( \Lambda ^{t+1,i}) ^{-1}+e_m^T \er{t,i} \iter{t+1,i} ( \Lambda ^{t,i}) ^{-1}-e_m^T \truf \G{t+1,i}\\
 =  &e_m^T \truf \true \bigr [ (\truf)^T\iter{t+1,i}( \Lambda ^{t+1,i}) ^{-1}-(\true)^{-1}\G{t+1,i}\bigr]+e_m^T \er{t,i} \iter{t+1,i} ( \Lambda ^{t+1,i}) ^{-1}\\
 =  &\underbrace{e_m^T \truf\true\bigr [ (\truf)^T\iter{t+1,i}(\true)^{-1}-(\true)^{-1}\G{t+1,i}\bigr]}_{T_1}\\
 &+ \underbrace{e_m^T \truf \true (\truf)^T\iter{t+1,i}\bigr[( \Lambda ^{t+1,i}) ^{-1}-(\true)^{-1}\bigr]}_{T_2}+\underbrace{e_m^T \er{t,i} \iter{t+1,i} ( \Lambda ^{t+1,i}) ^{-1}}_{T_3}.
 \end{align*}  
We can bound the $ \ell_2 $ norms of $T_1,T_2,T_3$ by following  the same arguments used in Section \ref{sec: prelimstep0fullsvp} for bounding $T_1,T_2,T_3$ therein. Doing so yields that \whp
 \begin{equation}
 \begin{aligned}\label{fullkey}
   \twonorm{ \ddf{t+1,i}_{m\cdot}  }
   \leq &  \frac{1}{C} \fulladd \sqrt{\frac{\inco \rk}{\dm}} \lic{t+1} +  \twonorm{e_m^T \er{t,i} \iter{t+1,i}} \frac{C+1}{C\sss} .
 \end{aligned}
 \end{equation}
To proceed, we control $\twonorm{e_m^T \er{t,i} \iter{t+1,i}}$ and thereby establish the  $\ell_{2,\infty}$ error bound~\eqref{eq:hypothesis_2inf} for $ t+1 $. 
Note that if $i=m$, then $e_m^T\er{t,i} = 0$ by construction and we are done. In the following, we assume $i\not=m$.

\subsubsection{Bounding $\twonorm{e_m^T \er{t,i} \iter{t+1,i}}$ and establishing the $\ell_{2,\infty}$ bound}
 \label{fullstep2} 
 
Let $Q \defn  (\G{t+1,m})^T\G{t+1,i}\in \real^{r\times r}$, which satisfies $QQ^T=I$. The reason for this choice shall become clear later. We use the decomposition 
 \begin{equation}\label{fullef}
 \begin{aligned}
 &e_m^T \er{t,i} \iter{t+1,i}\\
 =& e_m^T \er{t,i} \iter{t+1,m}Q + e_m^T\er{t,i}(\iter{t+1,i}-\iter{t+1,m}Q)\\ 
 = &e_m^T\hpp{i}(M^{t,i}-\tru)\iter{t+1,m}Q + e_m^T\er{t,i}(\iter{t+1,i}-\iter{t+1,m}Q)\\
 =&\underbrace{  e_m^T\hpp{i}(M^{t,m}-\tru)\iter{t+1,m}Q}_{\tilde{T}_1} +\underbrace{e_m^T\hpp{i}(M^{t,i}-M^{t,m})\iter{t+1,m}Q}_{\tilde{T}_2} +\underbrace{ e_m^T\er{t,i}(\iter{t+1,i}-\iter{t+1,m}Q)}_{\tilde{T}_3}.
 \end{aligned}
 \end{equation}
 Below we control each of the terms $ \tilde{T}_1 $, $ \tilde{T}_2 $ and $ \tilde{T}_3 $.
 
\paragraph{Controlling $\tilde{T}_1$} 
 
We begin with the inequality
 \begin{equation}
 \begin{aligned}\label{fulls1t1}
 \twonorm{\tilde{T}_1} & = \twonorm{ e_m^T\hpp{i}(M^{t,m}-\tru)\iter{t+1,m}}\\
 & \leq \sqrt{r}\max_{j\leq r} \biggr|\sum_{k=1}^d \Big(1-\frac{1}{p}\delta_{mk}\Big) (M^{t,m}_{mk}-\tru_{mk})\iter{t+1,m}_{kj}\biggr| .
 \end{aligned}
 \end{equation} 
Note that $\iter{t+1,m}$ is independence of $\{\delta_{mk}, k\in[\dm] \}$ by construction. Therefore, Bernstein's inequality (Lemma~\ref{bern}) ensures that for each  $j\in [\rk]$, with probability at least $1- \dm^{-12}$, there holds the inequality
\begin{equation}\label{fulls1t1_intermediate}
 \begin{aligned}
 &\bigg| \sum_{k=1}^d \Big(1-\frac{1}{p}\delta_{mk}\Big) (M^{t,m}_{mk}-\tru_{mk})\iter{t+1,m}_{kj} \bigg| \\\leq  &\underbrace{\sqrt{\frac{C\log d}{p}} \twoinfnorm{\iter{t+1,m}} \bigr( \twonorm{M^{t,m}_{m\cdot}-\tru_{m\cdot}}\bigr) }_{\tilde{T}_{1a}}
 +\underbrace{ \frac{C\log\dm}{p} \twoinfnorm{\iter{t+1,m}}\bigr(\infnorm{M^{t,m}- \tru}\bigr)}_{\tilde{T}_{1b}}.
 \end{aligned}
\end{equation}
To further bound $ \tilde{T}_{1a} $ and $ \tilde{T}_{1b} $, we shall apply Lemma~\ref{lemma: inftwoinfnorm} to control the 
$\ell_2$ and $\ell_{\infty}$ norms of the vector $M^{t,m}_{m\cdot}-\tru_{m\cdot}$. To this end, we recall the bounds proved before~\eqref{eqn:iterm-truminfnorm}: $\twoinfnorm{\ddf{t,i} } \leq \fulladd \lic{t} \inc $, $\fronorm{\ddf{t}} \leq \fulladd \lic{t} \sqrt{\inco\rk}$, $\twoinfnorm{\iter{t,i}} \leq2\inc$,  $\opnorm{\iterl{t,i} - \true}\leq \opnorm{E^{t-1,i}}\leq \fulladd \lic{t} \sss $ and 
$ \opnorm{\iterl{t,i}} \le 2 \ls $ \whp. 
With these bounds, we apply Lemma~\ref{lemma: inftwoinfnorm} to obtain that \whp 
  \begin{align*}
  \twonorm{M^{t,m}_{m\cdot}-\tru_{m\cdot}} & \leq 2\ls\twoinfnorm{\Delta^{t,m}} +2\ls\inc \fronorm{\Delta^{t,m}}+6\cond\inc \opnorm{E^{t-1,i}}\ls,   \\ 
  \infnorm{M^{t,m}_{m\cdot}-\tru_{m\cdot}} & \leq 4\ls\twoinfnorm{\Delta^{t,m}}\inc +2\ls\inc \twoinfnorm{\Delta^{t,m}}+6\cond\incc\opnorm{E^{t-1,i}}\ls .
  \end{align*}
Also note that $\twoinfnorm{\iter{t+1,m}}\leq \sqrt{\frac{\inco \rk}{\dm}}  +\twoinfnorm{\Delta^{t+1,m}}$ since $\truf$ is $\inco$-incoherent.
Combining these bounds with the induction hypothesis on $\twoinfnorm{\Delta^{t,\infty}}, \opnorm{E^{t-1,\infty}}$ as well as the assumption $p \gtrsim \frac{\log (d)\cond^6 \mu ^4 r^6}{d}$, we obtain
 \begin{align*}
\max \big\{ \tilde{T}_{1a}, \tilde{T}_{1b} \big \} & \leq  \sss \Big( \frac{1}{128\sqrt{\rk}}\twoinfnorm{\ddf{t+1,m}} + \frac{1}{128\times 2\sqrt{\rk}}\fullindb{t+1}\Big) .
 \end{align*}
Plugging the above bound into~\eqref{fulls1t1} and~\eqref{fulls1t1_intermediate}, we conclude that \whp
 \begin{equation}\label{fulls1t1f}
 \begin{aligned}
 \twonorm{\tilde{T}_1} 
 & \leq \frac{\sss}{64}\twoinfnorm{\ddf{t+1,m}}+\frac{\sss}{128}\fullindb{t+1}.
 \end{aligned}
 \end{equation}

 \paragraph{The $\tilde{T}_2$ term} 
 
To bound the $\tilde{T}_2$ in inequality \eqref{fullef}, we first note that the matrix $M^{t,i}-M^{t,m}$ can be decomposed into three terms as
\begin{equation}
\begin{aligned}\label{eq: decompositionofMiMm}
M^{t,m}- M^{t,i} =&\, (\iter{t,m}\iterl{t,m}) \otimes \iter{t,m} - (\iter{t,i} \iterl{t,i})\otimes\iter{t,i} \\
=&\,(\iter{t,m}\iterl{t,m}) \otimes \iter{t,m} - (\iter{t,i} \G{t,m,i}\iterl{t,m}) \otimes\iter{t,m} \\
\,&+(\iter{t,i} \G{t,m,i}\iterl{t,m}) \otimes\iter{t,m} - (\iter{t,i}\iterl{t,i}\G{t,m,i})\otimes \iter{t,m}\\
\,& + (\iter{t,i}\iterl{t,i}\G{t,m,i}) \otimes \iter{t,m} - (\iter{t,i} \iterl{t,i}) \otimes\iter{t,i} \\
 =&\, (D^{t,m,i}\iterl{t,m}) \otimes \iter{t,m} - (\iter{t,i}S^{t,m,i}) \otimes \iter{t,m} + (\iter{t,i}\iterl{t,i}\G{t,m,i}) \otimes D^{t,m,i}.
\end{aligned}
\end{equation}
Therefore, we can bound $\tilde{T}_2$ by splitting it into three terms accordingly:
\begin{equation} \label{fulls1t2}
 \begin{aligned}
 \twonorm{\tilde{T}_2}=&\twonorm{e_m^T\hpp{i}(M^{t,i}-M^{t,m})\iter{t+1,m}Q} \\
 \leq &\underbrace{\twonorm{e_m^T\hpp{i}( (D^{t,m,i}\iterl{t,m}) \otimes \iter{t,m}  )\iter{t+1,m}Q}}_{\tilde{T}_{2a}} 
 +\underbrace{\twonorm{e_m^T\hpp{i}( (\iter{t,i}S^{t,m,i})\otimes \iter{t,m} )\iter{t+1,m}Q} }_{\tilde{T}_{2b}} \\
 &+ \underbrace{\twonorm{e_m^T\hpp{i}( (\iter{t,i}\iterl{t,i}\G{t,m,i}) \otimes D^{t,m,i} )\iter{t+1,m}Q} }_{\tilde{T}_{2c}}  .
 \end{aligned} 
\end{equation}
We control each of the above three terms. For $\tilde{T}_{2a}$, we have
 \begin{equation} \label{eq: fullsvpstep1T2r1b1b2}
 \begin{aligned}
 \tilde{T}_{2a} &=\twonorm{e_m^T\hpp{i}( (D^{t,m,i}\iterl{t,m}) \otimes \iter{t,m} )\iter{t+1,m}Q}\\
 & \niea\underbrace{\twonorm{e_m^T\hpp{i}( (D^{t,m,i}\iterl{t,m}) \otimes \iter{t,m} )\truf}}_{B_1} +\underbrace{\twonorm{e_m^T\hpp{i}( (D^{t,m,i}\iterl{t,m}) \otimes \iter{t,m} )\ddf{t+1,m}}}_{B_2}.\\ 
 \end{aligned} 
 \end{equation}
 The first term $B_1$ can be written explicitly as 
 \begin{equation} 
 \begin{aligned} \label{eq: fullsvpstep1T2r1b1} 
B_1
 =& \sqrt{ \sum_{l\leq \rk}\biggr(\sum_{k\leq \dm}\sum_{j\leq \rk}(D^{t,m,i}\iterl{t,m})_{mj}F^{t,m}_{kj} \Big(1-\frac{\delta_{mk}}{p}\Big) \truf_{kl}\biggr)^2}\\
 =& \sqrt{ \sum_{l\leq \rk}\biggr(\sum_{j\leq \rk}(D^{t,m,i}\iterl{t,m})_{mj}\sum_{k\leq \dm}F^{t,m}_{kj} \Big(1-\frac{\delta_{mk}}{p}\Big) \truf_{kl}\biggr)^2}\\
 \niea &  \sqrt{ \sum_{l\leq \rk}\twonorm{e_m^TD^{t,m,i}\iterl{t,m}}^2\biggr[\sum_{j\leq r}\biggr(\sum_{k\leq \dm}F^{t,m}_{kj} \Big(1-\frac{\delta_{mk}}{p}\Big) \truf_{kl}\biggr)^2\biggr]},\\
 \end{aligned} 
 \end{equation}
where we use Cauchy-Schwarz in step $(a)$. It follows that
\begin{equation}
\begin{aligned} \label{eq: fullsvpstep1T2r1b1second} 
B_1 \leq \twoinfnorm{D^{t,m,i}\iterl{t,m}} \cdot \rk \cdot  \max_{l,j\leq r} \left|\sum_{k\leq \dm}\iter{t,m}_{kj} \Big(1-\frac{\delta_{mk}}{p}\Big)\truf_{kl}\right|.
\end{aligned} 
\end{equation}
Recalling that $\delta_{mk}$ and $\iter{t,m}_{kj}$ are independent by construction, we apply Bernstein inequality (Lemma \ref{bern}) to obtain that \whp
 \begin{equation}\label{eq: fullsvpstep1T2r1b1bernstein}
 \begin{aligned} 
 \biggr|\sum_{k\leq \dm}\iter{t,m}_{kj} \Big(1-\frac{\delta_{mk}}{p}\Big) \truf_{kl}\biggr|&
 \leq   \sqrt{C\frac{\log \dm} {p}} \twoinfnorm{\truf} + \frac{C \log \dm}{p} \twoinfnorm{\truf}\twoinfnorm{\iter{t,m}}\\
 &\leq  \sqrt{C\frac{\inco \rk \log \dm}{p\dm}} + \frac{2C(\log \dm)\inco \rk}{pd}.
 \end{aligned} 
 \end{equation}
Combining inequalities \eqref{eq: fullsvpstep1T2r1b1second} and \eqref{eq: fullsvpstep1T2r1b1bernstein}, we have \whp
 \begin{equation} 
 \begin{aligned} \label{eq: fullsvpstep1T2r1b1final} 
 B_1\leq & \twoinfnorm{D^{t,m,i}\iterl{t,m}}r\left(\sqrt{C\frac{\inco \rk \log \dm}{p\dm}} + \frac{2C(\log \dm)\inco \rk}{pd}\right)
 \niea  \frac{\sss}{128}\fullindb{t+1},
 \end{aligned} 
 \end{equation}
where in step $\nia$ we use proximity condition~\eqref{eq:hypothesis_proximity} in the induction hypothesis, the bound $ \opnorm{\iterl{t,i}} \le 2 \ls $ proved before~\eqref{eqn:iterm-truminfnorm}, and the assumption that $p\gtrsim  \frac{\log(d)\inco^4\rk^6\cond^4}{ \dm}$.
 
For the term $ B_2 $, we follow a similar argument as in bounding $B_1$. In particular, we have \whp
 \begin{equation}\label{eq: fullsvpstep1T2r1b2}
 \begin{aligned} 
B_2 \defn &\twonorm{e_m^T\hpp{i}( (D^{t,m,i}\iterl{t,m}) \otimes \iter{t,m} )\ddf{t+1,m}}\\
 \niea  & \twoinfnorm{D^{t,m,i}\iterl{t,m}} \cdot \rk \cdot \max_{l,j\leq r} \biggr|\sum_{k\leq \dm}F^{t,m}_{kj} \Big(1-\frac{\delta_{mk}}{p}\Big) \ddf{t+1,m}_{kl}\biggr|\\
 \nieb& \frac{\sss}{128}\fullindb{t+1} + \frac{\sss}{128} \twoinfnorm{\ddf{t+1,m}}.\\
 \end{aligned} 
 \end{equation} 
Here in step $\nia$ we apply the same arguments as in \eqref{eq: fullsvpstep1T2r1b1} and \eqref{eq: fullsvpstep1T2r1b1second}; in step $\nib$ we apply the same argument as  in~\eqref{eq: fullsvpstep1T2r1b1bernstein} and~\eqref{eq: fullsvpstep1T2r1b1final}, noting in addition that $\ddf{t+1,m}$ is independent of $\delta_{mk}$ by construction. 
 
Plugging the bounds~\eqref{eq: fullsvpstep1T2r1b1final} and~\eqref{eq: fullsvpstep1T2r1b2} for $B_1$ and $B_2$ into the inequality~\eqref{eq: fullsvpstep1T2r1b1b2}, we obtain that  \whp 
 \begin{equation} \label{eq: fullsvpstep1T2r1final}
 \begin{aligned}
 \tilde{T}_{2a}  & \leq \sss \bigg[  \frac{1}{64}\fullindb{t+1}+\frac{1}{128}\twoinfnorm{\ddf{t+1,m}} \bigg] .
 \end{aligned} 
 \end{equation} 
 
We next consider the quantity $\tilde{T}_{2b} \defn \twonorm{e_m^T\hpp{i}( (\iter{t,i}S^{t,m,i})\otimes \iter{t,m} )\iter{t+1,m}Q}$ in~\eqref{fulls1t2}.
Note that \whp
$$\opnorm{S^{t,m,i}} 
\niea 4\cond \cdot \opnorm{E^{t-1,i}-E^{t-1,m}}
\nieb 8\ls\fulladd\lic{t},
$$
where step $ \nia $ follows from Lemma \ref{clemma2}, and step $ \nib $ follows from the triangle inequality and the induction hypothesis~\eqref{eq:hypothesis_opnorm} on $\opnorm{E^{t-1,\infty}}$.   
Combining the above bound with the bound $\twoinfnorm{\iter{t,i}}\leq 2\inc$ proved before~\eqref{eqn:iterm-truminfnorm}, we obtain that \whp
\[
\twoinfnorm{\iter{t,i} S^{t,m,i}} \leq  16\ls \fullindb{t}.
\]
To bound $\tilde{T}_{2b}$, we apply a similar argument as in bounding $\tilde{T}_{2a}$,  replacing each appearance of $D^{t,m,i}\Lambda^{t,m}$ by $\iter{t,i}(S^{t,m,i})$ everywhere and using the bound on $\twoinfnorm{\iter{t,i} S^{t,m,i}}$. Doing so gives that  \whp
 \begin{equation} \label{eq: fullsvpstep1T2r2final}
 \begin{aligned}
  \tilde{T}_{2b} & \leq \sss \bigg[ \frac{1}{64}\fullindb{t+1}+\frac{1}{128}\twoinfnorm{\ddf{t+1,m}} \bigg] .
 \end{aligned}
\end{equation} 
 
Finally, we turn to the quantity $\tilde{T}_{2c} \defn \twonorm{e_m^T\hpp{i}( \iter{t,i}\iterl{t,i}\G{t,m,i}\otimes D^{t,m,i} )\iter{t+1,m}Q}$ in~\eqref{fulls1t2}. Using triangle inequality and the fact that $Q = (\G{t+1,m})^T\G{t+1,i}$ is an orthogonal matrix, we get 
\begin{equation*}
\begin{aligned}
\tilde{T}_{2c} 
& \leq \underbrace{\twonorm{e_m^T\hpp{i}( (\iter{t,i}\iterl{t,i}\G{t,m,i}) \otimes D^{t,m,i})\truf}}_{B_1} + \underbrace{\twonorm{e_m^T\hpp{i}( (\iter{t,i}\iterl{t,i}\G{t,m,i}) \otimes D^{t,m,i})\ddf{t+1,m}}}_{B_2}.
\end{aligned}
\end{equation*}
For the first term $B_1$, using the same argument as in~\eqref{eq: fullsvpstep1T2r1b1} and \eqref{eq: fullsvpstep1T2r1b1second}, we have 
\begin{equation}
\label{eq:B1}
\begin{aligned}
B_1 \leq  & \twoinfnorm{ \iter{t,i}\iterl{t,i}\G{t,m,i}} \cdot \rk \cdot \max_{l,j\leq r} \biggr|\sum_{k\leq \dm}D^{t,m,i}_{kj} \Big(1-\frac{\delta_{mk}}{p}\Big) \truf_{kl}\biggr|.
\end{aligned}
\end{equation}
First consider the quantity inside the maximum above. Denoting by $\onevec$  the 
all one vector, we find that 
\begin{equation}
\begin{aligned}\label{eq: fullsvpstep1T2r3b1branch1} 
\biggr|\sum_{k\leq \dm}D^{t,m,i}_{kj} \Big(1-\frac{\delta_{mk}}{p}\Big) \truf_{kl}\biggr| 
& =  \abs{e_m^T\hp(\onevec\otimes \truf_{\cdot l})D_{\cdot j}^{t,i,m}}
\leq \opnorm{\hp(\onevec\otimes \truf_{\cdot l})}\twonorm{D_{\cdot j}^{t,i,m}}.
\end{aligned} 
\end{equation}
Applying Lemma \ref{candes0} with $\twoinfnorm{\truf}\leq \sqrt{\frac{\inco \rk}{\dm}}$, we  have \whp $\opnorm{\hp(\onevec\otimes \truf_{\cdot l})}\leq c\sqrt{\frac{\dm \log \dm}{p}}\sqrt{\frac{\inco \rk}{\dm}} $. 
Moreover, the proximity condition~\eqref{eq:hypothesis_proximity} in the induction hypothesis implies $\twonorm{D_{\cdot j}^{t,i,m}}\leq \fullindb{t} $, and the $ \ell_{2,\infty} $ bound~\eqref{eq:hypothesis_2inf} in the hypothesis implies   $\twoinfnorm{\iter{t,i}\iterl{t,i}\G{t,m,i}}\leq \twoinfnorm{\iter{t,i}}\opnorm{\iterl{t,i}}\leq 2\sss \inc$.
Plugging these bounds into~\eqref{eq:B1} and~\eqref{eq: fullsvpstep1T2r3b1branch1} and recalling  the assumption $p \gtrsim \frac{\cond^6\inco^4 r^6\log d}{\dm}$, we obtain that \whp 
\begin{equation*} 
\begin{aligned}\label{eq: fullsvpstep1T2r3b1final}
B_1
	\leq  \frac{\sss}{128}\fullindb{t+1}.
\end{aligned}
\end{equation*}
By a similar argument, we can bound the term $ B_2 \defn  \twonorm{e_m^T\hpp{i}( (\iter{t,i}\iterl{t,i}\G{t,m,i}) \otimes D^{t,m,i})\ddf{t+1,m}}$ as
\begin{equation*}
\begin{aligned}
B_2 
\leq  \twoinfnorm{\iter{t,i} \iterl{t,i}\G{t,m,i} } \cdot \rk \cdot \max_{l,j\leq r} \abs{e_m^T\hp(\onevec\otimes \ddf{t+1,m}_{\cdot l})D_{\cdot j}^{t,i,m}}
\leq  \frac{1}{32} \twoinfnorm{\ddf{t+1,m}}.
\end{aligned}
\end{equation*}
Combining the above bounds on $B_1$ and $B_2$, 
we obtain that \whp,
\begin{equation} 
  \begin{aligned}\label{eq: fullsvpstep1T2r3final}
  \tilde{T}_{2c}  & \leq \sss \bigg[ \frac{1}{32}\fullindb{t+1}+\frac{1}{32}\twoinfnorm{\ddf{t+1,m}} \bigg].\\
  \end{aligned} 
  \end{equation}  
  
Plugging the above bounds~\eqref{eq: fullsvpstep1T2r1final}, \eqref{eq: fullsvpstep1T2r2final} and \eqref{eq: fullsvpstep1T2r3final} on $\tilde{T}_{2a}, \tilde{T}_{2b}, \tilde{T}_{2c}$ into inequality~\eqref{fulls1t2}, we obtain that \whp
 \begin{align}\label{fullt2}
\twonorm{\tilde{T}_2} &\leq \sss \bigg[ \frac{1}{16}\twoinfnorm{\ddf{t+1,m}} + \frac{3}{32}\fullindb{t+1} \bigg].
 \end{align} 
 
\paragraph{The $\tilde{T}_3$ term}

To bound the third term $\tilde{T}_3$ in inequality \eqref{fullef},
we observe that 
 \begin{equation}\label{fullt3} 
 \begin{aligned}
 \twonorm{\tilde{T}_3} & = \twonorm{e_m^T\er{t,i}(\iter{t+1,i}-\iter{t+1,m}Q)}
 \leq  \underbrace{\twonorm{e_m^T\er{t,i}\ddf{t+1,i}}}_{\tilde{T}_{3a}} +\underbrace{\twonorm{e_m^T\er{t,i}\ddf{t+1,m}}}_{\tilde{T}_{3b}},
 \end{aligned}
 \end{equation}
where the last step is due to the choice of the orthogonal matrix $Q = (\G{t+1,m})^T\G{t+1,i}$.
 
Recalling $\er{t,i} \defn \hpp{i}(M^{t,i}-\tru)$, we write $\tilde{T}_{3a}$ explicitly as  
 \begin{equation*}\label{fullt31}
 \begin{aligned} 
 \tilde{T}_{3a} = &  \twonorm{e_m^T\hpp{i}(M^{t,i}-\tru)\ddf{t+1,i}} 
 = \sqrt{ \sum_{l\leq r}\biggr (\sum_{j\leq \dm} (M^{t,i}-\tru)_{ml} \Big(1-\frac{\delta_{mj}}{p}\Big) \ddf{t+1,i}_{jl}\biggr)^2}.
 \end{aligned} 
 \end{equation*}
Since $p\gtrsim \frac{\log \dm} {\dm}$, we have  
 $\sum_{j} \delta_{ij} \leq 2pd$ \whp uniformly for all $j$. It follows that
 \[
 \tilde{T}_{3a}\leq  \sqrt{\sum_{l\leq r} \big( 2 \dm \infnorm{M^{t,i}_{m\cdot}-\tru_{m\cdot}}\infnorm{\ddf{t+1,i}} +\dm \infnorm{M^{t,i}_{m\cdot}-\tru_{m\cdot}}\infnorm{\ddf{t+1,i}} \big)^2 }
 \]
The term $\infnorm{M^{t,i}_{m\cdot}-\tru_{m\cdot}}$ can be bounded using the \eqref{eqn:iterm-truminfnorm} proved in Section~\ref{fullsubsection1}: \whp
 \begin{align*}
 \infnorm{M^{t,i}_{m\cdot}-\tru_{m\cdot}}\leq 20\ls \fulladd \lic{t}\incc.
 \end{align*}
Putting together, we obtain that \whp   
 \begin{equation*}
 \begin{aligned} 
 \tilde{T}_{3a}
 \leq 
 3\sss \sqrt{r} \dm \times 20 \add\lic{t}\incc\twoinfnorm{\ddf{t+1,i}}
 \leq 
  \frac{\sss}{128}\twoinfnorm{\ddf{t+1,i}}.
 \end{aligned} 
 \end{equation*}
The term $\tilde{T}_{3b}$ in \eqref{fullt3} can be bounded using the same argument as above, which gives that \whp $  \tilde{T}_{3b} \leq  \frac{1}{128} \twoinfnorm{\ddf{t+1,m}}. $ 
Plugging the above bounds on~$\tilde{T}_{3a} $ and $ \tilde{T}_{3b} $  into~\eqref{fullt3}, we have \whp
 \begin{equation} \label{fullt33}
 \begin{aligned}
 \twonorm{\tilde{T}_3} 
 & \leq \frac{\sss}{64}\twoinfnorm{\ddf{t+1,\infty}}.
 \end{aligned} 
 \end{equation} 
 
Plugging the bounds~\eqref{fulls1t1f},\eqref{fullt2} and \eqref{fullt33} on $ \{\tilde{T}_i, i=1,2,3\} $ into inequality~\eqref{fullef}, we obtain that \whp
\begin{align}\label{eq:fullef_bound}
\twonorm{e_m^T \er{t,i} \iter{t+1,i}}
\leq \sss \biggr(\frac{1}{64} + \frac{1}{16} + \frac{1}{64} \biggr) \twoinfnorm{\ddf{t+1,m}}
+\sss \biggr( \frac{1}{128} +\frac{3}{32} \biggr) \fullindb{t+1} .
\end{align}
 
\paragraph{Completing proof of $\ell_{2,\infty}$ bound in the induction hypothesis} 
 
We now plug the bound~\eqref{eq:fullef_bound} on $ \twonorm{e_m^T \er{t,i} \iter{t+1,i}} $ into the inequality~\eqref{fullkey}, thereby obtaining that \whp 
 \begin{equation}\label{fullh1} 
 \begin{aligned}
 \twonorm{\Delta^{t+1,i}_{m\cdot}} 
 \leq &\frac{1}{4} \fullindb{t+1} + \frac{100}{99\sss} \biggr( \sss\frac{3}{32}\twoinfnorm{\ddf{t+1,m}} + \sss\frac{13}{128}\fullindb{t+1}  \biggr) \\
 \leq & \frac{1}{2}\fullindb{t+1}+ \frac{25}{99}\twoinfnorm{\ddf{t+1,\infty}}.\\
 \end{aligned}
 \end{equation}
Taking the maximum of both sides of~\eqref{fullh1} over $i,m$ and rearranging terms, we obtain that \whp
 \begin{equation}
 \begin{aligned} \label{eq: fullsvpell2infnormbound}
\twoinfnorm{\ddf{t+1,\infty}} \leq \fullindb{t+1}, 
 \end{aligned} 
 \end{equation}
thereby proving the $ \ell_{2,\infty} $ norm bound~\eqref{eq:hypothesis_2inf} in the induction hypothesis for $ t+1 $.

 \subsubsection{Proximity bound~\eqref{eq:hypothesis_proximity} and non-commutativity bound~\eqref{eq:hypothesis_commute} in the induction hypothesis} \label{fullsubsection4}
 
Recall that by definition,
\begin{align*}
\iterm{t+1,i} &= (\iter{t+1,i}\iterl{t+1,i} )\otimes \iter{t+1,i} = \svp \bigr [ \tru + \hpp{i} ( \iterm{t,i}- \tru ) \bigr ],\\
\iterm{t+1,m} &= (\iiter{t+1}{m} \iterl{t+1,m}) \otimes \iter{t+1,m} = \svp\bigr [ \tru + \hpp{m} ( \iterm{t,m}- \tru ) \bigr].
\end{align*}
By Weyl's inequality, the eigen gap $ \delta $ between the $r$-th and $(r+1)$-th eigenvalues of $\tru + \hpp{i} (M^{t,i} - \tru )  $ is at least $\delta \ge \sss -2\opnorm{\er{t,i}} \geq \sss-\frac{\sss}{16}$ \whp, where we use the bound~\eqref{fullopnorm} on $\opnorm{\er{t,i}}$.
 
We consider $\tru + \hpp{m} (M^{t,m} - \tru ) $ as a perturbed version of $\tru + \hpp{i} (M^{t,i} - \tru )$.  Let 
$$
W \defn \underbrace{(\hpp{i}-\hpp{m})(M^{t,m}-\tru)}_{W^{(a)}: \text{ discrepancy term}} +\underbrace{\hpp{i}(M^{t,i}-M^{t,m})}_{W^{(b)}: \text{ $ \ell_2 $ contraction term}}
$$
be the corresponding perturbation matrix, decomposed into two terms following the strategy outlined in equation~\eqref{pkey2} in Section~\ref{lootemplate}. Using the bound~\eqref{fullopnorm} on $\opnorm{\er{t,i}} $ again, we have \whp 
 $\opnorm{W} \leq \opnorm{\er{t,i}}+\opnorm{\er{t,m}} \leq \frac{\sss}{16}$. Consequently, Davis-Kahan's inequality (Lemma \ref{daviskahan}) ensures that \whp
 \begin{equation}
 \begin{aligned}\label{fulls3t1D1}
 \fronorm{D^{t+1,i,m}} 
 \leq \frac{\sqrt{2}\fronorm{ W\iter{t+1,m}}}{\delta - \opnorm{W}}
\leq \frac{2}{\sss} \big( \fronorm{W^{(a)}\iter{t+1,m}}+\fronorm{W^{(b)}\iter{t+1,m}} \big).
 \end{aligned}
 \end{equation}
To proceed, we control the two RHS terms to obtain a bound on $  \fronorm{D^{t+1,i,m}} $. We first consider the case $i=0$, deferring the case $ i \neq 0 $ to later.

\paragraph{The $W^{(a)}$ term}  
 
Introduce the shorthand $M_{\delta k}^{t,m} \defn (1-\frac{\delta_{mk}}{p})(M^{t,m}_{mk}-\tru_{mk})$. Noting that  $W^{(a)}$ is only nonzero at its $m$-th row and column, we have the following explicit expression:
 \begin{equation}\label{fullWa}
 \begin{aligned}
 \fronorm{W^{(a)}\iter{t+1,m}} =& \sqrt{ \sum_{j\leq r}\sum_{k\leq \dm, k\not=m} \bigr(M_{\delta k}^{t,m}\iter{t+1,m}_{mj}\bigr)^2 + \sum_{j\leq r}\bigr(\sum_{k\leq \dm} M_{\delta k}^{t,m}\iter{t+1,m}_{kj}\bigr)^2}\\
  \leq & \underbrace{\sqrt{ \sum_{j\leq r} (\iter{t+1,m}_{mj})^2\sum_{k\leq \dm, k\not=m} (M_{\delta k}^{t,m})^2}}_{T_1}+ \underbrace{\sqrt{\sum_{j\leq r}\bigr(\sum_{k\leq \dm} M_{\delta k}^{t,m}\iter{t+1,m}_{kj}\bigr)^2}}_{T_2}.
 \end{aligned}
 \end{equation}
For the term $T_1$, recalling that $\iter{t+1,m} =\ddf{t+1,m} + \truf\G{t+1,m} $ and  $ \er{t,m} \defn \hp(M^{t,m}-\tru)$, we find that 
 \[
 T_1 \leq \twoinfnorm{\iter{t+1,m}}\twonorm{\hp(M^{t,m}-\tru)e_m}\leq \left(\twoinfnorm{\ddf{t+1,m}} +\twonorm{\truf}\right)\opnorm{\er{t,m}}.
 \]
Combining this inequality with the assumption $\twoinfnorm{\truf}\leq \inc$ and the bounds~\eqref{fullopnorm} and~\eqref{eq: fullsvpell2infnormbound} on $\opnorm{\er{t,m}}$ and $\twoinfnorm{\ddf{t+1,m}}$, we obtain that \whp
 \[
 T_1\leq  \frac{\sss}{64}\fullindb{t+1}.
 \]
For the term $T_2$, we begin with the bound
 \begin{equation}
 \begin{aligned}
 T_2
 \leq  \sqrt{r} \max_{j\leq r}\biggr|\sum_{k\leq \dm} \Big(1-\frac{\delta_{mk}}{p}\Big) (M^{t,m}_{mk}-\tru_{mk})\iter{t+1,m}_{kj}\biggr|.
 \end{aligned} 
 \end{equation}
Bounding the last RHS using the same argument as in the derivation of~\eqref{fulls1t1f}, we find that \whp $T_2 \leq \frac{1}{64}\twoinfnorm{\ddf{t+1,m}}  +  \frac{\sss}{128}\fullindb{t+1}$. Combining with the bound~\eqref{eq: fullsvpell2infnormbound} on $\twoinfnorm{\ddf{t+1,m}}$ just proved above, we obtain that \whp
 \[
 T_2\leq  \frac{\sss}{64}\fullindb{t+1}.
 \]
Plugging the above bounds on $ T_1 $ and $ T_2 $ into~\eqref{fullWa}, we conclude that \whp
 \begin{equation}
 \begin{aligned}\label{eq: fullsvpwaterm}
 \fronorm{W^{(a)}\iter{t+1,m}}\leq  \frac{\sss}{32}\fullindb{t+1}.
 \end{aligned}
 \end{equation}
 
\paragraph{The $W^{(b)}$ term}
 
Specializing the decomposition in~\eqref{eq: decompositionofMiMm} to $ \iterm{t,0}  \defn \iterm{t} $, we have  
\[
 	\iterm{t,m}- \iterm{t} = (D^{t,m,0}\iterl{t,m}) \otimes \iter{t,m} - (\iter{t} S^{t,m,0}) \otimes \iter{t,m} + (\iter{t}\iterl{t}\G{t,m,0}) \otimes D^{t,m,0}.
\]
Accordingly, we split $\fronorm{W^{(b)}\iter{t+1,m}}$ into three terms as
 \begin{equation}\label{fullWb1}
 \begin{aligned} 
 \fronorm{W^{(b)}\iter{t+1,m}} & = \fronorm{\hp(M^t-M^{t,m})\iter{t+1,m}}\\
 &\leq \underbrace{\fronorm{\hp((D^{t,m,0}\iterl{t,m})\otimes \iter{t,m}  )\iter{t+1,m}}}_{R_1} +\underbrace{\fronorm{\hp( (\iter{t}S^{t,m,0})\otimes \iter{t,m} )\iter{t+1,m}} }_{R_2} \\
 &+ \underbrace{\fronorm{\hp( (\iter{t}\iterl{t}\G{t,m,0}) \otimes D^{t,m,0} )\iter{t+1,m}} }_{R_3}  .\\
 \end{aligned} 
 \end{equation}

For the term $R_1$, we introduce the shorthand $D_\Lambda^{t,m,0} \defn D^{t,m,0}\iterl{t,m}$ and further split $R_1$ into three terms: 
\begin{equation}\label{eq: fullsvpproxWbR1} 
\begin{aligned}
R_1 =&\fronorm{\hp(D_\Lambda^{t,m,0}\otimes \iter{t,m}  )\iter{t+1,m}} \\
\niea & \fronorm{\hp(D_\Lambda^{t,m,0}\otimes \iter{t,m}) \ddf{t+1,m}} + \fronorm{\hp(D_\Lambda^{t,m,0}\otimes \iter{t,m}) \truf}\\
\nieb &\underbrace{\fronorm{\hp(D_\Lambda^{t,m,0}\otimes \iter{t,m}) \ddf{t+1,m}}}_{R_{1a}} 
+\underbrace{\fronorm{\hp(D_\Lambda^{t,m,0}\otimes \ddf{t,m}) \truf}}_{R_{1b}}+ 
	\underbrace{\fronorm{\hp(D_\Lambda^{t,m,0}(\G{t,m})^T\otimes \truf ) \truf}}_{R_{1c}}, \\
\end{aligned} 
\end{equation}
where we use triangle inequality in steps $\nia$ and $\nib$, and the unitary invariance of $\fronorm{\cdot}$ in step $\nia$. 

We write the term $ R_{1a} $ in~\eqref{eq: fullsvpproxWbR1}  explicitly as 
\begin{equation} \label{eq: fullsvpproxWbR1B1}
\begin{aligned}
R_{1a}
= & \sqrt{ \sum_{j\leq d,l_1\leq r} \biggr(\sum_{l_2\leq r} \big(D_\Lambda^{t,m,0}\big)_{jl_2}\sum_{k\leq d}\iter{t}_{kl_2}
	\ddf{t+1,m}_{kl_1} \Big(1-\frac{\delta_{jk}}{p}\Big) \biggr)^2}\\ 
\niea & \sqrt{ \sum_{j\leq d,l_1\leq r} \bigg[\sum_{l_2\leq r}\Big(\big(D_\Lambda^{t,m,0}\big)_{jl_2}\Big)^2\bigg] \bigg[\sum_{l_2\leq r} \bigg(\sum_{k\leq d}\iter{t}_{kl_2}
	\ddf{t+1,m}_{kl_1} \Big(1-\frac{\delta_{jk}}{p}\Big) \bigg)^2\bigg]}\\
\leq   & \sqrt{ \sum_{l_1\leq r} \bigg[ \sum_{j\leq \dm,l_2\leq r}\Big(\big(D_\Lambda^{t,m,0}\big)_{jl_2}\Big)^2\bigg] \bigg[\max_{j\leq \dm} \sum_{l_2\leq \rk} \bigg(\sum_{k\leq d}\iter{t}_{kl_2}
	\ddf{t+1,m}_{kl_1}\Big(1-\frac{\delta_{jk}}{p}\Big) \bigg)^2\bigg]},\\
\end{aligned} 
\end{equation} 
where we use Cauchy-Schwarz in step $\nia$. 
Note that $\sum_{j\leq \dm,l_2\leq r}\big((D_\Lambda^{t,m,0})_{jl_2}\big)^2 =\fronorm{D_\Lambda^{t,m,0}}^2$ 
and  
\[
\sum_{l_1\leq r}\max_{j\leq \dm} \bigg\{ \sum_{l_2\leq \rk} \bigg(\sum_{k\leq d}\iter{t}_{kl_2}
\ddf{t+1,m}_{kl_1}\Big(1-\frac{\delta_{jk}}{p}\Big) \bigg)^2 \bigg\} 
\leq \rk^2 \max_{j\leq \dm,l_1,l_2\leq \rk}  \bigg( \sum_{k\leq d}\iter{t}_{kl_2}
\ddf{t+1,m}_{kl_1}\Big(1-\frac{\delta_{jk}}{p}\Big) \bigg)^2 .
\]
Therefore, we may continue from~equation~\eqref{eq: fullsvpproxWbR1B1} to obtain that \whp 
\begin{equation}
\begin{aligned}\label{eq: fullsvpproxWbR1B1step2}
R_{1a} 
\leq & \fronorm{D_\Lambda^{t,m,0}} \cdot \rk \max_{l_2,l_1\leq \rk,j\leq d} \biggr |\sum_{k\leq d}\iter{t}_{kl_2} \ddf{t+1,m}_{kl_1} \Big(1-\frac{\delta_{jk}}{p}\Big) \biggr| 
\\
\niea  &3\fronorm{D_\Lambda^{t,m,0}} \cdot \rk\dm\infnorm{\iter{t}}\infnorm{\ddf{t+1,m}}  \\
\nieb & \frac{\sss}{256}\fullindb{t+1}.
\end{aligned}
\end{equation}
Here in step $\nia$, we use the fact that whenever $p\geq 10\frac{\log \dm}{\dm}$, with probability $1-\dm^{-6}$, all rows have at most  $2pd$ observed entries, hence  $\big|\sum_{k\leq d}\iter{t}_{kl_2}
\ddf{t+1,m}_{kl_1}(1-\frac{\delta_{jk}}{p})\big| \leq 3\dm\infnorm{\iter{t}}\infnorm{\ddf{t+1,m}}$; in step $\nib$, we use the bounds on  $\twoinfnorm{\iter{t}}$ and $\opnorm{\iterl{t,m}}$ proved before~\eqref{eqn:iterm-truminfnorm}, the proximity bound~\eqref{eq:hypothesis_proximity} on $\fronorm{D^{t,0,m}}$ in the induction hypothesis, and the bound~\eqref{eq: fullsvpell2infnormbound} on $\twoinfnorm{\ddf{t+1,m}}$ proved previously.

For the term $R_{1b}=\fronorm{\hp(D_\Lambda^{t,m,0}\otimes \ddf{t,m}) \truf}$  in~\eqref{eq: fullsvpproxWbR1}, we apply a similar argument as in bounding $R_{1a}$ above, which gives that \whp
\begin{equation}\label{eq: fullsvpproxWbR1B2}
\begin{aligned} 
R_{1b}
\leq & \fronorm{D^{t,m}}\rk\times  3\dm \twoinfnorm{\truf}\twoinfnorm{\ddf{t,m}}
\leq  \frac{\sss}{256}\fullindb{t+1} .
\end{aligned} 
\end{equation}
For the term $R_{1c}=\fronorm{\hp(D_\Lambda^{t,m,0}(\G{t,m})^T\otimes \truf ) \truf}$ in~\eqref{eq: fullsvpproxWbR1}, we recall the assumption $p \gtrsim \frac{\cond^6\inco^4 r^6\log d}{\dm}$ and apply Lemma \ref{hpterm} to obtain that \whp
\begin{equation}\label{eq: fullsvpproxWbR1B3} 
\begin{aligned} 
R_{1c}
&\leq \frac{1}{512\kappa }\fronorm{D_\Lambda^{t,m,0}(\G{t,m})^T}
&\leq  \frac{\sss}{256}\fullindb{t+1}.
\end{aligned} 
\end{equation}
Plugging the above bounds~\eqref{eq: fullsvpproxWbR1B1step2}, \eqref{eq: fullsvpproxWbR1B2} and \eqref{eq: fullsvpproxWbR1B3} into the inequality~\eqref{eq: fullsvpproxWbR1},  we find that \whp
\begin{equation}
\begin{aligned}\label{eq: fullsvpproxWbR1final}
R_1 \leq  \frac{\sss}{64}\fullindb{t+1}.
\end{aligned}
\end{equation}

We next turn to the term $R_2 \defn \fronorm{\hp( (\iter{t}S^{t,m,0})\otimes \iter{t,m} )\iter{t+1,m}}$ in~\eqref{fullWb1}. Introducing the shorthand $S^{t,m,0}_F \defn \iter{t,m}(S^{t,m,0})^T$, we see $R_2=\fronorm{\hp( \iter{t}\otimes S^{t,m,0}_F )\iter{t+1,m}}$ and can be written explicitly as 
\begin{equation}\label{eq: fullsvpproxR2step1}
\begin{aligned}
R_2 & =\sqrt{ \sum_{j\leq d,l_1\leq r} \biggr(\sum_{l_2\leq r,k\leq d}\iter{t}_{jl_2}
	\big(S^{t,m,0}_F\big)_{kl_2} \iter{t+1,m}_{kl_1} \Big(1-\frac{\delta_{jk}}{p}\Big) \biggr)^2}\\
& \leq \sqrt{\rk}\max_{l_1\leq \rk} \underbrace{\sqrt{ \sum_{j\leq d} \biggr(\sum_{l_2\leq r,k\leq d}\iter{t}_{jl_2}
	\big(S^{t,m,0}_F\big)_{kl_2} \iter{t+1,m}_{kl_1} \Big(1-\frac{\delta_{jk}}{p}\Big) \biggr)^2}}_{R^\star_2 }.\\
\end{aligned} 
\end{equation} 
Note that $R^\star_2$ can be written compactly as $\fronorm{\sum_{l_2\leq r}\hp(\iter{t}_{\cdot l_2}\otimes \iter{t+1,m}_{\cdot l_1})	(S^{t,m,0}_F)_{\cdot l_2}}$, from which we obtain the bound
\begin{equation}
\begin{aligned} \label{eq: fullsvpproxWbr2step2}
 R^\star_2 
&\leq \rk \max_{l_1,l_2\leq\rk}\fronorm{\hp(\iter{t}_{\cdot l_2}\otimes \iter{t+1,m}_{\cdot l_1})(S^{t,m,0}_F)_{\cdot l_2}}\\
&\leq \rk \max_{l_1,l_2\leq\rk}\opnorm{\hp(\iter{t}_{\cdot l_2}\otimes \iter{t+1,m}_{\cdot l_1})}\fronorm{(S^{t,m,0}_F)_{\cdot l_2}}.\\
\end{aligned} 
\end{equation}
We have $\opnorm{\hp(\iter{t}_{\cdot l_2}\otimes \iter{t+1,m}_{\cdot l_1})}\leq 2c\sqrt{\frac{\dm \log \dm}{p} }\twoinfnorm{\iter{t}}\twoinfnorm{\iter{t+1,m}}$ \whp
by Lemma \ref{yudong0}, and  $\fronorm{S^{t,m,0}_F}\leq \fronorm{S^{t,m,0}}\opnorm{\iter{t,m}}\leq \fronorm{S^{t,m,0}}$ by construction. Plugging these bounds into~\eqref{eq: fullsvpproxWbr2step2}, we obtain that \whp 
\begin{equation}
\begin{aligned} \label{eq: fullsvpproxWbr2step3}
R^\star_2\leq  \rk \max_{l_1,l_2\leq\rk}2c\sqrt{\frac{\dm \log \dm}{p} }\twoinfnorm{\iter{t}}\twoinfnorm{\iter{t+1,m}}\fronorm{S^{t,m,0}}.
\end{aligned}
\end{equation}
To bound the last RHS, note that the  $\ell_{2,\infty}$ bound~\eqref{eq:hypothesis_2inf} in the induction hypothesis implies that $ \twoinfnorm{\iter{t,i}} \le 2 \sqrt{\frac{\inco \rk}{\dm}}, \forall i=0,\ldots,\dm$, which is valid for both $t$ and $t+1$ as we have established. Also recall the non-commutativity bound~\eqref{eq:hypothesis_commute} in the  induction hypothesis for  $\fronorm{S^{t,m,0}} =\fronorm{S^{t,0,m}}$, as well as the assumption that $p\gtrsim \frac{\cond^2\inco^2 r^5\log\dm}{\dm}$. 
Assembling these bounds into~\eqref{eq: fullsvpproxWbr2step3} and~\eqref{eq: fullsvpproxR2step1}, we obtain that \whp
 \begin{equation}\label{eq: fullsvpproxWbr2final}
 \begin{aligned}
 R_2 \leq \sqrt{r} R^\star_2
 \leq \frac{\sss}{64}\fullindb{t+1}.
 \end{aligned} 
 \end{equation}
 
Finally, consider the term  $R_3= \fronorm{\hp( (\iter{t}\iterl{t}\G{t,m,0})\otimes D^{t,m,0} )\iter{t+1,m}}$ in~\eqref{fullWb1}. Following the sames steps in~\eqref{eq: fullsvpproxR2step1}, \eqref{eq: fullsvpproxWbr2step2} and \eqref{eq: fullsvpproxWbr2step3} for bounding
$R_2$ by treating $\iter{t}\iterl{t,i}\G{t,m,0}$ as $\iter{t}$ and $D^{t,m,0}$ as $S^{t,m,0}_F$, we obtain that \whp
  \begin{equation}\label{eq: fullsvpproxWbr3}
  \begin{aligned}
  R_3 
  & \leq \rk^{\frac{3}{2}} \max_{l_1,l_2\leq\rk}2c\sqrt{\frac{\dm \log \dm}{p} }\twoinfnorm{\iter{t}\iterl{t,i}\G{t,m,0}}\twoinfnorm{\iter{t+1,m}}\fronorm{D^{t,m,0}}.\\ 
  \end{aligned} 
  \end{equation}
To bound the last RHS, we use the bound  $ \max\{\twoinfnorm{\iter{t}}, \twoinfnorm{\iter{t+1,m}} \} \le 2 \sqrt{\frac{\inco \rk}{\dm}}$ derived before~\eqref{eq: fullsvpproxWbr2final}, the proximity bound~\eqref{eq:hypothesis_proximity} on $\fronorm{D^{t,m,0}}$ in the induction hypothesis, the bound  $ \opnorm{\iterl{t,i}} \le 2 \ls $ derived before~\eqref{eqn:iterm-truminfnorm}, and the assumption $p\gtrsim \frac{\cond^6\inco^4 r^6\log\dm}{\dm}$. Doing so yields that \whp
  \begin{equation}\label{eq: fullsvpproxWbr3final}
\begin{aligned}
R_3&\leq \frac{\sss}{64}\fullindb{t+1}.
\end{aligned} 
\end{equation}

Plugging the above bounds~\eqref{eq: fullsvpproxWbR1final}, \eqref{eq: fullsvpproxWbr2final} and~\eqref{eq: fullsvpproxWbr3final} for $ \{R_i, i\in[3]\} $ into~\eqref{fullWb1}, we obtain that \whp 
 \begin{equation}\label{fullWb}
 \begin{aligned} 
 \fronorm{W^{(b)}\iter{t+1,m}} \leq \frac{\sss}{16}\fullindb{t+1}.
 \end{aligned} 
 \end{equation}
 
\paragraph{Completing proof of proximity and non-commutativity bounds in induction hypothesis}
 
Plugging the bounds~\eqref{eq: fullsvpwaterm} and \eqref{fullWb} on $\fronorm{W^{(a)}\iter{t+1,m}} $ and $ \fronorm{W^{(b)}\iter{t+1,m}} $ into inequality~\eqref{fulls3t1D1}, we get \whp
 \begin{equation}
 \begin{aligned}\label{fullproxi0}
 \fronorm{D^{t+1,0,m}} & \leq \frac{2}{\sss} (\fronorm{W^{(a)}\iter{t+1,m}}+\fronorm{W^{(b)}\iter{t+1,m}}) 
 \leq \frac{1}{2}\fullindb{t+1}.
 \end{aligned}
 \end{equation}
To bound $ \fronorm{D^{t+1,i,m}}  $ for $ i\neq 0 $, we follow the same argument used in deriving the inequality~\eqref{eq:D1im_bound}. This argument yields that \whp
 \begin{equation*}
 \begin{aligned}
 \fronorm{D^{t+1,i,m}} \leq \fronorm{D^{t+1,0,i}}  + \fronorm{D^{t+1,0,m}}  \leq \fullindb{t+1}.
 \end{aligned}
 \end{equation*}
Taking the maximum of both sides of the last two equations over $i,m$, we establish the proximity bound~\eqref{eq:hypothesis_proximity} in the induction hypothesis for $ t+1 $. 
 
Finally, to establish the non-commutativity bound~\eqref{eq:hypothesis_commute} in the induction hypothesis for $ t+1 $, we apply Lemma~\ref{clemma2} to obtain that \whp 
$$
\fronorm{S^{t+1,0,m}}  =\fronorm{S^{t+1,m,0}}\leq 4\cond \fronorm{W\iter{t+1,m}} \leq \frac{\ls}{2}\fullindb{t+1}.
$$
Taking maximum over $m$ on both sides proves the non-commutativity bound.\\

Recall we proved the operator norm and $ \ell_{2,\infty} $ norm bounds of the induction hypothesis for $ t+1 $ in~\eqref{fullopnorm} and~\eqref{eq: fullsvpell2infnormbound}, respectively. Therefore, we have completed the induction step and concluded that the induction hypothesis~\eqref{eq: inductionsvphyothesis} holds \whp for each $t \in [t_0]$, where $ t_0 \defn \tot $. Invoking Lemma \ref{lemma: inftwoinfnorm} and taking a union bound over $ t\in[t_0] $, we  deduce from the hypothesis~\eqref{eq: inductionsvphyothesis} that the desired error bound~\eqref{fullind1} on the original matrix $\iterm{t}$ holds \whp for the first $t_0 $ iterations; that is, $ \infnorm{ M^{t} - \tru} \leq \frac{ 1}{d}\lic{t}\ls, \forall t \in [t_0] . $

\subsection{Part 2: Bounds for an infinite number of iterations} \label{fullsubsection5}

\yccomment{Rewrote this section; please double check.}

The previous induction argument is insufficient for controlling all iterations $t > t_0 $, as the union bound would fail for an unbounded number of $t$'s. To control the error for an infinite number of \PGD iterations, we employ a different argument and establish a uniform  \emph{Frobenius} norm error bound as in~\eqref{fullind2}, i.e., \whp there holds $ \fronorm{\iterm{t}-\tru}
\leq \lic{t-t_0} \fronorm{\iterm{t_0}-\tru}, \forall t \geq t_0. $

 We prove the Frobenius bound~\eqref{fullind2} by induction. The base case $t=t_0$ trivially holds. Below we assume that~\eqref{fullind2} holds for all iterations $ t_0, t_0+1,\ldots, t  $.  For each $ t $,  define the shorthand $\td{t}\defn \iterm{t} - \tru$.

The proof relies on the following \emph{uniform} bound, which is proved in Appendix~\ref{sec:proof_prop1}.
\begin{lem}\label{prop1} 
	In the setting of $\SMC(\tru, p)$, suppose that $p \gtrsim \frac{\inco^2\rk^2\cond^6\log \dm}{\dm}$. Then with probability at least $1-4d^{-2}$, the bound 
	$$ 
	\Bigr | \inprod{M^+-\tru}{ M-\tru}-\frac{1}{p}\inprod{\proj(M^+-\tru)}{\proj( M-\tru)}\Bigr| 
	\leq \frac{1}{4}\fronorm{M^+-\tru}\fronorm{ M-\tru}
	$$ 
	holds simultaneously for all rank-$r$ matrices $M,M^+ \in \SymMat{d}$ satisfying $\max\{\infnorm{M}, \infnorm{M^+} \} \leq2\incc \ls (\tru) $ and $\max\{\fronorm{M-\tru},\fronorm{M^+-\tru} \}\leq\rhoo \times  \frac{1}{\cond^2} \ls(\tru)$. 
\end{lem}

We record several facts that are useful in verifying the premise of Lemma~\ref{prop1}. First note that
$$ 
\fronorm{ \td{t}} 
\niea \fronorm{\td{t_0}}
\leq \dm \infnorm{\td{t_0}}
\nieb \fulladd\frac{1}{2\dm^4}\incc,
$$ 
where step $ \nia $ follows from the induction hypothesis, and step $ \nib $ holds \whp and follows from specializing~\eqref{eqn:iterm-truminfnorm} to $t=t_0\defn\tot$.
It follows that
\begin{align}
\infnorm{\td{t}}&\leq \fronorm{\td{t}} \leq \rhoo\frac{1}{\dm^4}\incc \ls, \label{fullincc2t+1}\\
\infnorm{\iterm{t}} &\leq \infnorm{\tru} + \infnorm{\td{t}}  \leq 2\incc\ls .\label{fullincct+1}
\end{align}
Applying the \emph{uniform} bound in Lemma~\ref{yudong0}, and combining with the above bound on $ \infnorm{\td{t}} $ and the assumption $p\gtrsim \frac{\cond^6\inco^4\rk^6\log \dm}{\dm}$, we have  
$$
\opnorm{\er{t}} = \opnorm{\hp(\td{t})} 
\leq 2c \sqrt{\frac{2\rk \dm \log \dm}{p}} \infnorm{\td{t}}
\leq \rhoo\frac{1}{\cond^2 \dm^4} \ls . 
$$

By definition, $\iterm{t+1}$ is the best rank-$\rk$ approximation of $\tru+\er{t}$ in Frobenius norm, whence 
\begin{equation}
\begin{aligned}\label{fulloptsvd}
\fronorm{\iterm{t+1}-(\tru+\er{t})}^2 \leq& \fronorm{\er{t}}^2.\\
\end{aligned} 
\end{equation}
Consequently, we have 
\begin{equation}
\begin{aligned}\label{fulldt+1b}
\fronorm{\td{t+1}}
& \leq \fronorm{\iterm{t+1}-(\tru+\er{t})} +\fronorm{\er{t}}
\niea 2\fronorm{\er{t}}
\nieb 2\times \rhoo\frac{1}{\cond^2 \dm^3} \ls, \\
\end{aligned} 
\end{equation}
where step $ \nia $ follows from~\eqref{fulloptsvd}, and step $ \nib $ follows from $ \fronorm{\er{t}}\leq \sqrt{\dm} \opnorm{\er{t}} $ and the previous bound on $ \opnorm{\er{t}} $. We also have
\begin{equation}
\begin{aligned}\label{fulldt+1b1}
\infnorm{\iterm{t+1}}
& \leq \infnorm{\tru} + \infnorm{\td{t+1}}
\leq \incc + \fronorm{\td{t+1}}
\leq 2 \incc\ls . 
\end{aligned} 
\end{equation}

In view of the inequalities~\eqref{fullincct+1}, \eqref{fullincc2t+1}, \eqref{fulldt+1b}  and \eqref{fulldt+1b1}, we see that the premise of Lemma~\ref{prop1} is satisfied by letting $M^{+} = \iterm{t+1}, M = \iterm{t}$. Expanding the square on the LHS of inequality \eqref{fulloptsvd}, we have
\begin{equation*}
\begin{aligned} 
\fronorm{\td{t+1}}^2 &\leq -2 \inprod{\td{t+1}}{E^{t}}\\
& \leq -2 \inprod{\td{t+1}}{(\Id - p^{-1} \proj)(\td{t})}\\
&\leq  2 \Big| \frac{1}{p}\inprod{\proj \td{t+1}}{\proj\td{t}} - \inprod{\td{t+1}}{\td{t}} \Big|\\
&\leq 2 \times \frac{1}{4}\fronorm{\td{t+1}}\fronorm{\td{t}},\\
\end{aligned} 
\end{equation*} 
where in last step we apply Lemma~\ref{prop1}. The above inequality implies the contraction 
$
\fronorm{\td{t+1}}\leq \frac{1}{2}\fronorm{\td{t}},
$
hence the Frobenius norm bound~\eqref{fullind2} also holds for $ t+1 $. We have thus completed the induction step and established~\eqref{fullind2} for all the $ t\ge t_0. $ \\

We can now complete the proof of Theorem~\ref{fullmt0}. We have \whp
\begin{align*} 
\infnorm{M^t-\tru}\leq \fronorm{M^{t} -\tru }
&\niea \lic{t-t_0} \fronorm{M^{t_0}-\tru} \\
&\leq \lic{t-t_0} \cdot \dm \cdot \infnorm{M^{t_0}-\tru} \\
& \nieb \lic{t-t_0} \cdot \dm \cdot \frac{1}{\dm} \lic{t_0} \ls
= \lic{t} \ls,
\qquad \forall t\ge t_0,
\end{align*}
where step $ \nia $ follows from~\eqref{fullind1} and step $ \nib $ follows from~\eqref{fullind2}.
With the above bound, as well as the bound~\eqref{fullind1} for $ 1\le t\le t_0 $, we have established the claim in Theorem~\ref{fullmt0}.

\renewcommand{\hpp}[1]{\mathcal{H}_\ob^{(#1) }}

\section{Proof of Theorem \ref{mt1}}\label{sec: prfofthmmt1full}

In this section we prove Theorem~\ref{mt1} for \NNM using a combination of our \loo framework and the Golfing Scheme introduced in~\citep{recht2011simpler,gross2011recovering}. We assume that $\dm =\dm_1=\dm_2$ for simplicity; the proof of the general case follows the same lines. We shall make use of the auxiliary lemmas given in Appendix~\ref{auxillary lemmas}.


\subsection{Preliminaries}

Let the singular value decomposition of $ \tru $ be $\tru =U\Sigma V^T$. For a matrix $Z\in \real^{\dm\times \dm}$, we define the projections $\pj(Z) \defn UU^TZ+ZVV^T-UU^TZVV^T$ and 
$\pjp(Z)\defn Z-\pj(Z) = (I-UU^T)Z(I-VV^T)$. Introduce the shorthand $\rproj \defn \frac{1}{p}\proj$, which has the explicit expression  $[\rproj(Z)]_{ij}= \frac{1}{p}\delta_{ij}Z_{ij}$. We also define the operator $\hp \defn \Id  - \rproj = \Id - \frac{1}{p}\proj $ and the linear subspace $\mathcal{T} \defn  \{\pj(Z)\mid Z\in \real^{\dm\times \dm}\}$. For a linear map $ \mathcal{A} $ on matrices, its operator norm is defined as $\opnorm{\mathcal{A}} := \max_{Z:\fronorm{Z}=1}\fronorm{\mathcal{A}(Z)}$.  

We make use of the following standard result, which provides a deterministic sufficient condition for the optimality of $\tru$ to the nuclear norm minimization problem.
\begin{prop}[{\citealp[Proposition 2]{chen2015incoherence}}]
	\label{prop:suff_cond}
	Suppose that $p\geq \frac{1}{\dm}$. The matrix $\tru$ is the unique optimal solution to the \NNM problem~\eqref{mmininuc} if the following conditions hold:
	\begin{enumerate}
		\item $\opnorm{ \pj \rproj \pj -\pj } \leq \frac{1}{2}$. 
		\item There exists a dual certificate $Y\in \real^{\dm \times \dm}$ that satisfies $\proj(Y) = Y$ and 
	    \begin{enumerate}
			\item $\opnorm{\pjp(Y)}\leq \frac{1}{2}$,
			\item $\fronorm{\pj(Y)- UV^T} \leq \frac{1}{4\dm}$.
		\end{enumerate}
	\end{enumerate}
\end{prop}
The first condition in Proposition~\ref{prop:suff_cond} can be verified using the following well-known result from the matrix completion literature.
\begin{prop}[{\citealp[Theorem 4.1]{candes2009exact}}, {\citealp[Lemma 11]{chen2013low}}]
	\label{candes1}
	If $p \gtrsim \frac{\inco \rk \log \dm}{\dm} $, then with high probability $$\opnorm{ \pj \rproj \pj -\pj } \leq \frac{1}{64}.$$
\end{prop}

We are left to construct a dual certificate $Y$ such that Condition 2 in Proposition \ref{prop:suff_cond} is satisfied. Recalling the definition of the row-wise $\ell_2 $ norm in Section~\ref{sec:setup}, we further define the doubly $\ell_{2,\infty}$ norm 
$ \twotwoinfnorm{Z} := \max \bigr\{ \twoinfnorm{Z},\twoinfnorm{Z^T}\bigr\}, $ 
which plays a crucial role in the dual certificate construction.

\paragraph{Constructing the Dual Certificate} 

Our strategy is to construct the desired certificate $ Y $ by running an iterate procedure that uses the same set of random samples, and then apply \loo to analyze these correlated iterations. This procedure is warm-started by first employing the Golfing Scheme for $ \bigo(\log \inco \rk) $ iterations, each using an independent set of samples; as mentioned, doing so allows us to achieve tighter dependence on $ \inco \rk $,  

Now for the details. Set $k_0 \defn C_0\max\{1,\log (\inco \rk)\}$ for some large enough numerical constant $C_0$. Suppose that the set $\ob$ of observed entries is generated from $\ob = \cup_{t=1}^{k_0} \ob_t$, where for each $t$ and matrix index $(i,j)$ we have $\Prob[(i,j)\in \ob_t] = q \defn 1 -(1-p)^{\frac{1}{k_0}}$ independently of all others. Clearly this $\ob$ has the same distribution as the original model $\MC(\tru,p)$. 
Denote the projection $ \projk{t} $ by $[\projk{t}(Z)]_{ij}= Z_{ij} \indic\{(i,j) \in \ob_t \}$, and $\rprojk{t} \defn \frac{1}{q}\projk{t}$. 
Following our strategy, we use independent samples in the first $ k_0 $ iterations: set  $W^0 \defn  UV^T$ and 
\begin{align}\label{eq:defW}
	W^t \defn \pj \hpk{t}(W^{t-1}), \quad t =1,2,\dots, k_0 -1,
\end{align}
where $\hpk{t} = \Id - \frac{1}{q}\projk{t}$. We then use the same sample set $ \ob_{k_0}  $ in the next $\To \defn  2\log \dm +2 $ iterations: set  $Z^{0} = W^{k_0-1}$ and
\begin{align}\label{eq:defZ}
Z^{t} \defn \pj \hpk{k_0}(Z^{t-1}) = (\pj \hpk{_{k_0}})^t(W^{k_0-1}), \quad t = 1,2,\dots, \To -1.
\end{align} 
The final dual certificate $ Y $ is constructed by summing up the above iterates: set
\begin{align}\label{eq: defofY}
Y_1 \defn  \sum_{t=1}^{k_0-1} \rprojk{t}\pj(W^{t-1}),
\quad \quad 
Y_2 \defn \sum_{t=1}^{\To} \rprojk{k_0}\pj (Z^{t-1}),
\quad \text{and} \quad
Y \defn Y_1+ Y_2.
\end{align} 
Below we show that the matrix $ Y $ satisfies the conditions in Proposition \ref{prop:suff_cond}. 

\paragraph{Validating Condition 2(b)} Note $q \geq \frac{p}{k_0}\gtrsim \frac{\inco \rk \log \dm}{\dm}$ as $p \gtrsim \frac{\inco \rk \log\dm \log(\inco\rk)}{\dm}$. Applying Proposition~\ref{candes1} with $\ob$ replaced by $\ob_t$, we obtain that \whp,
\begin{align}\label{eq: Wt2b}
 \fronorm{W^t} \leq \opnorm{\pj\hpk{t}\pj}\fronorm{W^{t-1}}\leq \frac{1}{16}\fronorm{W^{t-1}} \quad \text{for all} \;t = 1,2,\dots,k_0-1,
 \end{align} and 
\begin{align} \label{eq: Zt2b}
\fronorm{Z^{t}}\leq \opnorm{\pj\hpk{k_0}\pj}\fronorm{Z^{t-1}}\leq \frac{1}{16}\fronorm{Z^{t-1}}\quad \text{for all} \; t = 1,2,\dots,\To.
\end{align}  
Using the last two inequalities, we obtain that \whp,
\begin{align}
\fronorm{\pj(Y) -UV^T} \overset{(a)}{=} \fronorm{Z^{\To}} \leq \lic{\To}\fronorm{Z_0}=\lic{\To}\fronorm{W_{k_0-1}}\leq \lic{\To}\fronorm{UV^{T}} \leq \frac{\sqrt{r}}{4\dm^2},
\end{align}
where step $ \nia $ follows from the fact that $ \pj(Y) -UV^T = -Z^{\To} $, which can be verified by definition and direct computation.
Therefore, Condition 2(b) in Proposition \ref{prop:suff_cond} is satisfied.

\paragraph{Validating Condition 2(a)} From the definitions of  $Y_1$,  $Y_2$ and $ Y $ in~\eqref{eq: defofY}, we have 
\begin{align*}
\opnorm{\pjp(Y)} & \le \sum_{t=1}^{k_0-1} \opnorm{\pjp(\rprojk{t} \pj - \pj)(\W^{t-1})}  + \sum_{t=1}^{\To} \opnorm{\pjp(\rprojk{k_0} \pj - \pj)(Z^{t-1})}  \\
&\niea \underbrace{\sum_{t=1}^{k_0-1}\opnorm{\hpk{t}\W^{t-1}}}_{T_1} +\underbrace{ \sum_{t=1}^{\To}\opnorm{\hpk{k_0}\Z^{t-1}}}_{T_2},
\end{align*}
where in step $ \nia$ we use the facts that $\opnorm{\pjp(A)}=\opnorm{(I - UU^T)A(I-VV^T)}\leq \opnorm{A}$ for any $A\in \real^{\dm\times \dm}$ and that $Z^{t-1},W^{t-1}\in \mathcal{T}$. To bound the term $T_1$, we follow exactly the same arguments in \cite[``Validating Condition 2(a)'', pp 12-13]{chen2015incoherence}, which gives that
$\label{eq: Y1bound}
T_1 \leq \frac{1}{4}
$
\whp
Introduce the shorthand
$$G \defn \frac{1}{2^{k_0-1}}\frac{1}{(\inco \rk)^{10}},$$
which will be used throughout the rest of the proof.
Turning to the term $T_2$, we claim that \whp, 
\begin{equation}
\begin{aligned}\label{eq: infzbound}
\infnorm{\Z^t} \leq \frac{1}{2^{t}}\incc G, \quad t= 0,1,\dots,\To.
\end{aligned}
\end{equation} 
We prove this bound later. Taking it as given for now, we apply the second inequality in Lemma \ref{yudong0} with $ \ob $ replaced by $ \ob_{k_0} $, which gives that \whp 
\begin{align*}
\opnorm{\hpk{k_0}(\Z^{t})} \leq \lic{t+3},\quad t=0,1,\dots,\To.
\end{align*}
Plugging into the expression of $T_2$, we obtain $ T_2 \leq \frac{1}{4}. $
Combining the bounds $T_1$ and $T_2$, we see that Condition 2(a) in Proposition~\ref{prop:suff_cond} is satisfied, thereby establishing Theorem~\ref{mt1}. \\

The rest of this section is devoted to proving the bound~\eqref{eq: infzbound}. We first show that the bound holds for $t=0$. Recall the definition of $ \W^t $ in~\eqref{eq:defW}. Note that $\infnorm{UV^T}\leq \incc$, and that $\ob_t$ is independent of $W^{t-1}$ for each $t\leq k_0$. Applying Lemma \ref{candes0} gives that \whp,
\begin{align}\label{eq: Wtinfnorm}
\infnorm{\W^t}\leq \frac{1}{16^t}\incc,\quad t = 1,2,\dots,k_0.
\end{align}
Using the above inequality and recalling the definition $\Z^0 \defn W^{k_0-1}$, we obtain that \whp,
\begin{equation}
\begin{aligned}\label{eq: nnmfullindcutionbase}
\infnorm{\Z^0}& \leq  G \cdot  \incc \cdot \frac{1}{(\inco \rk)^5} \leq \incc G ,\\ 
\fronorm{\Z^0}& \leq  d \cdot  G \cdot  \incc \cdot \frac{1}{(\inco \rk)^5} \leq G, \\
\twotwoinfnorm{\Z^0} & \leq   \sqrt{\dm} \cdot G \cdot \incc \cdot \frac{1}{(\inco \rk)^5}  \leq \inc G,
\end{aligned}
\end{equation}
provided that the constant $C_0$ in $k= C_0\log (\inco \rk)$ is sufficiently large. Therefore, the bound~\eqref{eq: infzbound} holds for $t=0$. Below we prove the bound  for $1\leq t\leq \To$ using the \loo technique.

\subsection{\Loo Analysis of the $ \Z^t $ Sequence }  

In this subsection, we abuse the notation and write $\ob_{k_0}$ as $\ob$, whose observation probability is $q  \defn 1 -(1-p)^{\frac{1}{k_0}} \gtrsim \frac{\inco \rk \log \dm}{\dm}$ since $p \gtrsim \frac{\inco \rk \log \dm \log (\inco \rk)}{\dm}$. For each $ w\in [\dm]\times[\dm] $, define the operator $\hpw: \real^{\dm\times\dm }\rightarrow \real^{\dm \times \dm}$ by
$$ 
\biggr (\hpw Z \biggr )_{ij} = 
\begin{cases} 
(1- \frac{1}{q}\delta_{ij} )Z_{ij},  & i\not =w_1 \text{ and }  j\not =w_2, \\
0, &  i = w_1 \;\text{or}\; j =w_2.
\end{cases}
$$ 
Let $\hpww \defn \hp - \hpw$.  For each $w\in[\dm]\times[\dm] $, we introduce  the \loo sequence
$$ \Z^{0,w}=\Z^0 ; \quad \Z^{t,w}=( \pj \hpw )(\Z^{t-1,w}), \;\; t = 1,2,\ldots.$$
By construction, this sequence is independent of $ \delta_{w_1w_2} $ and $ \hpww $, a property we crucially rely on below.
 
We first record a few technical lemmas that provide concentration bounds for the operators $\hp$ and $\hpw$. The first  lemma is proved in Appendix~\ref{sec: s2l2}.
\begin{lem}\label{s2l2}
	If $q \gtrsim \frac{\inco \rk\log \dm}{\dm}$, then we have \whp $$ \infnorm{\pj \hp(Z)} \leq \frac{1}{8}\inc \fronorm{Z}
	\qquad  \text{uniformly for all }  Z\in \mathcal{T}.$$ 
	The same statement holds with $\hp$ replaced by $\hpw$ for each $ w \in [\dm] \times [\dm] $.
\end{lem}

The lemma below is proved in Appendix~\ref{sec: leml2normbound}.
\begin{lem}\label{lem: l2normbound}
If $q\gtrsim \frac{\inco \rk \log \dm}{\dm}$, then for each $t=1,2,\ldots, \To$, we have \whp 
\begin{align*}	
&\max\bigr\{\twonorm{(\pj \hp(\Z^{t-1}))_{v_1\cdot}},\twonorm{(\pj \hp(\Z^{t-1}))_{\cdot v_2}}\bigr\}  
\leq \frac{1}{32} \twotwoinfnorm{\Z^{t-1,w}} + \frac{1}{32} \sqrt{\frac{\log \dm}{q}}\infnorm{\Z^{t-1,v}} \\
& \quad +\frac{1}{32}\inc \fronorm{\Z^{t-1}} + \twotwoinfnorm{\pj \hp (\Z^{t-1}-Z^{t-1,v})}
\qquad  \text{uniformly for all }  v \in [\dm] \times [\dm].
\end{align*} 
The same statement holds with $\hp$ replaced by $\hpw$  for each $ w \in [\dm] \times [\dm] $.
\end{lem}

The lemma below is proved in Appendix~\ref{sec: lem: linftynormbound}
\begin{lem}\label{lem: linftynormbound}
		If $q\gtrsim \frac{\inco \rk \log \dm}{\dm}$, then for each $t=1,2,\ldots, \To$, we have \whp 
		\begin{align*} 
		\biggr| \bigr( \pj \hp (\Z^{t-1})\bigr)_{v_1,v_2}\biggr| \leq& \frac{1}{32} \infnorm{\Z^{t-1,v}} + \frac{1}{32} \incc \fronorm{\Z^{t-1}} + \frac{1}{32} \inc \fronorm{\Z^{t-1}-\Z^{t-1,v}} \\
		&\qquad \qquad  \text{uniformly for all } v\in [\dm] \times [\dm].
		\end{align*}
		The same statement holds with $\hp$ replaced by $\hpw$ for each $ w \in [\dm] \times [\dm] $.
\end{lem}
The lemma below is proved in Appendix~\ref{sec: leml2lip}.
\begin{lem} \label{lem: l2lip}
		If $q\gtrsim \frac{\inco \rk \log \dm}{\dm}$, then we have \whp 
	$$\twotwoinfnorm{\pj \hp (Z)} \leq \frac{1}{32} \fronorm{Z}
	\quad  \text{uniformly for all }  Z\in \mathcal{T}.$$
	The same statement holds with $\hp$ replaced by $\hpw$ for each $ w \in [\dm] \times [\dm] $.
\end{lem}

The lemma below is proved in Appendix~\ref{sec: lemdiscrepancy}.
\begin{lem}\label{lem: discrepancy}
	If $q\gtrsim \frac{\inco \rk \log \dm}{\dm}$, then for each $w\in [\dm]\times [\dm]$ and each fixed $Z\in \mathcal{T}$, we have with probability at 
	least $1-d^{-4}$,
	$$\fronorm{\pj \hp^{(w)}(Z)} \leq \frac{1}{32}\twotwoinfnorm{Z} + \sqrt{\frac{\log \dm}{q}} \infnorm{Z}.$$
\end{lem} 
Finally, we note that the inequalities~\eqref{eq: Zt2b} and~\eqref{eq: nnmfullindcutionbase} imply that \whp
\begin{align}\label{eq: fornormnnmfullloo}
\fronorm{\Z^t} \leq \frac{1}{2^t}G,\quad \forall t = 1,2,\ldots, \To.
\end{align}

We are now ready to prove the inequality \eqref{eq: infzbound} by induction on $ t $. The induction hypothesis is
\begin{subequations}\label{eq:nnm_hypothesis}
\begin{align} 
\twotwoinfnorm{\Z^{t-1}} &\leq \lic{t-1} \inc G, \label{eq: twotwoinfnormindfullnnm}\\ 
\twotwoinfnorm{\Z^{t-1,w}}& \leq \lic{t-1} \inc G, \quad \forall w\in[\dm]\times [\dm], \label{eq: twotwoinfnormindfullnnmloo}\\
\infnorm{\Z^{t-1}} & \leq \lic{t-1}\incc G, \label{eq: infnormindfullnnm}\\
 \infnorm{\Z^{t-1,w}}& \leq \lic{t-1}\incc G, \quad \forall w\in[\dm]\times [\dm],\label{eq: infnormindfullnnmloo}\\ 
 \fronorm{\Z^{t-1}-\Z^{t-1,w}}& \leq \lic{t-1}\inc G, \quad \forall w\in[\dm]\times [\dm]. \label{eq: proximitynnmfullind}
\end{align} 
\end{subequations}
We have proved the base case in equation~\eqref{eq: nnmfullindcutionbase}, noting that $ \Z^{0,w}=\Z^0 $. Assuming that the bounds in~\eqref{eq:nnm_hypothesis} hold for $t-1$, we show below that each of them also holds for $t$ \whp

\paragraph{The $\twotwoinfnorm{\cdot}$ bound \eqref{eq: twotwoinfnormindfullnnm}} 
We focus on a fixed $w=(w_1,w_2)\in [\dm]\times [\dm]$ and bound the quantity 
$\twonorm{\Z^{t}_{w_1\cdot}} =\twonorm{(\pj \hp (\Z^{t-1}))_{w_1\cdot}}$. Lemma~\ref{lem: l2normbound} ensures that \whp
\begin{equation}
\begin{aligned}\label{eq: nnmloostepa2infnormlemma4application}
&\twonorm{\Z^{t}_{w_1\cdot}} \leq 
\underbrace{\frac{1}{32} \twotwoinfnorm{\Z^{t-1,w}} + \frac{1}{32}\sqrt{\frac{\log \dm}{q}}\infnorm{\Z^{t-1,w}}}_{T_1}
 + \underbrace{\frac{1}{32}\inc \fronorm{\Z^{t-1}}}_{T_2} 
 + \underbrace{\twotwoinfnorm{\pj\hp(\Z^{t-1}-Z^{t-1,w})}}_{T_3}.
\end{aligned}
\end{equation}
We bound the term $T_1$ using the induction hypothesis \eqref{eq: twotwoinfnormindfullnnmloo} and \eqref{eq: infnormindfullnnmloo}, and bound $T_2$ using
inequality~\eqref{eq: fornormnnmfullloo}.
For the term $ T_3 $, we apply Lemma~\ref{lem: l2lip} and the induction hypothesis~\eqref{eq: proximitynnmfullind} to obtain that \whp
\begin{equation*}
\begin{aligned}
T_3 \leq \frac{1}{32} \fronorm{\Z^{t-1}-Z^{t-1,w}}\leq \lic{t-1}\inc G.
\end{aligned}
\end{equation*}
Combining the above bounds, we obtain that
$
\twonorm{\Z^{t}_{w_1\cdot}}\leq  \lic{t}\inc G
$
 \whp
We can bound $\twonorm{\Z^{t}_{\cdot w_2}}$ in a similar way. Taking a union bound over all $w\in [\dm]\times [\dm]$ proves the inequality~\eqref{eq: twotwoinfnormindfullnnm} for $ t $.

\paragraph{The $\twotwoinfnorm{\cdot}$ bound \eqref{eq: twotwoinfnormindfullnnmloo}} 
Fix $v,w \in [\dm] \times [\dm]$. We have 
\begin{equation*}
\begin{aligned}
\twonorm{\Z^{t,w}_{v_1\cdot}} 
=\twonorm{(\pj \hpw (\Z^{t-1,w}))_{v_1\cdot}}
\leq \twonorm{(\pj \hpw(\Z^{t-1})_{v_1\cdot}}+ \twonorm{(\pj \hpw (\Z^{t-1}-Z^{t-1,w})_{v_1\cdot}}. \\  
\end{aligned}\
\end{equation*} 
The first RHS term can be bounded in a similar way as in the above proof of~\eqref{eq: twotwoinfnormindfullnnm}. 
The second term can be bounded using Lemma~\ref{lem: l2lip} and the induction hypothesis~\eqref{eq: proximitynnmfullind}. Combining the two bounds gives  $\twonorm{\Z^{t,w}_{v_1\cdot}} \leq \lic{t}\inc G$ \whp Taking a union bound over $v $ and $w$ proves the inequality~\eqref{eq: twotwoinfnormindfullnnmloo} for $ t $.

\paragraph{The $\infnorm{\cdot}$ bound \eqref{eq: infnormindfullnnm}}
Fix $w \in [\dm] \times [\dm]$. Lemma~\ref{lem: linftynormbound} ensures that \whp
\begin{align*}
|\Z^{t}_w | = \Big| \bigr(\pj \hp (\Z^{t-1})\bigr)_{w_1,w_2} \Big|
\leq \underbrace{ \frac{1}{32} \infnorm{\Z^{t-1,w}}}_{T_1} + \underbrace{\frac{1}{32}\incc \fronorm{\Z^{t-1}}}_{T_2}
+ \underbrace{\frac{1}{32}\inc \fronorm{\Z^{t-1}-Z^{t-1,w}}}_{T_3} .
\end{align*}
We bound $T_1$ and $T_3$ using the induction hypothesis \eqref{eq: infnormindfullnnm} and \eqref{eq: proximitynnmfullind}, respectively, and bound $ T_2 $ using inequality~\eqref{eq: fornormnnmfullloo}. Doing so gives $|\Z^{t}_w | \le \lic{t}\inc G$ \whp, and taking a union over $ w $ proves the inequality~\eqref{eq: infnormindfullnnm} for $ t $.

\paragraph{The $\infnorm{\cdot}$ bound \eqref{eq: infnormindfullnnmloo}} 
Fix $v,w\in [\dm]\times [\dm]$. We have
\begin{align*} 
|\Z^{t,w}_{v}| &= \Bigr|\bigr(\pj \hpw (\Z^{t-1,w})\bigr)_{v} \Bigr| 
\leq \Bigr|\bigr(\pj \hpw (\Z^{t-1})\bigr)_{v} \Bigr| + \Bigr|\bigr(\pj \hpw (\Z^{t-1,w}-\Z^{t-1})\bigr)_{v} \Bigr| .
\end{align*}
The first RHS term can be bounded in a similar way as in the above proof of~\eqref{eq: infnormindfullnnm}. 
To bound the second RHS term, we apply Lemma~\ref{s2l2} to obtain that \whp
\[
  \Bigr|\bigr(\pj \hpw (\Z^{t-1,w}-\Z^{t-1})\bigr)_{v} \Bigr| \leq \frac{1}{8}\inc\fronorm{\Z^{t-1,w}-\Z^{t-1}}.
\]
Combining the above bounds and the induction hypothesis~\eqref{eq: proximitynnmfullind}, we get that $|\Z^{t,w}_{v}| \le \lic{t}\inc G$ \whp Taking a union bound over $v $ and $w$ proves the inequality~\eqref{eq: infnormindfullnnmloo} for $ t $.

\paragraph{The proximity condition \eqref{eq: proximitynnmfullind}}
Fix $ w \in [\dm] \times [\dm]$. We have \whp
\begin{equation*}
\begin{aligned}
\fronorm{\Z^t - \Z^{t,w}} &= \fronorm{\pj \hp (\Z^{t-1}) -\pj \hpw(\Z^{t-1,w}) }\\ 
& \leq \fronorm{\pj \hp (\Z^{t-1} -\Z^{t-1,w}) } + \fronorm{\pj \bigr( \hp - \hpw\bigr)(Z^{t-1,w})} \\ 
&\niea \frac{1}{8} \fronorm{\Z^{t-1}-\Z^{t-1,w}} + \fronorm{\hpww (\Z^{t-1,w})}\\
& \nieb \frac{1}{8} \fronorm{\Z^{t-1}-\Z^{t-1,w}} +\biggr (\frac{1}{4}\twotwoinfnorm{\Z^{t-1,w}} +\sqrt{\frac{\log \dm}{q}}\infnorm{\Z^{t-1,w}} \biggr),
\end{aligned}
\end{equation*}
where we use Proposition~\ref{candes1} in step $\nia$ and Lemma \ref{lem: discrepancy} in step $\nib$. Bounding the last RHS using the induction hypothesis~\eqref{eq:nnm_hypothesis} and the fact that $q\gtrsim \frac{\inco \rk\log \dm}{\dm}$, we obtain $\fronorm{\Z^t - \Z^{t,w}} \leq \lic{t}\inc G$ \whp. Taking a union bound over $v $ and $w$ proves the inequality~\eqref{eq: proximitynnmfullind} for $ t $.

We have completed the induction step. Running this argument for  $\To \defn 2\log_2 d+2 $ steps and taking a union bound, we establish the claimed inequality~\eqref{eq: infzbound}.

\yccomment{I remove the conclusion section, as it is not very useful anyway.}

\section*{Acknowledgment}

L. Ding and Y. Chen were partially supported by the National Science Foundation CRII award 1657420 and grant 1704828. Y. Chen would like to thank Yuxin Chen for  inspiring discussion.

\appendixpage
\appendices

\section{Proof of Lemmas in Section~\ref{fullsvp}} \label{sec:proof_lemma_svp}

In this section, we prove the technical Lemmas~\ref{clemma1}, \ref{clemma2} and~\ref{prop1} used in the proof of \PGD in Section~\ref{fullsvp}.

\subsection{Proof of Lemma~\ref{clemma1}} \label{sec:proof_clemma1}
	
	\begin{proof} 
		We only prove the first inequality in the lemma. The other two inequalities can be proved similarly.
		
		 We make use of the following known result.
		\begin{lem}[{\citealp[Lemma 3]{abbe2017entrywise}}]
			\label{clemma}
			Under the setting of Lemma \ref{clemma1}, we have the bounds
			$$\opnorm{H} \leq 1, \qquad 
			\opnorm{H- G}^{\frac{1}{2}} \leq \frac{\opnorm{E\truf}}{\sss -\opnorm{E}} \qquad\text{and}\qquad
			\opnorm{\true H- H\true}\leq 2\opnorm{E}.$$
		\end{lem}
		Returning to the proof of the first inequality in Lemma~\ref{clemma1}, we have
		\begin{align*}
		\opnorm{\true G- G\true} & \niea \opnorm{\true G- \true H} +\opnorm{ H\true -G\true} +\opnorm{H\true -\true H} \\
		& \nieb  \opnorm{\true}\opnorm{G-H} +\opnorm{H-G}\opnorm{\true} + \opnorm{H\true -\true H} \\
		& \niec \Big( 2 + 2\ls\frac{1}{\sss-\opnorm{E}} \Big)\opnorm{E}, 
		\end{align*}
		where we use the triangle inequality in step $\nia$, the sub-multiplicative property of the operator norm $\opnorm{\cdot}$ in step $\nib$, and Lemma \ref{clemma} and the assumption $\opnorm{E}<\frac{1}{2}\sss$ in step $\nic$. 
	\end{proof}

\subsection{Proof of Lemma~\ref{clemma2}} \label{sec:proof_clemma2}

\begin{proof} 
	Using the definition of $\tilde{F}$, we have 
	\[
	\tilde{A}\tilde{F}   = \tilde{F}\tilde{\Lambda }\implies (A+ E)\tilde{F} = \tilde{F}\tilde{\Lambda} \implies  F_1 \Lambda _1 F_1^T \tilde{F} + F_2 \Lambda_2 F_2^T \tilde{F} +E \tilde{F}= \tilde{F} \tilde{\Lambda}.
	\]
	Right multiplying the last equation by $F_1^T$ on both sides, we get 
	\begin{align*} 
	\Lambda_1F_1^T \tilde{F} +  F_1^TE \tilde{F} = F_1^T \tilde{F} \tilde{\Lambda}
	\implies  \Lambda_1 H - H \tilde{\Lambda} = -F_1^T E \tilde{F} 
	\implies \fronorm {\Lambda_1 H - H \tilde{\Lambda}} =\fronorm{F_1^TE\tilde{F}}.
	\end{align*}
	Consequently, with $\Theta = \diag(\cos\theta_1,\dots,\cos \theta_r)$ denoting the matrix of the principal angles between the column spaces of $\tilde{F}$ and $F_1$,  we obtain
	\begin{align*}
	\fronorm{ \Lambda_1 G - G\tilde{\Lambda}} 
	& \leq \fronorm { \Lambda_1 G - \Lambda_1H}+ \fronorm{H\tilde{\Lambda} - G\tilde{\Lambda}} +\fronorm{\Lambda_1 H - H\tilde{\Lambda} }\\
	& \leq \lambda_1(A) \fronorm{G-H} + (\lambda_1(A)+\opnorm{E})\fronorm{G-H} +\fronorm{F_1^T E \tilde{F}}\\
	& \niea (2\lambda_1(A)+\opnorm{E})\fronorm{\sin \Theta} +\fronorm{F_1^T E \tilde{F}}\\
	& \leq \Bigr(\frac{ 2\lambda_1(A)+\opnorm{E}}{\lambda_{r}(A) -\opnorm{E}}+1\Bigr) \fronorm{E\tilde{F}},
	\end{align*}
	where step $\nia$ holds because $\fronorm{H-G}  = \fronorm{I - \cos \Theta}\leq  \fronorm{ (\sin \Theta)(\sin \Theta) } \leq \fronorm{\sin \Theta}$. This proves the Frobenius norm bound in the lemma. The operator norm bound can be proved in a similar way.
\end{proof} 

\subsection{Proof of Lemma~\ref{prop1}} \label{sec:proof_prop1}

	\begin{proof}
	In this proof, we make use of the auxiliary lemmas given in Appendix~\ref{auxillary lemmas}.
	
	Let $M = F\otimes F$ and $M^+ = F^+\otimes F^+$, where $F,F^+\in \real^{n\times r}$. Set $\W \defn M-\tru$ and $\W^+ \defn M^+-\tru$. Also let $Q \defn \inf_{O\in \real^{r\times r} ,OO^T=I}\fronorm{FO - \truf }, Q^+ \defn  \inf_{O\in \real^{r\times r} ,OO^T=I}\fronorm{F^+O - \truf } $.  Define $\Delta \defn FQ-\truf $ and $\Delta^+ \defn F^+Q-\truf $. We first record two useful inequalities. Lemma~\ref{Tu} ensures that 
	\begin{align}\label{e5}
	\fronorm{\Delta}\leq \frac{3}{\sqrt{\sss}}\fronorm{\W}\leq 3\times \rhoo\sqrt{\frac{\sss(\tru)}{\cond^2}} 
	\quad\text{and}\quad
	\fronorm{\Delta^+}\leq 3\frac{1}{\sqrt{\sss}}\fronorm{\W^+}\leq 3\times \rhoo \sqrt{\frac{\sss(\tru)}{\cond^2}} .
	\end{align}
	The differences $\W$ and $ \W^+ $ can be expressed as 
	$$ 
	\W = \truf \otimes \Delta +\Delta \otimes \truf    + \Delta \otimes \Delta
	\quad\text{and}\quad
	\W ^+= \truf \otimes \Delta^+ +\Delta^+ \otimes \truf    + \Delta^+ \otimes \Delta^+.
	$$ 
	With these expressions, we decompose the inner product of interest as 
	\begin{equation}
	\begin{aligned} 
	&\inprod{\W^+}{\W}-\frac{1}{p}\inprod{\proj(\W^+)}{\proj(\W)}\\
	=& \underbrace{\inprod{\truf \otimes \Delta +\Delta \otimes \truf}{\truf \otimes \Delta^+ +\Delta^+ \otimes \truf}-\frac{1}{p}\inprod{\proj(\truf \otimes \Delta +\Delta \otimes \truf)}{\proj(\truf \otimes \Delta^+ +\Delta^+ \otimes \truf)}}_{T_1}\\
	&+ \underbrace{\inprod{\truf \otimes \Delta +\Delta \otimes \truf}{\Delta^+\otimes \Delta^+}-\frac{1}{p}\inprod{\proj(\truf \otimes \Delta +\Delta \otimes \truf)}{\proj(\Delta^+\otimes \Delta^+)}}_{T_2}\\
	&+\underbrace{\inprod{\truf \otimes \Delta^+ +\Delta^+ \otimes \truf}{\Delta\otimes \Delta}-\frac{1}{p}\inprod{\proj(\truf \otimes \Delta^+ +\Delta^+ \otimes \truf)}{\proj(\Delta\otimes \Delta)}}_{T_3}\\
	&+\underbrace{\inprod{\Delta^+ \otimes \Delta^+}{\Delta\otimes \Delta}-\frac{1}{p}\inprod{\proj(\Delta^+\otimes \Delta^+)}{\proj(\Delta\otimes \Delta)}}_{T_4}.
	\end{aligned} 
	\end{equation}
	
	For $T_1$, we apply Lemma~\ref{yudong1} with $\epsilon \leq  \frac{1}{\cond}$, which ensures \whp,  
	$$|T_1| \leq \frac{1}{24}\fronorm{\truf \otimes \Delta}\fronorm{\truf \otimes \Delta^+} \leq 4\epsilon \ls(\tru)  \fronorm{\Delta}\fronorm{\Delta^+}\leq \frac{1}{8}\fronorm{\W^+}\fronorm{\W},$$
	where the last step follows from the bounds in~\eqref{e5}.
	
	To bound $T_2$, $ T_3 $ and $T_4$, we recall the premise of the proposition that $\max\{\{\infnorm{F\otimes F},\infnorm{F^+\otimes F^+}\}\leq 2\incc\ls(\tru)$, which implies that $\max \{ \twoinfnorm{F},\twoinfnorm{F^+}\}\leq \sqrt{\frac{2\inco \rk\ls(\tru)}{\dm}}$. It follows that $\twoinfnorm{\Delta}\leq \twoinfnorm{F}+\twoinfnorm{\truf} \leq 2\sqrt{ \frac{2\inco \rk\ls(\tru)}{\dm}}$ and similarly $\twoinfnorm{\Delta^+}\leq  2\sqrt{ \frac{2\inco \rk\ls(\tru)}{\dm}}$. These bounds allows us to use Lemmas~\ref{yudong1} and~\ref{yudong2}. In particular, for $ T_2 $, letting $\epsilon = \frac{1}{c\cond^2}$ for a sufficiently large constant $ c $, we have \whp,
	\begin{equation*}
	\begin{aligned}
	|T_2| & \leq \fronorm{ \truf \otimes \Delta +\Delta \otimes \truf}\fronorm{\Delta^+\otimes \Delta^+} + \fronorm{ (1/\sqrt{p}) \proj(\truf \otimes \Delta +\Delta \otimes \truf)} \fronorm{ (1/\sqrt{p}) \proj(\Delta^+\otimes \Delta^+)}\\
	& \niea  2\fronorm{\truf\otimes \Delta}\fronorm{\Delta^+}^2 + 4 \fronorm{ \truf \otimes \Delta}\sqrt{(1+\epsilon)\fronorm{\Delta^+}^4 +\epsilon\ls(\tru) \fronorm{\Delta^+}^2} \\
	&\leq 2 \fronorm{\Delta}\fronorm{\Delta^+}^2 + 4\sqrt{\ls(\tru)}\fronorm{\Delta} \Big( 2\sqrt{(1+\epsilon)}\fronorm{\Delta^+}^2 + 2\sqrt{\epsilon\ls(\tru)}\fronorm{\Delta^+} \Big) \\
	& \leq \Big( 24\sqrt{\frac{1+\epsilon}{\cond^2}}\times \rhoo+4\sqrt{\epsilon}\Big) \ls(\tru)\fronorm{\Delta}\fronorm{\Delta^+}\\
	&\nieb \frac{1}{24}\fronorm{\W^+}\fronorm{\W},
	\end{aligned}
	\end{equation*}
	where in step $ \nia $ we use Lemma \ref{yudong1} for the term $\fronorm{ \frac{1}{\sqrt{p}} \proj(\truf \otimes \Delta +\Delta \otimes \truf)} $ and Lemma \ref{yudong2} with the above $ \epsilon $ for  $\fronorm{\frac{1}{\sqrt{p}} \proj(\Delta^+\otimes \Delta^+)}$, and in step $ \nib $ we use~\eqref{e5} and the above choice of $ \epsilon $. Note that applying Lemma~\ref{yudong2} with the above $ \epsilon $ requires  $p\gtrsim \frac{\inco^2\rk^2(\log \dm)\cond^4}{\dm}$, which is satisfied under the premise of the proposition.  A similar argument shows that $|T_3|\leq\frac{1}{24}\fronorm{\W^+}\fronorm{\W}$ \whp
	
	For $T_4$,  with the same $ \epsilon $ as above and applying Lemma~\ref{yudong1}, we have \whp,
	\begin{equation*}
	\begin{aligned}
	|T_4| & \leq \fronorm{ \frac{1}{\sqrt{p}} \proj(\Delta\otimes \Delta)} \fronorm{\frac{1}{\sqrt{p}} \proj(\Delta^+\otimes \Delta^+)}+\fronorm{ \Delta}^2\fronorm{\Delta^+}^2\\
	& \leq \sqrt{(1+\epsilon)\fronorm{\Delta}^4 +\epsilon \ls(\tru)\fronorm{\Delta}^2} \sqrt{(1+\epsilon)\fronorm{\Delta^+}^4 +\epsilon\ls(\tru) \fronorm{\Delta^+}^2}+\fronorm{ \Delta}^2\fronorm{\Delta^+}^2\\
	&\leq (2\sqrt{(1+\epsilon)}\fronorm{\Delta}^2 + 2\sqrt{\epsilon\ls(\tru)}\fronorm{\Delta})(2\sqrt{(1+\epsilon)}\fronorm{\Delta^+}^2 + 2\sqrt{\epsilon\ls(\tru)}\fronorm{\Delta^+})+ \fronorm{\Delta}^2\fronorm{\Delta^+}^2\\
	& \leq \frac{1}{24}\fronorm{\W^+}\fronorm{\W},
	\end{aligned}
	\end{equation*}
	where the last step follows from the above choice of $ \epsilon  $ and the bounds~\eqref{e5}. 
	
	Combing the above bounds on $ T_1 $--$ T_4 $, we obtain that 
	\begin{align*}
	\bigg| \inprod{\W^+}{\W}-\frac{1}{p}\inprod{\proj(\W^+)}{\proj(\W)} \bigg|
	\leq \sum_{i=1}^4 |T_i|
	\leq \frac{1}{4}\fronorm{\W}\fronorm{\W^+},
	\end{align*}
	thereby completing the proof of Lemma~\ref{prop1}.
\end{proof}
\section{Proof of Lemmas in Section \ref{sec: prfofthmmt1full}} \label{sec: proofoflemmafornnm}

In this section, we prove the technical Lemmas~\ref{s2l2}--\ref{lem: discrepancy} used in the proof of \NNM in Section~\ref{sec: prfofthmmt1full}. For the first four lemmas, we prove the bounds for $\hp$ only; the bounds for $\hpw$ can be proved similarly. 

\subsection{Proof of Lemma \ref{s2l2}}\label{sec: s2l2}
\begin{proof}
	For each fixed $(i,j)\in [\dm]\times [\dm]$, we have 
	\begin{align*}
	e_i^T[\pj \hp(Z)]e_j &= \inprod{\pj \hp(Z)}{e_ie_j^T}
	\overset{(a)}{=} \inprod{ \pj \hp \pj (Z)}{ e_ie_j^T}  =\inprod{Z}{\pj \hp \pj (e_ie_j^T)},
	\end{align*}
	where the step $(a)$ is due to $Z\in \mathcal{T}$. Applying the Cauchy-Schwarz inequality, we obtain that  \whp
	\begin{align*}
	\inprod{Z}{\pj \hp \pj (e_ie_j^T)}	\leq \fronorm{Z}\fronorm{\pj\hp\pj(e_ie_j^T)}
	\nieb\frac{1}{16}\fronorm{Z}\fronorm{\pj(e_ie_j^T)}
	\leq  \frac{1}{8}\inc\fronorm{Z},
	\end{align*}
	where the inequality $\nib$ holds because $\opnorm{\pj \hp\pj }\leq \frac{1}{16}$ \whp by Proposition~\ref{candes1}, and the last inequality follows from direct computation using the definition $\pj$.
\end{proof}

\subsection{Proof of Lemma \ref{lem: l2normbound}}\label{sec: leml2normbound}
	\begin{proof} 
		We first recored two useful identities:
		\begin{subequations}
		\begin{align}
		UU^T \pj(Z) &= UU^T(UU^TZ+ZVV^T -UU^TZVV^T) = UU^TZ, \label{eq: ptproperty0}\\
		\pj(Z) VV^T &= (UU^TZ+ZVV^T -UU^TZVV^T)VV^T = Z VV^T. \label{eq: ptproperty0a}
		\end{align}
		\end{subequations}
		Now fix $ v\in[\dm]\times [\dm] $.  By definition of $\pj$ we have
		\begin{equation*}
		\begin{aligned}
		&\twonorm{\bigr ( \pj \hp (\Z^{t-1})_{v_1\cdot}\bigr)} \\
		 \leq& \twonorm{ e_{v_1}^TUU\hp (\Z^{t-1})} + \twonorm{e_{v_1}^TUU^T\hp(\Z^{t-1})VV^T} + \twonorm{e_{v_1}^T\hp(\Z^{t-1})VV^T}\\
		\leq& 2 \twonorm{e_{v_1}^TUU^T\hp(\Z^{t-1})} + \twonorm{e_{v_1}^T\hp(\Z^{t-1})VV^T}\\
		\leq& 2\underbrace{\twonorm{e_{v_1}^TUU^T\hp(\Z^{t-1})}}_{T_1}+ \underbrace{\twonorm{e_{v_1}^T\hp(\Z^{t-1,w})VV^T}}_{T_2}  + \underbrace{\twonorm{e_{v_1}^T \hp (\Z^{t-1}-\Z^{t-1,v})VV^T}}_{T_3}. 
		\end{aligned}
		\end{equation*}
		
		For $ T_1 $, we have \whp
		\begin{equation*}
		\begin{aligned}
		T_1 &  \overset{(a)}{=}2 \twonorm{e_{v_1}^T UU^T UU^T \hp \pj \Z^{t-1}} \\ 
		& \nieb2 \twonorm{e_{v_1}^TU}\fronorm{UU^T \pj \hp \pj (\Z^{t-1})}\\
		& \leq 2 \twoinfnorm{U} \opnorm{UU^T} \fronorm{\pj \hp\pj(\Z^{t-1})}\\
		& \niec \frac{1}{64}\inc \fronorm{\Z^{t-1}},
		\end{aligned}
		\end{equation*}
		where we use $\Z^{t-1}\in \mathcal{T}$ in step $\nia$, the identity \eqref{eq: ptproperty0} in step $\nib$, and Proposition~\ref{candes1} in step $(c)$.
		
		For $T_2$, note that $\Z^{t.v}$ is independent of $\{\delta_{v_1j}, j\in [\dm]\}$. Conditioning on $\Z^{t.v}$, we write $ T_2 $ as sum of independent vectors:
		\begin{align}
		\label{eq: boundt2lemma2nnm} 
		T_2 & = \twonorm{e_{v_1}^T \bigr[ \hp (\Z^{t-1,v})V\bigr]}  = \twonorm{ {\textstyle \sum_{1\leq j\leq \dm} } (1- q^{-1}\delta_{v_1j}) \Z_{v_1j}^{t-1,v} V_{j\cdot}}.
		\end{align}
		We compute the bounds
		\begin{equation*} 
		\begin{aligned}
		&\twonorm{(1-q^{-1}\delta_{v_1j})\Z_{v_1 j}^{t-1,v} V_j} 
		\leq \frac{1}{q} \infnorm{\Z^{t-1,v}}\twoinfnorm{V}  
		\leq \frac{1}{q}\inc \infnorm{\Z^{t-1,v}} =: B, 
		\end{aligned}
		\end{equation*}
		and
		\begin{equation*} 
		\begin{aligned}
		&\sum_{j}\Exs \biggr[\twonorm{ (1-q^{-1} \delta_{v_1j})\Z_{v_1j}^{t-1,v} V_{j\cdot}}^2 \biggr] 
		 = \sum_{j} \frac{1-q}{q}|\Z^{t-1,v}_{v_1j}|^2\twonorm{V_{j\cdot}}^2 
		 \leq \frac{\inco \rk }{q \dm} \twotwoinfnorm{\Z^{t-1,v}}^2 =: \sigma^2.
		\end{aligned}
		\end{equation*}
		Applying the vector Bernstein's inequality (Lemma \ref{lem: vecbernstein}) with the above $ B $ and $ \sigma^2 $, we have \whp 
		\begin{align*}
		\twonorm{T_2} & \leq c \biggr( \sqrt{\frac{\inco \rk \log \dm}{q\dm} }\twotwoinfnorm{\Z^{t-1,v}} + \frac{1}{q} \sqrt{\frac{\inco \rk \log \dm}{\dm}} \infnorm{\Z^{t-1,v}}\biggr)\\
		&\leq \frac{1}{256} \twotwoinfnorm{\Z^{t-1,v}} + \frac{1}{256} \sqrt{\frac{\log\dm}{q}} \infnorm{\Z^{t-1,v}},
		\end{align*}
		where we use $q\gtrsim \frac{\inco \rk\log \dm}{\dm}$ in the last step. 
		
		For $T_3$, we have
		\begin{equation*}
		\begin{aligned}
		\twonorm{e_{v_1}^T \hp (\Z^{t-1}-Z^{t-1,v})VV^T} & \overset{(a)} {=}\twonorm{e_{v_1}^T \pj \hp (\Z^{t-1}-Z^{t-1,v})VV^T} \\
		&\leq \twonorm{e_{v_1}^T \bigr ( \pj \hp (\Z^{t-1}-\Z^{t-1,v})\bigr)} \\
		& \leq \twotwoinfnorm{\pj \hp (\Z^{t-1}-Z^{t-1,v})},
		\end{aligned}
		\end{equation*}
		where we use the identity \eqref{eq: ptproperty0a} in step $(a)$. 
		
		Combining the above bounds for $ T_1,T_2 $ and $ T_3 $, we conclude that \whp $ 	\twonorm{\bigr ( \pj \hp (\Z^{t-1})\bigr)_{v_1\cdot}} $ is bounded as in the statement of the lemma. By a similar argument, the same bound holds $\twonorm{\bigr ( \pj \hp (\W^{t-1})_{\cdot v_2}\bigr)}$. The lemma then follows from a union bound over all $ v \in [\dm]\times[\dm]. $
	\end{proof}

	\subsection{Proof of Lemma \ref{lem: linftynormbound}} \label{sec: lem: linftynormbound}
	\begin{proof} 
		Fix  $ v\in[\dm]\times [\dm] $. By definition of $\pj$, we have the bound 
		\begin{align*}
		 \Bigr |\bigr( \pj \hp (\Z^{t-1})\bigr)_{v_1,v_2} \Bigr| 
		& 	\leq \underbrace{|e_{v_1}^T UU^T \hp (\Z^{t-1})e_{v_2} |}_{T_1} + \underbrace{ |e_{v_1}^T \hp (\Z^{t-1})VV^T e_{v_2} | }_{T_2} + \underbrace{|e_{v_1}^T UU^T \hp (\Z^{t-1})VV^T e_{v_2}|}_{T_3}.
		\end{align*}
		For $T_1$, we have \whp
		\begin{align*}
		T_1 & \leq | e_{v_1}^T UU^T \hp (\Z^{t-1,v})e_{v_2}| + | e_{v_1}^T UU^T \hp ( \Z^{t-1}-Z^{t-1,v})e_{v_2}|\\
		& \overset{(a)}{=} |e_{v_2}^T  UU^T \hp (\Z^{t-1,v})e_{v_2}|+ | e_{v_1}^T UU^T \pj \hp (\Z^{t-1}-\Z^{t-1,v})e_{v_2}|\\
		& \nieb |e_{v_1}^T UU^T \hp (\Z^{t-1,v})e_{v_2}| + \twonorm{e_{v_1}^T UU^T }\fronorm{ \pj \hp \pj(\Z^{t-1}-\Z^{t-1,v})}\twonorm{e_{v_2}} \\
		&\niec | e_{v_1}^T UU^T  \hp (\Z^{t-1,v})e_{v_2}| + \frac{1}{64}\inc \fronorm{\Z^{t-1}-\Z^{t-1,v}},
		\end{align*}
		where we use the equality \eqref{eq: ptproperty0} in step $(a)$, $\Z^{t-1},\Z^{t-1,v} \in \mathcal{T}$ in step $\nib$, and Proposition~\ref{candes1} in step $\nic$.
		To proceed, note that $\Z^{t-1,v}$ is independent of $\{\delta_{kv_2},k\in [\dm]\}$ by construction. Therefore, we have \whp 
		\begin{align*}
		| e_{v_1}^T UU^T  \hp (\Z^{t-1,v})e_{v_2}| & = \bigg| \sum_{k}(UU^T)_{v_1k}(1-\frac{1}{q}\delta_{kv_2})\Z^{t-1,v}_{kv_2} \bigg|\\
		& \niea c \sqrt{\frac{\log \dm}{\dm} \inc } \infnorm{\Z^{t-1,v} }+ \frac{\log \dm}{q} \incc \infnorm{\Z^{t-1,v}}\\
		& \nieb \frac{1}{64} \infnorm{\Z^{t-1,v}},
		\end{align*}
		where we use Bernstein's inequality (Lemma \ref{bern}) in step $\nia$, and $q\gtrsim \frac{\inco \rk \log \dm}{\dm}$ in step $\nib$. Thus, $T_1$ satisfies 
		$$  T_1 \leq \frac{1}{64} \infnorm{\Z^{t-1,v}}+ \frac{1}{64} \sqrt{\frac{\inco \rk}{\dm}} \fronorm{\Z^{t-1} -\Z^{t-1,v}} \quad \whp $$ 
		By a similar argument, the same bound holds for  $T_2$.
		For $T_3$, we have \whp
		\begin{align*}
		T_3 &\overset{(a)}{=}| e_{v_1}^T UU^T \pj \hp \pj(\Z^{t-1})VV^T e_{v_2}| \\
		&\leq \twonorm{e_{v_1}^T UU^T} \fronorm{\pj \hp \pj (Z^{t-1})} \twonorm{VV^Te_{v_2}} \\
		&\nieb \inc \cdot \frac{1}{64} \fronorm{\Z^{t-1}}\inc,
		\end{align*}  
		where we use the equality \eqref{eq: ptproperty0} in step $\nia$, and Proposition~\ref{candes1} in step $\nib$.
		Combining the above bounds for $T_1$, $T_2$ and $T_3$, and applying a union bound over all $ v\in[\dm]\times[\dm] $, proves the lemma. 
	\end{proof}

	\subsection{Proof of Lemma \ref{lem: l2lip}}\label{sec: leml2lip}
	\begin{proof} 
		Fix $j\in [\dm]$. Since $Z\in\mathcal{T}$, we have 
		\begin{align*}
		\twonorm{\pj \hp (Z)e_j} 
		&= \sup_{\twonorm{x}=1}\inprod{[\pj \hp(Z)]e_j}{x}  \\
		&=\sup _{\twonorm{x}=1}\inprod{\pj \hp\pj(Z)}{xe_j^T}
		\leq \fronorm{Z}\sup_{\twonorm{x}=1}\fronorm{\pj\hp\pj(xe_j^T)},
		\end{align*}
		Bounding the last RHS using Proposition~\ref{candes1}, we obtain that \whp
		\begin{align*}
		\twonorm{\pj \hp (Z)e_j} 
		& \le  \frac{1}{64}\fronorm{Z}\sup_{\twonorm{x}=1}\fronorm{(xe_j^T)} 
		 = \frac{1}{64} \fronorm{Z}.
		\end{align*}
		The same bound holds for $\twonorm{e_i \pj \hp(Z)}$ for each $ i\in[\dm] $ by a similar argument.
	    The lemma then follows from a union bound over $ i\in[\dm] $ and $ j\in[\dm] $.
	\end{proof}

	\subsection{Proof of Lemma \ref{lem: discrepancy}}\label{sec: lemdiscrepancy}
	\begin{proof} 
		Fix $ w\in[\dm] \times [\dm] $ and $ Z\in\mathcal{T} $. 
		By definition of $\pj $ and the fact that $ \opnorm{U}\le1,\opnorm{V}\le 1 $, we have 
		$$ 
		\fronorm{\pj \hpww (Z)} 
		= \fronorm{UU^T \hpww (Z) (I+VV^T) + \hpww (Z)VV^T } 
		\leq 2 \fronorm{U^T \hpww (Z)} + \fronorm{\hpww(Z)V}. 
		$$
		Below we bound the first RHS term; the second term can bounded similarly. Since only the $w_1$-th row and $w_2$-th column of $\hpww(Z)$ are non-zero, we have 
		\begin{align*}
		\fronorm{U^T \hpww (Z)} & \leq \twonorm{U^T \bigr [\hpww(Z) \bigr]_{\cdot w_2}} + \fronorm{ (U_{w_1\cdot})^T \bigr [ \hpww(Z)\bigr]}\\
		& = \twonorm{U^T \bigr [\hpww(Z) \bigr]_{\cdot w_2}} + \twonorm{U_{w_1\cdot}}\twonorm{\bigr [ \hpww(Z)\bigr]_{w_1\cdot}}\\
		& \leq \underbrace{ \twonorm{ {\textstyle \sum_{i=1}^{\dm}} U_{i\cdot} (1-q^{-1}\delta_{iw_2})Z_{iw_2}}}_{T_1} + \underbrace{\inc \twonorm{\bigr[\hpww(Z)\bigr]_{w_1\cdot}}}_{T_2}.
		\end{align*}
		Note that $ T_1 $ is the sum of independent vectors and has the same form as the term $ T_2 $ in equation~\eqref{eq: boundt2lemma2nnm} in the proof of Lemma \ref{lem: l2normbound}. Following the same arguments therein, we obtain that \whp  
		$$T_1 \leq \frac{1}{256}\sqrt{\frac{\log \dm}{q}} \infnorm{Z} + \frac{1}{256}\twotwoinfnorm{Z}.$$
		To bound $T_2$, define the operator $\pw: \real^{\dm \times \dm} \rightarrow \real^{\dm \times \dm}$ by $  (\pw Z)_{ij} = Z_{ij} \indic\{ i =w_1 \,\text{or}\, j=w_2 \} $.
		We have \whp 
		\begin{align*}
		T_2  \niea  \inc \opnorm{\hpww(Z)} 
		& \overset{(b)}{=} \inc \opnorm{\hp(\pw(Z))}\\
		& \niec\inc c \biggr( \sqrt{\frac{\log \dm}{q}}\twotwoinfnorm{\pw(Z)} + \frac{\log \dm}{q}\infnorm{\pw (Z)}\biggr) \\
		& \leq \frac{1}{256} \sqrt{\frac{\log \dm}{q}}\infnorm{Z} + \frac{1}{256}\twotwoinfnorm{Z},
		\end{align*}
		where step $ \nia $ follows from the inequality $ \twoinfnorm{A} \le \opnorm{A},\forall A $, step $ \nib $ follows from $ \hpww = \hp \pw  $, and step $ \nic $ follows from Lemma \ref{lem: fixedfrotwotwoinfinfnorm}.
		
		Combining the bounds for $ T_1 $ and $ T_2 $, we get that \whp $$ \fronorm{U^T \hpww(Z)} \leq \frac{1}{128} \sqrt{\frac{\log \dm}{q}}\infnorm{Z} + \frac{1}{128}\twotwoinfnorm{Z}.$$ 
		Plugging this inequality into the bound for $ \fronorm{\pj \hpww (Z)}  $, we prove the lemma.
	\end{proof} 
\renewcommand{\hpp}[1]{\mathcal{H}_\ob^{(-#1) }}

\section{Auxiliary lemmas}\label{auxillary lemmas}

In this section, we record several technical lemmas that are used in the proofs of our main theorem.

\subsection{Standard Concentration and Perturbation Bounds}
\label{sec:auxiliary_standard}

The lemmas in this subsection are standard concentration and matrix perturbation inequalities.

\begin{lem}[Bernstein]\label{bern}
	Let $X_1,\dots, X_n$ be $n$ independent random variable with $|X_i|\leq B$ with mean $0$. For each $ t>0 $ we have  $$ \Prob \big(| {\textstyle \sum_{i=1}^n X_i} |\geq t \big) \leq 2\exp\biggr( \frac{-t^2}{2\sum_{i=1}^n\Exs(X_i^2)+ \frac{2}{3}Bt}\biggr).$$ 
\end{lem}

\begin{lem}[Vector Bernstein {\citealp[Theorem 11]{gross2011recovering}}]
	\label{lem: vecbernstein}
	Let $\{v_k\}$ be a finite sequence of independent $ \dm $ dimensional random vectors. Suppose that $\Exs v_k =0$ and $\twonorm{v_k} \leq B$, a.s., and put $\sigma \geq \sum_k \Exs\twonorm{v_k}^2$. Then for all $t\geq 0$, 
	$$ \Prob \big( \twonorm { {\textstyle\sum _k v_k} } \geq t \big) \leq (n+1) \exp \biggr (-\frac{t^2}{2\sigma^2 +\frac{2}{3}Bt}\biggr).$$ 
\end{lem}

\begin{lem}[Matrix Bernstein \citep{tropp2012user}]
	\label{mbernstein}
	Consider a finite sequence $\{Z_k\}$ of independent  $n_1\times n_2$  random matrices that satisfy $\Exs Z_k = 0$ and $\opnorm{Z_k}\leq D$ a.s.
	Let $\sigma^2$ be the maximum of $\opnorm{\sum_k \Exs[Z_kZ_k^T]}$ and  $\opnorm{\sum_k \Exs Z_k^TZ_k}$. Then for all $t\geq 0$ we have 
	\begin{align*}
	\Prob\big( \opnorm{ {\textstyle \sum_k Z_k} }\geq t \big) \leq (n_1+n_2)\exp \biggr(-\frac{t^2}{2\sigma^2 +\frac{2}{3}Dt}\biggr).
	\end{align*}
\end{lem}

\begin{lem}[Subspace Distance Equivalence {\citealp[Proposition 2.2]{vu2013minimax}}]
	\label{sinthetadis}
	If the matrices $V_1,V_2\in \real^{d\times r}$ have orthonormal columns, then 
	$$ \frac{1}{2}\inf_{ Q\in \real^{r\times r}, QQ^T = I} \fronorm{V_1-V_2Q}^2 \leq 
	\fronorm{\sin (V_1,V_2)}^2 \leq \inf_{Q\in \real^{r\times r},QQ^T=I}\fronorm{V_1-V_2Q}^2,$$ where $(V_1,V_2)$ denotes the principal angles between the column spaces of $V_1$ and $V_2$.
\end{lem} 

\begin{lem}[Davis-Kahan sin$\Theta$ Theorem \citealp{davis1970rotation, li1998relative}]
	\label{daviskahan}
	Suppose that $A,W\in \real^{\dm\times \dm}$ are symmetric matrices, and $\tilde{A} = A+W$. Let $\delta = \lambda_{r}(A)- \lambda_{r+1}(A)$ be the gap between the top $r$-th and  $(r+1)$-th eigenvalues of $A$, and $U,\tilde{U}$ be matrices whose columns are the $r$ leading orthonormal eigenvectors of $A$ and $\tilde{A}$  respectively. If $\delta > \opnorm{W}$, then for any unitarily invariant norm $\matsnorm{\cdot}{}$, we have
	$$\matsnorm{\sin(U,\tilde{U} )}{} \leq \frac{\matsnorm{WU}{}}{\delta - \opnorm{W}}.$$
	Consequently, by Lemma \ref{sinthetadis} there exists a matrix $O\in \real^{r\times r}$ satisfying $OO^T=I$ and 
	$$ \fronorm{U - \tilde{U}O} \leq \frac{\sqrt{2}\fronorm{WU}}{\delta - \opnorm{W}}.$$
\end{lem}

\begin{lem}[{\citealt[Lemma 6]{ge2017no}}]
	\label{Tu}
	Given matrices $F,F^+\in \real^{\dm \times \rk}$, let $M = F\otimes F$ and $M^+= F^+\otimes F^+$. Also let $\Delta = F- F^+  O^\star$ where 
	$O^\star \defn \arg\min_{O \in \real^{\rk\times \rk},O\otimes O =I}\fronorm{F- F^+  O}$. We have 
	$$\fronorm{\Delta \otimes \Delta}^2 \leq 2\fronorm{M-M^+}^2
	\qquad\text{and}\qquad
	\sss(M^+)\fronorm{\Delta}^2 \leq \frac{1}{2(\sqrt{2}-1)}\fronorm{M-M^+}^2.$$
\end{lem}

\subsection{Technical Lemmas for Matrix Completion}
\label{sec:auxiliary_MC}

The lemmas in this subsection apply to the matrix completion settings $ \MC(\tru,p) $ and $ \SMC(\tru,p) $. The first three lemmas are known results in the literature.

\begin{lem}[{\citep[Lemma 2]{chen2015incoherence}}]
	\label{lem: fixedfrotwotwoinfinfnorm} Suppose $Z$ is a fixed $\dm \times \dm$ matrix. In the setting of $\MC(\tru,p)$, there exists a universal constant $c>1$ such that with probability at least $1-d^{-6}$
	$$\opnorm{\hp (Z)} \leq c \biggr ( \sqrt{\frac{\log\dm}{p}}\twotwoinfnorm{Z} + \frac{\log \dm}{p} \infnorm{Z} \biggr). $$
\end{lem} 

\begin{lem}[{\citealp[Lemma 4]{chen2015fast}}]
	\label{yudong1} 
	In the setting of $\SMC(\tru,p)$, for each $\epsilon \in (0,1)$, if $p\gtrsim \frac{\inco\rk\log\dm}{\epsilon^2\dm}$, then with probability at least $1-2d^{-3}$, the following holds: for all $H,G\in \real^{\dm\times \rk}$,
	\begin{align*}
	|p^{-1} \inprod{ \proj(\truf \otimes H)}{\proj (G\otimes \truf)} - \inprod{\truf \otimes H}{G\otimes \truf}| &\leq \epsilon \opnorm{\truf}^2\fronorm{H}\fronorm{G},\\
	|p^{-1} \inprod{ \proj(\truf \otimes G)}{\proj (\truf\otimes H)} - \inprod{\truf \otimes H}{\truf\otimes G}| &\leq \epsilon\opnorm{\truf}^2\fronorm{H}\fronorm{G}.
	\end{align*}
\end{lem} 

\begin{lem}[{\citealp[Lemma 5]{chen2015fast}}]
	\label{yudong2}
	In the setting of $\SMC(\tru,p)$, for each $\epsilon \in (0,1)$, if $p \gtrsim \frac{1}{\epsilon^2}(\frac{\inco^2 \rk^2}{\dm}+ \frac{\log\dm}{\dm})$, then with probability  at least $1-2\dm^{-4}$,  the following holds: for all $H\in \real^{\dm \times \rk}$ with $\twoinfnorm{H}\leq 6\inc$,
	$$ p^{-1} \fronorm{\proj(H\otimes H)}^2 \leq (1+\epsilon)\fronorm{H}^4 +\epsilon \fronorm{H}^2.$$
\end{lem}

\medskip

Below we state and prove several additional lemmas.
\begin{lem}\label{lemma: inftwoinfnorm}
	In the setting of $\SMC(\tru, p)$, let the eigenvalue 
	decomposition of $\tru = (\truf \true) \otimes \truf$ with $\truf \in 
	\real^{\dm \times r}$ and $\true \in \real^{r\times r}$ being diagonal. Let 
	$M = \svp(\tru + E)$ for some error matrix 
	$E\in \SymMat{\dm}$. Let the eigenvalue decomposition of 
	$M = (F\Lambda)\otimes F$ with $F\in \real^{\dm\times r}$ having orthonormal 
	columns and $\Lambda\in \real^{r\times r}$ being diagonal. 
	Suppose the rank-$r$ SVD of $(\truf)^T F$ is $ \bar{U} \bar{\Sigma}\bar{V}$. 
	Set $G \defn \bar{U}\bar{V}^T$, $\Delta \defn F -\truf G$.
	If $\opnorm{E}\leq \frac{1}{10}\sigma_r$,  then we have 
	$\opnorm{\Lambda}\leq \sigma_1+ \opnorm{E}$, and 
	\begin{align*}
	\infnorm{M-\tru}&\leq 2\twoinfnorm{\Delta}\twoinfnorm{\iter{}}\opnorm{\iterl{}}  +  (1+5\kappa)\twoinfnorm{\truf}^2\opnorm{E},\\
	\twoinfnorm{M-\tru}&\leq \twoinfnorm{\Delta}\opnorm{\iterl{}} 
	+\twoinfnorm{F}\opnorm{\iterl{}}\fronorm{\Delta} 
	+(1+5\kappa)\twoinfnorm{\truf}\opnorm{E}.
	\end{align*}
\end{lem}
\begin{proof}
	The inequality $\opnorm{\Lambda}\leq \sigma_1+ \opnorm{E}$ is a simple consequence of 
	Wely's inequality. 
	
	We next begin by decomposing the matrix $M-\tru$ into four terms as follows:
	\begin{equation}
	\begin{aligned}\label{eqn: decompose}
 M -\tru  & =  (F\iterl{})\otimes F -\tru\\
	&=  \underbrace{F\iterl{}F^T  - \truf G\iterl{}F^T}_{R_1}
	+ \underbrace{\truf G\iterl{}F^T-\truf \G{}\iterl{}(\truf G)^T}_{R_2}\\
	& \quad + \underbrace{\truf G\iterl{}(\truf G)^T -\truf G \true(\truf G)^T}_{R_3}
	+ \underbrace{\truf G \true(\truf G)^T  - \truf \true (\truf)^T }_{R_4} .\\
	\end{aligned} 
	\end{equation} 
	Let us first bound the matrices $R_i, i=1,2,3$ in terms of their infinity norms. We have
	\begin{equation} 
	\begin{aligned}\label{eqn: decomposeR123infinity} 
	&\infnorm{R_1}  \leq  \twoinfnorm{\Delta}\twoinfnorm{\iter{}\iterl{}}\leq \twoinfnorm{\Delta}\twoinfnorm{\iter{}}\opnorm{\iterl{}},   \\ 
	&\infnorm{R_2} \leq \twoinfnorm{\Delta} \twoinfnorm{\truf\G{}} \opnorm{\iterl{}}=\twoinfnorm{\Delta} \twoinfnorm{\truf} \opnorm{\iterl{}},\\
 	&\infnorm{R_3}  \leq \twoinfnorm{\truf \G{}}\opnorm{\iterl{} - \true}\twoinfnorm{\truf G}\niea   \twoinfnorm{\truf}\opnorm{E}\twoinfnorm{\truf},
	\end{aligned}
	\end{equation}
	where we use Wely's inequality in step $\nia$. We can also bound $R_4$ in term of the infinity norm:
\begin{equation} 
\begin{aligned}\label{eqn: decomposeR4infinity} 
 \infnorm{R_4} \leq  \twoinfnorm{\truf}\twoinfnorm{\truf}\opnorm{\G{}\true G^T -\true}
\niea5\kappa \twoinfnorm{\truf}\twoinfnorm{\truf}\opnorm{E} ,\\
\end{aligned} 
\end{equation}
where step $\nia$ follows from using Lemma \ref{clemma1} to bound 
$\opnorm{\G{}\true G^T -\true} = \opnorm{\G{}\true -\true G}$. Combining \eqref{eqn: decomposeR123infinity} and \eqref{eqn: decomposeR4infinity} yields 
the desired bound on $\infnorm{M-\tru}$. 

To bound the $ \ell_{2,\infty} $ norm of $M-\tru$, we first control $R_i$, $i=1,2,3$ as the following: 
\begin{equation} 
\begin{aligned}\label{eqn: decomposeR123twoinf} 
&\twoinfnorm{R_1}  \leq  \twoinfnorm{\Delta}\opnorm{\iter{}\iterl{}}\niea \twoinfnorm{\Delta}\opnorm{\iterl{}},   \\ 
&\twoinfnorm{R_2} \leq \twoinfnorm{\truf\G{}} \opnorm{\iterl{}\Delta^T}= \twoinfnorm{\truf} \opnorm{\iterl{}}\fronorm{\Delta},\\
&\twoinfnorm{R_3}  \leq \twoinfnorm{\truf \G{}}\opnorm{(\iterl{} - \true)\truf}\nieb   \twoinfnorm{\truf}\opnorm{E},
\end{aligned}
\end{equation}
where we use the fact that $\truf ,F $ has orthonormal columns $\nia$ and $\nib$, and Wely's inequality in step $\nib$. 
For $R_4$, we have the bound
\begin{equation} 
\begin{aligned}\label{eqn: decomposeR4twoinfinity} 
\infnorm{R_4} \leq  \twoinfnorm{\truf}\opnorm{(\G{}\true G^T -\true)(\truf)^T}
\niea5\kappa \twoinfnorm{\truf}\opnorm{\truf}\opnorm{E}\leq 5\kappa \twoinfnorm{\truf}\opnorm{E} ,
\end{aligned} 
\end{equation}
where step $\nia$ is due to Lemma \ref{clemma1}. Combining pieces yields the desired bound on $\twoinfnorm{M-\tru}$.
\end{proof}

\begin{lem}\label{hpterm} 
	In the setting of $\SMC(\tru, p)$, for each $ \epsilon \in (0,1) $ and fixed orthonormal matrix $\truf$, if $p\gtrsim \frac{\inco \rk \log \dm}{\epsilon^2\dm}$, then with probability at least $1-2\dm^{-3}$, we have 
	$$ \fronorm{\hp(\W\otimes \truf)\truf}\leq \epsilon \fronorm{\W}  \quad \text{for all } \W\in \mathbb{R}^{\dm\times \rk}. $$
\end{lem}
\begin{proof} 
	Using the variational characterization of Frobenius norm, we have 
	\begin{equation*}
	\begin{aligned}
	 \fronorm{\hp(\W\otimes \truf)\truf} 
	 &= \sup_{\fronorm{U}=1, U\in \real^{\dm\times \rk}} \inprod{\hp(\W\otimes \truf)\truf}{U}\\
	 & = \sup_{\fronorm{U}=1, U\in \real^{\dm\times \rk}} \inprod{\hp(\W\otimes \truf)}{U\otimes \truf}\\
	 & \overset{(a)}{=} \sup_{\fronorm{U}=1, U\in \real^{\dm\times \rk}} \inprod{\W\otimes \truf}{U\otimes \truf} -\frac{1}{p} \inprod{\proj(\W\otimes \truf)}{\proj(U\otimes \truf)},\\
	\end{aligned}
	\end{equation*}
	where we use the definition of $\hp$ in step $(a)$. Consequently, we find that with probability at least $1-2\dm^{-3}$,
	\[
	  \fronorm{\hp(\W\otimes \truf)\truf} \nieb \epsilon \sup_{\fronorm{U}=1, U\in \real^{\dm\times \rk}} \opnorm{\truf}^2\fronorm{\W}\fronorm{U}  \niec \epsilon \fronorm{W},
	\] 
	where step $ \nib $ follows from Lemma \ref{yudong1} and step $ \nic $ holds since $ \truf $ is orthonormal with $\opnorm{\truf}=1$.
	\end{proof} 
		
 	\begin{lem}\label{candes0} 
 		In the setting of $\SMC(\tru, p)$ or $\MC(\tru,p)$, there exists a numerical constant $c>1$ such that  if $p\geq \frac{\log \dm}{\dm}$, then for each fixed matrix $Z \in \real^{\dm \times \dm}$, the following inequalities hold \whp:
 		$$ \opnorm{\hpp{i}(Z)}\leq \opnorm{\hp(Z)} \leq 2c \sqrt{\frac{\dm \log \dm}{p}} \infnorm{Z}.$$
 	\end{lem}
 	\begin{proof} 
 		Since the $i$-th column and $i$-th row of $\hpp{i}(Z)$ is zero, we have $\opnorm{\hpp{i}(Z)}\leq \opnorm{\hp(Z)}$. Thus it remains to bound $ \opnorm{\hp(Z)} $.  Under $\MC(\tru,p)$, such a bound has been established in the literature using the matrix Bernstein inequality (Lemma~\ref{mbernstein}); see, e.g., \cite[Lemma 3.1]{candes2011robust} and~\citet[Lemma 12]{chen2013low} for the proof. 
 		The proof under the symmetric setting $\SMC(\tru,p)$ follows the same lines; we omit the details here.
 		\end{proof}

 		\begin{lem}[Uniform version of Lemma~\ref{candes0}]\label{yudong0}
 		In the setting of $\SMC(\tru,p)$ or $\MC(\tru,p)$, there exists a numerical constant $c>1$ such that	if  $p\geq \frac{\log \dm}{\dm}$, then  \whp the following bounds hold:
		\begin{align*}
 			\opnorm{\hp(U\otimes V)} &\leq 2 rc\sqrt{\frac{\dm \log \dm}{p}} \twoinfnorm{U}\twoinfnorm{V}, \quad \forall U,V \in \real^{\dm \times r}. \\
 			 \opnorm{\hp(A)} &\leq 2 c\sqrt{\frac{\rk \dm \log \dm}{p}} \infnorm{A},\quad \forall A\in \real^{\dm \times \dm}: \rank(A)\leq r, \\
 			 \opnorm{\hpp{i}(A)} &\leq 2c \sqrt{\frac{\rk \dm \log \dm}{p}} \infnorm{A},\quad \forall A\in \real^{\dm \times \dm}: \rank(A)\leq r, \forall i \in [\dm]. 				
 		\end{align*}
 		\end{lem}
 		\begin{proof}
 			The first and last inequalities in the lemma are immediate consequence of the second inequality. In particular, 
 			the second inequality follows from noting that $\sqrt{r}\leq r$ and $\infnorm{U\otimes V} \leq \twoinfnorm{U}\twoinfnorm{V}$. The last inequality follows from the fact $\hpp{i}$ sets the $i$-th row and column of $\hp(A)$ to $0$, hence  $\opnorm{\hpp{i}(A)}\leq \opnorm{\hp(A)}$. It remains to prove the second inequality in the lemma.
 			
 			Since $\rank(A)=r$, we have the decomposition $A = U\otimes V$ where $U,V \in \real^{\dm \times \rk}$. Let $u_i =(u_i^{(1)}, \dots,u_i^{(r)})$ and $v_j = (v_j^{(1)},\dots, v_j^{(r)})$ be the $i$-th row and $j$-th row of $U$ and $V$, respectively. We make use of the variational representation of the spectral norm:
 			$$\opnorm{ \hp(U\otimes V)} = \sup_{\twonorm{a}=\twonorm{b}=1}\inprod{\hp(U\otimes V)}{a\otimes b}.$$
 			Recalling the definition $\hp \defn \Id -\frac{1}{p}\proj$, we have
 			\begin{align*} 
 			\inprod{\hp (U\otimes V)}{a\otimes b} & = \inprod{U\otimes V}{a\otimes b} -\frac{1}{p}\inprod{\proj(U\otimes V)}{a\otimes b}\\
 			& = \sum_{i,j}\inprod{u_i}{v_j}a_ib_j - \frac{1}{p}\sum_{i,j\in \Omega} \inprod{u_i}{v_j} a_ib_j \\
 			& = \inprod{\onevec  \otimes \onevec}{(U\otimes V)\circ (a\otimes b)} -\frac{1}{p}\inprod{\proj(\onevec\otimes \onevec)}{(U\otimes V)\circ(a\otimes b)} \\
 			& =\inprod{\hp(\onevec \otimes \onevec)}{(U\otimes V)\circ (a\otimes b)},
 			\end{align*}
 			where $\circ$ denotes the Hadamard product. It follows that 
 			\begin{align*}
	 		\opnorm{ \hp(U\otimes V)} 
	 		&\leq \opnorm{ \hp(\onevec \otimes \onevec)}\sup_{\twonorm{a}=\twonorm{b}=1}\nucnorm{(U\otimes V)\circ (a\otimes b)}.
	 		\end{align*}
 		   On the one hand, Lemma \ref{candes0} applied to the fixed matrix $Z = \onevec \otimes \onevec $ guarantees that $ \opnorm{\hp(\onevec \otimes \onevec)} \leq 2c \sqrt{\dm \log(\dm)/p}$ \whp
 		   On the other hand, note that $[(U\otimes V)\circ (a\otimes b)]_{ij} = \inprod{a_iu_i}{b_jv_j}$, so the matrix $(U\otimes V)\circ (a\otimes b)$ has rank at most $\rk$. It follows that 
 		   \begin{align*}
 		   \nucnorm{ (U\otimes V)\circ (a\otimes b)} &\leq \sqrt{r} \cdot \fronorm{(U\otimes V)\circ (a\otimes b)}\\
 		   &=  \sqrt{r}\sqrt{ \sum_{i,j} (a_ib_j)^2( \inprod{u_i}{v_j}) ^2}\\
 		   &\leq \sqrt{r} \sqrt{ \sum_{i,j}(a_ib_j)^2 } \max_{i,j}\abs{\inprod{u_i}{v_j}} \\
 		   & =\sqrt{\rk}  \cdot 1 \cdot \infnorm{U\otimes V}.
 		   \end{align*}
 		   Combining pieces, we establish the second inequality in the lemma.
 		\end{proof}
 	
\renewcommand{\hpp}[1]{\mathcal{H}_\ob^{(#1) }}

\bibliographystyle{IEEEtranS} 

\bibliography{reference}

\end{document}